\pdfoutput=1
\documentclass[12pt]{article}
\usepackage{arxiv}
\usepackage[T1]{fontenc}
\usepackage[british]{babel}
\usepackage{csquotes}
\usepackage{microtype}
\usepackage{times}
\usepackage{xcolor}
\usepackage{mathtools,amssymb,amsthm}
\mathtoolsset{showonlyrefs,showmanualtags}
\usepackage{bm}
\usepackage{my_notation}
\usepackage[authoryear,square]{natbib}
\usepackage{graphicx}
\usepackage{placeins}
\usepackage{float}
\usepackage{algorithm}
\usepackage{algpseudocode}
\usepackage{booktabs}
\usepackage{enumitem}
\usepackage{array}
\usepackage{multirow}
\usepackage{etoolbox}
\usepackage{xurl}
\usepackage{hyperref}
\usepackage{bookmark}

\theoremstyle{plain}
\newtheorem{theorem}{Theorem}
\newtheorem{lemma}[theorem]{Lemma}
\newtheorem{proposition}[theorem]{Proposition}
\newtheorem{corollary}[theorem]{Corollary}
\theoremstyle{definition}
\newtheorem{definition}[theorem]{Definition}

\usepackage{paper_layout}
\renewenvironment{keywords}
 {\par\small\noindent\textbf{Keywords:}\ }
 {\par}
\definecolor{dkblue}{cmyk}{1,.54,.04,.19}
\hypersetup{
 colorlinks=true,
 linkcolor=black,
 citecolor=dkblue,
 urlcolor=dkblue,
 pdftitle={Minimax-Optimal Small-Loss Bounds for Online Linear Optimisation on lp Balls},
 pdfauthor={Jack J. Mayo},
 pdfsubject={Sharp cancellation-slack and dimension dependence for full-information online linear optimisation},
 pdfkeywords={online linear optimisation, small-loss regret, lp balls, cancellation slack, minimax regret}
}
\renewcommand{\R}{\mathbb R}
\renewcommand{\E}{\mathbb E}
\newcommand{\one}{\mathbf 1}
\newcommand{\ip}[2]{\mathopen{}\mathclose\bgroup\left\langle #1,#2\right\rangle\egroup}
\newcommand{\ipM}[3]{\mathopen{}\mathclose\bgroup\left\langle #1,#2\right\rangle_{#3}\egroup}
\newcommand{\Reg}{\operatorname{Reg}}
\newcommand{\FTL}{\mathrm{FTL}}
\newcommand{\FTRL}{\mathrm{FTRL}}

\title{Minimax-Optimal Small-Loss Bounds for Online\\Linear Optimisation on \(\ell_p\) Balls}
\renewcommand{\shorttitle}{Minimax-Optimal Small-Loss Bounds for Online Linear Optimisation on \(\ell_p\) Balls}
\author{Jack J. Mayo\\
Kurtos.ai\\
Korteweg-de Vries Institute for Mathematics\\
University of Amsterdam\\
\texttt{jack@kurtos.ai}}
\date{}

\begin{document}
\noeqref{eq:slack-boundary,eq:width-profile,eq:quadratic-barrier,eq:stochastic-profile-limit}
\raggedbottom
\setlength{\emergencystretch}{3em}
\maketitle

\begin{abstract}
We study full-information online linear optimisation on the unit
\(\ell_p\) ball in \(\R^d\), for fixed \(2\le p<\infty\) and losses of dual
norm at most one. Let \(\Delta_T\) denote the cumulative loss of the best fixed
action above each round's minimum. A follow-the-regularised-leader (FTRL)
algorithm attains, on every loss sequence and without knowing \(T\) or
\(\Delta_T\), regret
\[
 O_p\!\left(1+\min\left\{\Delta_T^{1-1/p},\
 d^{1/2-1/p}\sqrt{\Delta_T}\right\}\right).
\]
Matching lower bounds show that this rate is minimax for each fixed \(p\), and
its constant \({(2+o(1))p/\log p}\) is optimal as \(p\to\infty\) among
guarantees uniform in \(d\) and \(T\). The algorithm uses a bounded radial
regulariser and switches at most once to its quadratic augmentation. On some
paths where follow-the-leader (FTL) has regret one, its regret grows with
\(T\). A switching extension keeps the sharp constant and also guarantees
\(\Reg_T\le2\Reg_T^{\FTL}\). For i.i.d.\ losses whose mean has no zero
coordinates, we compute the exact \(\log T\) coefficient of the expected
regret, which separates FTL's variance term from the bias of the regulariser.
\end{abstract}


\section{Introduction}
\label{sec:olo-introduction}

We study full-information online linear optimisation (OLO) on the unit
\(\ell_p\) ball in \(\R^d\), for fixed \(2\le p<\infty\). With losses
normalised in the dual norm, the minimax regret over horizon \(T\) is of
order \(\min\{T^{1-1/p},d^{1/2-1/p}\sqrt T\}\); adaptive regularisation
attains this rate without knowing the horizon \citep{johnson2025necessity}.
We ask whether the best comparator's loss can replace the horizon
without losing the transition between these two rates. Linear losses can be negative, so a small-loss bound first subtracts each round's minimum.
Write \(r=p/(p-1)\) and
\(B_p^d=\{x\in\R^d:\norm x_p\le1\}\), and assume
\(\norm{\theta_t}_r\le1\). The minimum loss on round \(t\) is
\(-\norm{\theta_t}_r\), so subtracting it gives the \emph{shifted loss}
\(\norm{\theta_t}_r+\ip{x}{\theta_t}\ge0\).
The best fixed action's cumulative shifted loss is the
\emph{cancellation slack}
\[
 \Delta_T=\sum_{t=1}^T\norm{\theta_t}_r
 -\norm[\Big]{\sum_{t=1}^T\theta_t}_r.
\]
This is the triangle-inequality slack in the support function of
\(B_p^d\), and the linear instance of constraint-aware small loss
\citep[Section~4.3]{gao2026small}. It vanishes on losses aligned along a common ray, regardless
of their number or magnitudes, and satisfies \(0\le\Delta_T\le T\).

Our follow-the-regularised-leader (FTRL) learner satisfies,
on every realised loss path,
\begin{equation}
 \Reg_T\le O_p\!\left(
 1+\min\{\Delta_T^{1-1/p},\
 d^{1/2-1/p}\sqrt{\Delta_T}\}\right),
 \label{eq:intro-rate}
\end{equation}
without knowing the horizon or the slack. Its leading coefficient is
\((2+o(1))p/\log p\) as \(p\to\infty\), uniformly over dimensions and
horizons, and no causal learner improves the constant two. For fixed
\(p\), matching lower bounds make the rate minimax under bounds on the
total loss norm and the slack.

\paragraph{Approach.}
We build on the smoothing viewpoint and regret decomposition of
\citet{abernethy2014smoothing}. In this view, FTRL replaces the support
function underlying follow-the-leader (FTL) by a differentiable convex
potential. The decomposition separates Bregman-divergence stability,
approximation error, and changes of potential. We construct a bounded radial regulariser whose stability term is at
most a multiple of the learner's shifted loss, so the learner's own loss
pays for it. A quadratic augmentation of this regulariser gives a similar bound with a dimension-dependent cap. These two geometric bounds supply the power
and square-root branches of \eqref{eq:intro-rate}, respectively.
The learner, which we call standalone FTRL, tunes from its own
cumulative shifted loss and changes regulariser at most once, from the
radial to the augmented form. The potential may increase at this change,
by at most the approximation error of the pre-switch potential. The sharp leading coefficient requires an additional radius-dependent FTL bound and a minor refinement for small slack.

\paragraph{Fast regimes and the cost of smoothing.}
For bounded i.i.d.\ losses with nonzero mean, our learners adapt
separately to cancellation and to the norm of the mean. At a mean with
no zero coordinates, both FTL and standalone FTRL have at most
logarithmic regret, and the exact coefficient separates FTL's variance
term from the augmentation bias. Smoothing also has a pathwise cost: on
some paths where FTL has regret one, standalone FTRL has regret at least
\(\sqrt T/16\) at \(p=2\), and of order \(\log T\) for \(p>2\), where
quadratic augmentation tilts the action away from the leader. A
switching extension, which alternates between FTL and standalone FTRL,
controls this cost: it preserves the sharp small-loss
coefficient and adds \(\Reg_T\le2\Reg_T^{\FTL}\), without restarts or an
additive comparison term.

\paragraph{Scope and outline.}
We assume fixed finite \(p\ge2\), a known loss-norm bound, and exact
support and action oracles.
Section~\ref{sec:olo-setting} states the guarantees,
Sections~\ref{sec:lp-algorithm} and~\ref{sec:one-clock} construct and
analyse standalone FTRL, and Section~\ref{sec:lp-switching} develops the
switching extension. Section~\ref{sec:p-dependence} proves the sharp
leading constant, and Section~\ref{sec:stochastic-fast} treats stochastic
losses.

\subsection{Related work}
\label{sec:baselines}
\label{sec:olo-related}
Table~\ref{tab:method-comparison} compares worst-case and small-loss
bounds for OLO, and guarantees relative to FTL.

\begin{table}[!htbp]
\caption{Worst-case and small-loss regret for OLO with losses of dual norm at most one. The last row competes with FTL. Rows 2--4 are our anytime reductions of
the cited methods (Appendix~\ref{app:baseline-reductions}). Here \(W_p(n,d)=\min\{n^{1-1/p},d^{1/2-1/p}\sqrt n\}\).}
\label{tab:method-comparison}
\centering
\small
\setlength{\tabcolsep}{4pt}
\renewcommand{\arraystretch}{1.22}
\begin{tabular}{@{}>{\raggedright\arraybackslash}p{0.27\textwidth}>{\raggedright\arraybackslash}p{0.27\textwidth}>{\raggedright\arraybackslash}p{\dimexpr0.46\textwidth-16pt\relax}@{}}
\toprule
Method & Regret upper bound & Decision set; scope and oracles \\
\midrule
Adaptive regularisation \citep{johnson2025necessity}
& \(O(1+W_p(T,d))\)
& \(B_p^d\), \(2<p<\infty\); anytime worst-case bound \\[3pt]
Norm perturbations \citep{abernethy2019privacy}
& \(O(d+\sqrt{d\Delta_T})\)
& Symmetric bodies; exact expected perturbed-leader and support oracles \\[3pt]
MetaGrad \citep{vanerven2021metagrad}
& \(O(d\log T+\sqrt{d\Delta_T\log T})\)
& Symmetric bodies; \(O(d^2\log T)\) time and up to \(\lceil\log_2T\rceil\) quadratic projections per round \\[3pt]
Cone multiplicative weights \citep{canyakmaz2023multiplicative}
& \(O(1+\sqrt{\Delta_T})\)
& \(B_2^d\); explicit actions, restarts on observed shifted loss \\[3pt]
Standalone FTRL, Algorithm~\ref{alg:one-clock} (\textbf{this work})
& \(O_p(1+W_p(\Delta_T,d))\)
& \(B_p^d\), \(2\le p<\infty\); sharp leading coefficient, scalar quadratic oracle \\[3pt]
Switching extension, Algorithm~\ref{alg:lp-hybrid} (\textbf{this work})
& \(O_p(1+W_p(\Delta_T,d))\)
& \(B_p^d\), \(2\le p<\infty\); also \(\Reg_T\le2\Reg_T^{\FTL}\), no restarts \\
\bottomrule
\end{tabular}
\end{table}

\citet[Proposition~1]{levy2019geometries} obtain the low-dimensional
worst-case rate, and \citet{johnson2025necessity} adapt FTRL
regularisation across the dimension crossover to obtain an anytime
worst-case bound. Their fixed-shape obstruction (their Theorems~4.2
and~4.6) covers sign-invariant regularisers with one nonincreasing
learning rate that are separable, within constants of separable, or
strongly convex; ours is not strongly convex until its cumulative shifted
loss reaches a threshold, and its change of regulariser is not one
learning rate on a fixed function. Retuning the nonincreasing learning
rates of their time-triggered crossover cannot give our bound for fixed
\(p>2\): once a prefix of zero losses has brought their learner to its
final quadratic phase, their Proposition~4.1 gives regret at least
\(m/16\) on \(m\le d^{1-2/p}\) rounds with slack at most \(m\), whereas our
bound there is \(O_p(1+m^{1-1/p})\). Optimal regularisers for fixed
symmetric sets \citep{gatmiry2024computing} and offset minimax values
\citep{foster2015adaptive} do not by themselves give the joint dependence
on \(T,d,\Delta_T\) on \(B_p^d\).

Shifting each loss and halving makes the comparator loss \(\Delta_T/2\),
the usual small-loss setting
\citep{cesabianchi2006prediction,cesabianchi2007improved}. The
smoothed-loss bound of \citet[Proposition~4.2]{gao2026small} adds a
squared-gradient term to their constraint-aware loss, and this term grows
on aligned losses. Differential stability of perturbed leaders
\citep{abernethy2019privacy} with the norm mechanism of
\citet{hardt2010geometry} gives row~2 of Table~\ref{tab:method-comparison}
for the mean perturbed leader \citep[see also][Appendix~A.5]{wang2022adaptive},
which is FTRL with a bounded regulariser of cap \(d\); our rate replaces
its additive \(d\) by one and its factor \(\sqrt d\) by
\(\min\{\Delta_T^{1/2-1/p},d^{1/2-1/p}\}\).

Second-order bounds imply small-loss bounds, as for experts
\citep{gaillard2014second,koolen2015second}: MetaGrad
\citep{vanerven2016metagrad,vanerven2021metagrad} bounds the regret
against every \(u\in B_p^d\) by \(O(\sqrt{V_T^u\,d\log T}+d\log T)\), and
\(V_T^u=\sum_t\ip{x_t-u}{\theta_t}^2\le2\Reg_T+4\Delta_T\) at the best
fixed action, which gives row~3. Gradient-adaptive bounds, which tune from
observed losses as in \citet{auer2002adaptive}, scale with
\(\sum_t\norm{\theta_t}_r^2\), and comparator-adaptive bounds also with
the comparator norm
\citep{orabona2018scalefree,cutkosky2018blackbox,mhammedi2020lipschitz,jacobsen2022parameter};
this sum is \(T\) on aligned unit losses, whereas \(\Delta_T=0\).
Gradient-variation bounds
\citep{chiang2012online,jacobsen2022parameter,zhao2026gradient} are
incomparable with \(\Delta_T\). Cone multiplicative weights
\citep{canyakmaz2023multiplicative}, restarted on the observed shifted
loss, give the dimension-free Euclidean bound in row~4, which does not
transfer to \(p>2\).

Set curvature stabilises FTL when cumulative loss vectors grow fast
enough to avoid the origin \citep[Theorem~5]{huang2017following}; see
also \citet[Section~4]{molinaro2023curvature}. On \(B_p^d\), \(p>2\),
\citet[Theorem~3.1]{kerdreux2021projectionfree} give horizon dependence
\(O(T^{(p-2)/(p-1)})\) when \(\norm{\sum_{s\le t}\theta_s}_r\ge\lambda t\) at
every prefix. \citet{kerdreux2021linear} use radial regularisation
with a logarithmic barrier, whereas our profile has finite boundary value
and divergent boundary slope; Follow the Shrunken Leader
\citep[Section~4.1.1]{huang2017following} damps Euclidean FTL radially
with a different rule.

For bounded i.i.d.\ losses with nonzero mean, \citet{huang2017following}
also treat quadratic FTRL on the Euclidean ball.
\citet[Theorems~15 and~17]{tsuchiya2024fast} combine curvature around the
population minimiser with a second-order bound, in a framework that also
covers convex losses and adversarial corruption, and obtain expected
regret \(O(T^{(p-2)/(2(p-1))}(\log T)^{p/(2(p-1))})\) on \(B_p^d\) against the
population optimum. The exponent is necessary in dimension two for regret
against the realised comparator \citep[Theorem~C.1]{bhaskara2020imperfect}.
For fixed \(p>2\), our stochastic bound attains it against the realised
comparator without the logarithmic factor, and keeps the norm of the mean
and the cancellation separate.

Combining FTL with a robust learner is standard
\citep{sani2014exploiting,derooij2014follow,wang2022adaptive}; SMART uses
full-history FTL with a restarted base learner and additive epoch terms
\citep[Theorem~5 and Algorithm~2, full version]{banerjee2024smart}. Our
extension retains \eqref{eq:intro-rate} and \(\Reg_T\le2\Reg_T^{\FTL}\)
without an additive term, following FlipFlop \citep{derooij2014follow}
with FTRL in place of AdaHedge and no resets of the cumulative loss
vector.

\FloatBarrier
\Needspace{12\baselineskip}

\section{Protocol and main results}
\label{sec:olo-setting}

\subsection{Protocol and loss geometry}

Fix \(d\in\mathbb N\), \(2\le p<\infty\), and the dual exponent
\(r=p/(p-1)\in(1,2]\), so that \(\norm{\cdot}_r\) is the dual norm of
\(\norm{\cdot}_p\).
Let \(\mathcal K=B_p^d\). At round \(t\), the learner selects
\(x_t\in \mathcal K\), then observes the full loss vector \(\theta_t\) and incurs
\(\ip{x_t}{\theta_t}\). The regret against the best fixed action is
\[
 \Reg_T=\sum_{t=1}^T\ip{x_t}{\theta_t}
 -\min_{x\in\mathcal K}\sum_{t=1}^T\ip{x}{\theta_t}.
\]
Write \(Y_t=-\sum_{i=1}^t\theta_i\) for the negative cumulative loss and
\(h(y)=\max_{x\in\mathcal K}\ip{x}{y}=\norm y_r\) for the support function.
We use the negative cumulative loss so that regularised leaders are
potential gradients. We fix a subgradient selection \(f:\R^d\to\mathcal K\) by
\[
 f(y)=\nabla h(y)
 =\left(\frac{\operatorname{sgn}(y_i)|y_i|^{r-1}}{h(y)^{r-1}}\right)_{i=1}^d
 \quad (y\ne0),\qquad f(0)=0.
\]
Thus \(f(y)\in\partial h(y)\) is a
support-maximising action: \(\ip{f(y)}{y}=h(y)\). It is unique for
\(y\ne0\); at the origin, \(\partial h(0)=\mathcal K\), and we select zero.
Follow-the-leader (FTL) plays \(x_t=f(Y_{t-1})\), and \(\Reg_T^{\FTL}\)
denotes its regret. Symmetry of \(\mathcal K\) makes the best fixed action's cumulative loss
\(-h(Y_T)\), so
\begin{equation}
 \Reg_T=\sum_{t=1}^T\ip{x_t}{\theta_t}+h(Y_T).
 \label{eq:protocol}
\end{equation}
For \(x\in\mathcal K\) and \(\theta\in\R^d\), symmetry
gives the per-round minimum \(-h(\theta)\). The \emph{shifted loss}
\(\ell^+(x,\theta)=h(\theta)+\ip{x}{\theta}\in[0,2h(\theta)]\)
measures the loss of \(x\) above that minimum.
Write \(H_t=\sum_{i=1}^t h(\theta_i)\) for the cumulative loss norm,
so \(\Delta_t=H_t-h(Y_t)\). For a fixed action \(x\in\mathcal K\), define
its cumulative shifted loss by
\(\Delta_t(x)=\sum_{i=1}^t\ell^+(x,\theta_i)=H_t-\ip{x}{Y_t}\).
The learner's cumulative shifted loss satisfies
\begin{equation}
 S_t=\sum_{i=1}^t\ell^+(x_i,\theta_i)=\Delta_t+\Reg_t.
 \label{eq:shifted}
\end{equation}
For every \(x_\star\in\argmin_{x\in \mathcal K}\ip{x}{\sum_t\theta_t}\),
\eqref{eq:protocol} gives
\begin{equation}
 \Delta_T=\min_{x\in\mathcal K}\Delta_T(x)=\Delta_T(x_\star).
 \label{eq:slack-boundary}
\end{equation}
When \(Y_T\ne0\),
strict convexity gives the unique best action \(x_\star=f(Y_T)\).
The contribution of a nonzero \(\theta_t\) to \(\Delta_T\) vanishes exactly
when \(f(-\theta_t)=x_\star\). The slack thus measures the losses'
disagreement with the best action.

\paragraph{Standing assumptions.}
Throughout, losses satisfy \(h(\theta_t)\le1\) and may be chosen
adaptively. The learners know \(p\) and \(d\), but neither the horizon nor
the slack, and use exact support evaluations and action oracles. Unless
stated otherwise, upper bounds hold pathwise and simultaneously for all
horizons \(T\).

\paragraph{Notation.}
For \(x,y\in\R^d\), \(\ip xy\) is the inner product, \(\norm x_q\) the
\(\ell_q\) norm, \(I\) the identity matrix, and \(\preceq\) the Loewner
order; \(\operatorname{sgn}\) is the sign
with \(\operatorname{sgn}(0)=0\), \(\one\{\cdot\}\) the indicator of an
event, and logarithms are natural. For convex \(G:\R^d\to(-\infty,\infty]\), \(G^*(y)=\sup_x\{\ip xy-G(x)\}\)
is the Fenchel conjugate and \(\partial G\) the subdifferential. For
differentiable convex \(G\), the Bregman
divergence is
\(D_G(a,b)=G(a)-G(b)-\ip{\nabla G(b)}{a-b}\), linearised at its second
argument; for the support function,
\(D_h(a,b)=h(a)-h(b)-\ip{f(b)}{a-b}=h(a)-\ip{f(b)}a\) uses the selection
\(f\), also at \(b=0\). The notation \(O_p\) and
\(\Theta_p\) allows constants depending only on \(p\).

\subsection{Regret guarantees}
For real \(n\ge0\), define the regret profile
\begin{equation}
 W_p(n,d)=\min\{n^{1/r},d^{1/2-1/p}\sqrt n\}.
 \label{eq:width-profile}
\end{equation}
Its branches cross at \(n=d\).
To compare learners, let \(C_{\mathsf L}(p)\) be the least coefficient
with \(\E\Reg_T\le C_{\mathsf L}(p)[1+W_p(\Delta_T,d)]\) for all \(d\),
\(T\), and deterministic loss sequences, where the causal learner
\(\mathsf L\) may randomise and may know \(p\), \(d\), and the horizon;
the expectation is over its randomisation.

\begin{theorem}[Small-loss regret]\label{thm:main} Algorithms~\ref{alg:one-clock} and \ref{alg:lp-hybrid} satisfy, at every horizon,
\begin{equation}
 \Reg_T\le C_{\mathsf L}(p)[1+W_p(\Delta_T,d)],
 \label{eq:explicit}
\end{equation}
with \(C_{\mathsf L}(p)\) finite for every \(p\) and
\(C_{\mathsf L}(p)=(2+o(1))p/\log p\) as \(p\to\infty\), uniformly over all dimensions and horizons. Algorithm~\ref{alg:lp-hybrid} guarantees, in addition, \(\Reg_T\le2\Reg_T^{\FTL}\) at every horizon.
\end{theorem}

Section~\ref{sec:p-dependence} proves the coefficient for both
algorithms and shows that no causal learner improves the leading
constant two. At \(p=2\), where
\(W_2(\Delta_T,d)=\sqrt{\Delta_T}\), the coefficient is at most \(71\) for
Algorithm~\ref{alg:one-clock} and \(159\) for Algorithm~\ref{alg:lp-hybrid},
and it remains bounded as \(p\downarrow2\). On aligned
losses (\(\Delta_T=0\)), Algorithm~\ref{alg:lp-hybrid} never leaves FTL and
\(\Reg_T=\Reg_T^{\FTL}\le1\) (Appendix~\ref{app:single-safe}).

\subsection{Minimax regret}
Fix \(T\ge1\) and \(\bar H,\bar\Delta\ge0\). At round \(t\), an adversary
\(\mathsf A\) chooses \(\theta_t\) using only the history
\((x_s,\theta_s)_{s<t}\) and its own randomness.
Define
\[
 \mathcal A=\{\mathsf A:h(\theta_t)\le1\ (1\le t\le T),\quad
 H_T\le\bar H,\quad \Delta_T\le\bar\Delta\},
\]
where the constraints must hold against every learner and on every path.
Learners \(\mathsf L\) are causal, may randomise, and may know the horizon and the bounds \(\bar H\) and \(\bar\Delta\).
The lower bound uses oblivious losses; the upper bound also
covers adversaries that see the current action.

\begin{proposition}[Minimax regret under upper bounds]\label{prop:minimax}
If \(\bar H\ge1\), then
\begin{equation}
 \inf_{\mathsf L}\sup_{\mathsf A\in\mathcal A}
 \E_{\mathsf L,\mathsf A}[\Reg_T]
 =\Theta_p\bigl(1+W_p(\min\{\bar H,\bar\Delta,T\},d)\bigr).
 \label{eq:minimax}
\end{equation}
If \(\bar H<1\), the value is exactly \(\bar H\).
\end{proposition}

Section~\ref{sec:p-dependence} optimises a single coefficient required to
hold for every dimension and horizon; sharp constants as a joint function
of \(p,d,\bar H,\bar\Delta,T\) remain open. Appendix~\ref{app:minimax} gives the
lower constructions, with independent random signs on coordinates.

\section{Smoothing the support function}
\label{sec:lp-algorithm}
Our construction must control
two costs: stability, measured by a Bregman divergence,
and the cost of approximating \(h\). For the adaptive result, we require the learner's shifted
loss to pay for both, at the rates that give the two branches of \(W_p\). Appendix~\ref{app:lp-proofs} proves the local geometric results used throughout this section.

\subsection{Regularisation and the regret decomposition}
A \emph{bounded regulariser} on \(\mathcal K\) with \emph{cap} \(C\) is a convex
lower semicontinuous map \(F:\R^d\to[0,\infty]\), finite exactly on
\(\mathcal K\), with \(F(0)=0\), \(F\le C\) on \(\mathcal K\), and
\((\beta F)^*\) differentiable for every \(\beta>0\).
With inverse learning rate (or temperature) \(\beta\), its potential is
\[
 \Phi_\beta(y)=(\beta F)^*(y)
 =\max_{x\in\mathcal K}\{\ip{x}{y}-\beta F(x)\}.
\]
The gradient \(\nabla\Phi_\beta(y)\) is the FTRL action and lies in
\(\mathcal K\). Boundedness controls the \emph{support deficit}:
\(0\le h(y)-\Phi_\beta(y)\le C\beta\); also \(\Phi_\beta(0)=0\)
(Lemma~\ref{lem:bounded-regulariser}).

We use the potential decomposition of \citet[Lemma~2]{abernethy2014smoothing};
see also the adaptive FTRL bounds of
\citet[Lemma~5 and Corollary~6]{mcmahan2017survey}.
Let \(\Phi_t\) be round \(t\)'s potential. A regularised action
\(x_t=\nabla\Phi_t(Y_{t-1})\) satisfies
\(\ip{x_t}{\theta_t}=\Phi_t(Y_{t-1})-\Phi_t(Y_t)+D_{\Phi_t}(Y_t,Y_{t-1})\).
If every round is regularised, summing with \(\Phi_1(0)=0\) gives
\begin{equation}
 \Reg_T=\sum_{t=1}^T D_{\Phi_t}(Y_t,Y_{t-1})
 +h(Y_T)-\Phi_T(Y_T)
 +\sum_{t<T}[\Phi_{t+1}(Y_t)-\Phi_t(Y_t)].
 \label{eq:one-clock-telescope}
\end{equation}
The three terms are one-round stability, the terminal support deficit,
and changes of potential. Nondecreasing regularisation weights make
\(\Phi_{t+1}\le\Phi_t\), so the last term is nonpositive except at the
single change of regulariser, which Section~\ref{sec:one-clock} charges
once.
To pay for stability from observed loss, we bound each stability term
by that round's shifted loss; at order two, the bound replaces one factor
\(h(\theta)\) in the usual bound of order \(h(\theta)^2/\beta\) by the
learner's shifted loss.

\begin{definition}[Shifted-loss stability]\label{def:stability}
For \(\bar r\in(1,2]\), a bounded regulariser \(F\) is
\emph{\((\bar r,\kappa)\)-shifted-loss stable} if, for every \(\beta>0\)
and \(y,\theta\in\R^d\), its associated potential \(\Phi_\beta\) satisfies
\begin{equation}
 D_{\Phi_\beta}(y-\theta,y)
 \le\kappa\left(\frac{h(\theta)}\beta\right)^{\bar r-1}\ell^+(\nabla\Phi_\beta(y),\theta).
 \label{eq:stability}
\end{equation}
\end{definition}

\subsection{Radial and quadratic regularisers}
For \(0\le\rho<1\), set
\(g(\rho)=\rho+(1-\rho)\log(1-\rho)\), extended to \(g(1)=1\) by
continuity; this is the generalised relative entropy of \(1-\rho\) with
respect to one \citep[Table~1]{banerjee2005clustering}.
Define on \(\mathcal K\)
\begin{equation}
 F_1(x)=g(\norm{x}_p),\qquad F_2(x)=F_1(x)+\tfrac12\norm{x}_2^2,
 \label{eq:quadratic-barrier}
\end{equation}
with value \(+\infty\) outside \(\mathcal K\). The finite height \(g(1)=1\)
bounds the regularisation cost, so \(F_1\le1\) and
\(F_2\le\bar F_2=1+d^{1-2/p}/2\) on \(\mathcal K\); the divergent boundary slope
\(g'(\rho)=-\log(1-\rho)\) keeps FTRL actions in the interior.
For \(F_1\), the action is an explicit damped leader:
\begin{equation}
 (\beta F_1)^*(y)=h(y)+\beta e^{-h(y)/\beta}-\beta,\qquad
 \nabla(\beta F_1)^*(y)=\bigl(1-e^{-h(y)/\beta}\bigr)f(y).
 \label{eq:families}
\end{equation}
For \(y\ne0\), the radial action lies on the ray through the FTL action
\(f(y)\), with \(\ell_p\) distance \(e^{-h(y)/\beta}\) from the boundary along this
ray. Thus \(h(y)/\beta\) controls how closely the radial action follows
FTL.

Both stability bounds rest on one observation: the shifted loss has
separate terms for radial and angular motion. For either regulariser,
write a nonzero interior action as \(x=\rho u\), where \(0<\rho<1\) and
\(\norm u_p=1\). For a nonzero loss, put \(s=h(\theta)\), \(w=-\theta/s\),
and \(\omega=1-\ip uw\ge0\). The shifted loss at \((x,\theta)\) is then
\[
 \ell^+(x,\theta)=s\bigl[(1-\rho)+\rho\omega\bigr].
\]
The first term measures the distance from the boundary along the action's
ray; the second measures its misalignment with the loss.

\begin{lemma}[Radial regularisation]\label{lem:power}
The regulariser \(F_1\) is bounded with cap \(1\) and
\((r,5/2)\)-shifted-loss stable.
\end{lemma}

For \(F_1\), a composition identity \eqref{eq:power-decomposition} splits
the stability term into a scalar divergence of the radii, at most
\(s\ell^+(x,\theta)/(2\beta)\), and an angular term \(\rho D_h(y-\theta,y)\),
at most \(2(s/\beta)^{r-1}\ell^+(x,\theta)\) when \(s\le\beta\). Since
\(s/\beta\le(s/\beta)^{r-1}\), this gives the constant \(5/2\); for
\(s>\beta\), the bound \(D_{\Phi_\beta}\le\ell^+\) suffices
(Appendix~\ref{app:power}).

\begin{theorem}[Stability of the augmented regulariser]\label{thm:quadratic-log}
The regulariser \(F_2\) is bounded with cap \(\bar F_2\) and
\((2,\bar\kappa_p)\)-shifted-loss stable, for a computable constant
\(1\le\bar\kappa_p<p\) depending only on \(p\), and
\(\bar\kappa_p\sim p/(2\log p)\). Uniformly over \(d\ge2\), the least
constant with this property is \((1+o(1))p/(2\log p)\) as \(p\to\infty\).
\end{theorem}

The dimension enters only through the cap \(\bar F_2\). With \(v=\nabla\norm{x}_p\) and
\(J=\nabla^2F_2(x)\), the quadratic term and \(g''(\rho)=1/(1-\rho)\) give
\(J^{-1}\preceq I\) and \(v^\top J^{-1}v\le1-\rho\): inverse curvature is
bounded, and in the normal direction by the distance from the boundary.
With strong convexity of the squared dual norm for the angular part, this
gives \(w^\top J^{-1}w\le2p[(1-\rho)+\rho\omega]\), and integration along
the loss step gives \((2,p)\)-shifted-loss stability
(Appendix~\ref{app:quad}). A finer boundary estimate gives \(\bar\kappa_p\),
and a two-dimensional witness shows that no smaller constant works
asymptotically (Proposition~\ref{prop:quad-constant-lower}).

\paragraph{Balancing the two costs.}
For a \((\bar r,\kappa)\)-shifted-loss stable regulariser with cap \(C\),
a fixed temperature \(\beta\) gives approximation cost \(C\beta\) and stability cost at most
\(\kappa S_T/\beta^{\bar r-1}\), using \(h(\theta_t)\le1\).
Balancing the costs
\(C\beta+\kappa S_T/\beta^{\bar r-1}\) gives the scale
\(C^{1-1/\bar r}(\kappa S_T)^{1/\bar r}\), up to a factor depending only
on \(\bar r\). For our two regularisers, \(\kappa\) depends only on \(p\).
Up to these \(p\)-dependent factors, Lemma~\ref{lem:power} and
Theorem~\ref{thm:quadratic-log} give
\(S_T^{1/r}\) and \(d^{1/2-1/p}\sqrt{S_T}\), respectively; their minimum is
\(W_p(S_T,d)\).

\section{Tuning by the shifted loss}
\label{sec:one-clock}

Algorithm~\ref{alg:one-clock} tunes a regulariser from \(S_t\), the
cumulative shifted loss of its played actions. It uses \(F_1\) until this
sum crosses a threshold, then uses \(F_2\). The single change of
regulariser costs one bounded potential increase.

\subsection{Regularisation schedule and algorithm}
The fixed-temperature balance suggests a radial temperature of order
\(u^{1/r}\) and an augmented temperature of order
\(\sqrt{u/\bar F_2}\), where \(u\) is the observed cumulative shifted loss.
With the constant \(\bar\kappa_p\) of Theorem~\ref{thm:quadratic-log},
define the crossover and the radial scale by
\begin{equation}
 m_p=\bar\kappa_p^{\min\{1,p-2\}},\qquad
 d_*=d\,\bar\kappa_p^{\,p\min\{1,1/(p-2)\}},\qquad
 \varepsilon_p=\min\{1,1/\log p\},
 \label{eq:calibrated-profile}
\end{equation}
where \(1/(p-2)=\infty\) at \(p=2\). Then \((d_*/d)^{1-2/p}=m_p\), with
\(m_2=1\) and \(m_p=\bar\kappa_p\) for \(p\ge3\). For \(u\ge0\), define
\[
 \beta_2(u)=\sqrt{m_p(u+2)/\bar F_2}\,\one\{u\ge d_*\},
 \qquad
 \beta_1(u)=
 \begin{cases}
  \varepsilon_p(u+2)^{1/r},&u<d_*,\\
  \beta_2(u),&u\ge d_*.
 \end{cases}
\]
For \(b>0\) and \(q\ge0\), write
\(F^{b,q}=bF_1+\tfrac q2\norm{\cdot}_2^2\) and
\(\Phi^{b,q}=(F^{b,q})^*\), so that \(F^{b,0}=bF_1\) and
\(F^{b,b}=bF_2\). The corresponding FTRL action is
\begin{equation}\label{eq:monotone-action}
 x(u,y)=\nabla\Phi^{\beta_1(u),\beta_2(u)}(y).
\end{equation}
Thus the regulariser is \(\varepsilon_p(u+2)^{1/r}F_1\) before crossover
and \(\beta_2(u)F_2\) afterwards. Each phase has one increasing
temperature. The radial weight may decrease at crossover; we charge
this single potential increase below. The offset \(2\) matches the
maximum increment \(\ell^+(x_t,\theta_t)\le2\) and keeps the initial
temperature positive. Algorithm~\ref{alg:one-clock} uses \(u=S_{t-1}\)
on every round and keeps the full cumulative loss vector across the change.

\Needspace{5\baselineskip}
Placing the crossover at \(d_*\ge d\), rather than at the branch crossing
\(d\) of \(W_p(\cdot,d)\), increases the quadratic temperature by \(\sqrt{m_p}\). For \(p\ge3\), the power scale
\(u^{1/r}\) and quadratic scale \(\sqrt{\bar\kappa_p}\,d^{1/2-1/p}\sqrt u\)
meet at \(u=d\,\bar\kappa_p^{p/(p-2)}=d_*\). The definition for \(2\le p<3\)
keeps \(m_p\to1\) as \(p\downarrow2\). The radial scale \(\varepsilon_p\le1\) improves
tracking of FTL.

The upper bounds use only that \(F_2\) is
\((2,\bar\kappa_p)\)-shifted-loss stable, that \(1\le\bar\kappa_p\le p\),
and that \(\bar\kappa_p/\sqrt p\to\infty\). The constant \(p\) of the
elementary argument in Section~\ref{sec:lp-algorithm} also has these properties
and gives the same
upper bounds and leading constants, with a later crossover and a larger
augmentation-bias coefficient (Proposition~\ref{prop:local-stochastic}).

\begin{algorithm}[H]
\begin{algbox}{FTRL tuned by shifted loss}{alg:one-clock}
\State \textbf{Input:} \(p\), \(d\), and gradient oracles; use \(d_*,\varepsilon_p\) from \eqref{eq:calibrated-profile}.
\State Initialise \(Y_0=0\), \(S_0=0\).
\For{\(t=1,2,\ldots\)}
 \State Play \(x_t=x(S_{t-1},Y_{t-1})\) and observe \(\theta_t\).
 \State \(S_t\gets S_{t-1}+\ell^+(x_t,\theta_t)\); \(Y_t\gets Y_{t-1}-\theta_t\).
\EndFor
\end{algbox}
\end{algorithm}

\paragraph{Paying for the crossover.}
Within either phase, increasing the temperature makes the potential
nonincreasing. At crossover, the new potential is at most \(h\) and
the old radial potential has support deficit at most
\(J=\varepsilon_p(d_*+2)^{1/r}\), so the potential increases by at most
\(J\) at any dual point. Once the cumulative shifted loss \(u\) has
crossed \(d_*\),
\[
 J\le3\varepsilon_p d_*^{1/r}\le3\varepsilon_p W_p(u,d_*).
\]
Proposition~\ref{prop:tuning} adds this increase once to the stability
increments and the terminal support deficit, and gives
\[
 \Reg_T\le O_p\bigl(1+W_p(S_T,d_*)\bigr),
\]
with an implied factor bounded as \(p\to\infty\).

\paragraph{From learner loss to comparator loss.}
The bound above contains the regret itself, through
\(S_T=\Delta_T+\Reg_T\). For each fixed \(p\), solving this implicit
inequality gives Theorem~\ref{thm:main} for Algorithm~\ref{alg:one-clock}:
on the square-root branch the inequality is quadratic in
\(\sqrt{\Delta_T+\Reg_T}\), whereas on the power branch subadditivity of
\(u\mapsto u^{1/r}\) gives a coefficient exponential in \(p\)
(Appendix~\ref{app:calibrated-standalone}). The switching extension and
the sharp coefficient instead compare with FTL, whose regret we bound in
the same slack.

\subsection{FTL regret against the slack}
For a step \(y\mapsto y-\theta\), FTL's regret increment is
\(D_h(y-\theta,y)\), and the slack increment is
\(h(y)+h(\theta)-h(y-\theta)\). We compare them using contraction of
Bregman divergences along segments (Lemma~\ref{lem:symmetry-contraction}).
For the power potential \(P(y)=h(y)^r/r\), this contraction follows from
the approximate-symmetry estimate
\(D_P(v,u)\le2(p-1)D_P(u,v)\)
\citep[Theorem~1]{kindermann2019symmetry}. The resulting bound improves
as the cumulative loss norm \(h(y)\) grows relative to \(h(\theta)\), and
summing it after the last step made from within a radius cutoff
\(\bar R>0\) bounds FTL's regret by the slack. Put
\(b_p(\bar R)=1/\{(1+\bar R)^{1/(2p-2)}-1\}\), and define
\[
 C_p^{\FTL}=\max\{2\bar R+2,b_p(\bar R)\}
 \qquad\text{at }\bar R=\frac{p-1}{\log(2p-2+e)}.
\]

\begin{lemma}[Radius-dependent FTL stability]\label{lem:radius-ftl}
For every \(y,\theta\in\R^d\) with \(h(y),h(\theta)>0\),
\[
 D_h(y-\theta,y)
 \le\frac{h(y)+h(\theta)-h(y-\theta)}
 {(1+h(y)/h(\theta))^{1/(2p-2)}-1}.
\]
\end{lemma}

\begin{proposition}[FTL regret against the slack]\label{prop:ftl-log-p}
For every \(\bar R>0\),
\begin{equation}
 \Reg_T^{\FTL}\le2(\bar R+1)+\max\{1,b_p(\bar R)\}\Delta_T.
 \label{eq:ftl-radius-cut}
\end{equation}
In particular, \(\Reg_T^{\FTL}\le C_p^{\FTL}(1+\Delta_T)\) and
\(C_p^{\FTL}\sim2p/\log p\).
\end{proposition}

Appendix~\ref{app:radius-ftl} derives Lemma~\ref{lem:radius-ftl} from
segment contraction and normalised norm increments, and proves
Proposition~\ref{prop:ftl-log-p} from it.
As \(p\) grows, \(B_p^d\) approaches the cube, and the
approximate-symmetry factor \(2(p-1)\) reflects the resulting loss of
curvature: for \(\log(1+\bar R)=o(p)\),
\(b_p(\bar R)\sim2(p-1)/\log(1+\bar R)\), so increasing the radius reduces
the cost per unit of slack only logarithmically.

\section{An extension for comparison with FTL}
\label{sec:lp-switching}

Standalone FTRL already attains the sharp small-loss bound. We now seek
the additional pathwise guarantee \(\Reg_T\le2\Reg_T^{\FTL}\), which
adapts to paths where FTL performs well even when the slack is
large. Proposition~\ref{prop:standalone-separation} shows that standalone
FTRL cannot provide it: radial damping has a cost at
\(p=2\), and quadratic augmentation adds a directional bias at \(p>2\).

\subsection{Standalone costs relative to FTL}

\begin{proposition}[Standalone costs relative to FTL]
\label{prop:standalone-separation}
\begin{enumerate}[label=(\alph*),nosep,leftmargin=*]
\item\label{item:radial-separation}
Let \(p=2\), \(d=1\), let \(T\ge4\) be even, put \(\lambda=1/(8\sqrt T)\),
and let \(\theta_{2k-1}=1\) and \(\theta_{2k}=-(1-2\lambda)\) for
\(k\le T/2\). For \(1\le t\le T\),
\[
 h(Y_t)\ge\lambda t,\qquad\Reg_t^{\FTL}=1,
\]
and Algorithm~\ref{alg:one-clock} satisfies \(\Reg_T\ge\sqrt T/16\).
\item\label{item:nonradial-separation}
Let \(p>2\), \(d=2\), \(\eta=(2,1)/(2^r+1)^{1/r}\), and let
\(\theta_{2k-1}=\eta\) and \(\theta_{2k}=-\eta/2\) for \(k\ge1\).
For every \(t\ge1\),
\[
 h(Y_t)\ge t/4,\qquad\Reg_t^{\FTL}=1,
\]
and Algorithm~\ref{alg:one-clock} satisfies
\[
 \lim_{T\to\infty}\frac{\Reg_T}{\log T}
 =\frac{m_p}{\bar F_2}\,
 \nabla h(\eta)^\top\nabla^2h(\eta)\,\nabla h(\eta)\in(0,\infty).
\]
The right side is \(O((p-2)^2)\) as \(p\downarrow2\).
\end{enumerate}
\end{proposition}

In part~(a), standalone FTRL stays far enough inside the ball to pay at
least \(\lambda\) per pair of rounds.
In part~(b), the quadratic term tilts the action away from the fixed
leader direction. Appendix~\ref{app:ftl-comparison} proves both claims.
These examples rule out a uniform multiplicative FTL comparison for
Algorithm~\ref{alg:one-clock}, despite its sharp small-loss guarantee.

\subsection{Switching rule}

We switch between full-history FTL and the regularised action
\eqref{eq:monotone-action}, accumulating the shifted loss only on FTRL
rounds. The triangle inequality gives
\(\Delta_t-\Delta_{t-1}=h(\theta_t)+h(Y_{t-1})-h(Y_t)\ge0\).
The regret increments satisfy
\begin{equation}
 \Reg_t-\Reg_{t-1}=\ell^+(x_t,\theta_t)-(\Delta_t-\Delta_{t-1}), \qquad
 \Reg_t^{\FTL}-\Reg_{t-1}^{\FTL}=D_h(Y_t,Y_{t-1})\ge0.
 \label{eq:ftl-increments}
\end{equation}
The FTL identity is the support-function Bregman decomposition of
\citet[Proposition~3]{huang2017following}, written for cumulative losses.
Let \(A_t\) and \(Q_t\) sum the learner's regret increments over FTL and
FTRL rounds \(i\le t\), respectively. Then \(\Reg_t=A_t+Q_t\).
On FTL rounds the learner plays FTL's action, so \(A_t\) is
nondecreasing and at most \(\Reg_t^{\FTL}\). The FTRL contribution
\(Q_t\) can decrease;
the tests therefore use its running maximum
\(M_t=\max\{0,Q_1,\ldots,Q_t\}\).
Let \(U_t\) sum the shifted losses on FTRL rounds \(i\le t\).
It stays fixed on FTL rounds; the state \(Y_t\) includes every loss.

The learner starts in FTL. It leaves
FTL when \(A>4(c+M)\) and returns when \(2(c+M)>A\), where
\(c=2^{1/r}\in(1,2)\). Between the
two thresholds it keeps its current mode. Neither switch resets the
accumulated quantities, the regulariser phase, or the cumulative loss
vector. On FTRL rounds the learner plays the action
\eqref{eq:monotone-action} of Algorithm~\ref{alg:one-clock} with \(U\) in
place of \(S\), \(x_t=x(U_{t-1},Y_{t-1})\). As for standalone FTRL, the
change of regulariser happens at most once, since \(U\) never decreases.

\begin{algorithm}[H]
\begin{algbox}{Switching on observed regret}{alg:lp-hybrid}
\State \textbf{Input:} \(p\), \(d\), and gradient oracles.
\State Initialise \(Y_0=0\), \(A_0=Q_0=M_0=U_0=0\), and mode \(\FTL\).
\For{\(t=1,2,\ldots\)}
 \State \((A_t,Q_t,M_t,U_t)\gets(A_{t-1},Q_{t-1},M_{t-1},U_{t-1})\).
 \If{mode is \(\FTL\)}
  \State Play \(x_t=f(Y_{t-1})\) and observe \(\theta_t\).
  \State \(A_t\gets A_{t-1}+h(Y_{t-1}-\theta_t)-h(Y_{t-1})+\ip{x_t}{\theta_t}\).
  \State \textbf{if} \(A_t>4(c+M_t)\) \textbf{then} set mode to \(\FTRL\).
 \Else
  \State Play \(x_t=x(U_{t-1},Y_{t-1})\) and observe \(\theta_t\).
  \State \(Q_t\gets Q_{t-1}+h(Y_{t-1}-\theta_t)-h(Y_{t-1})+\ip{x_t}{\theta_t}\);
         \(M_t\gets\max\{M_{t-1},Q_t\}\).
  \State \(U_t\gets U_{t-1}+h(\theta_t)+\ip{x_t}{\theta_t}\).
  \State \textbf{if} \(2(c+M_t)>A_t\) \textbf{then} set mode to \(\FTL\).
 \EndIf
 \State \(Y_t\gets Y_{t-1}-\theta_t\).
\EndFor
\end{algbox}
\end{algorithm}

\Needspace{10\baselineskip}
\subsection{Regret of the switching extension}

\begin{lemma}[Regret of Algorithm~\ref{alg:lp-hybrid}]\label{lem:single-safe}
Algorithm~\ref{alg:lp-hybrid} satisfies:
\begin{enumerate}[label=(\alph*),nosep,leftmargin=*]
\item\label{item:switch-comparison}
\(\Reg_T\le2\Reg_T^{\FTL}\);
\item\label{item:switch-counter}
\(U_T\le\Delta_T+\Reg_T^{\FTL}/2+1\);
\item\label{item:switch-rate}
\(\Reg_T\le155\,[1+W_p(U_T,d_*)]\)
(Appendix~\ref{app:switch-constants}).
\end{enumerate}
\end{lemma}

The switching tests control regret relative to FTL, and a bound for
each regularised block preserves the small-loss bound. Immediately before
every FTRL round, \(2(c+M)\le A\), and each FTRL round raises \(Q\) by at
most two, so \(M_T\le A_T/2+2-c\) at every horizon. Since
\(A_T\le\Reg_T^{\FTL}\), slack increments are nonnegative, and
\(A_T>4c\) after the first entry into FTRL, this gives
parts~\ref{item:switch-comparison} and~\ref{item:switch-counter}. For
part~\ref{item:switch-rate}, call a maximal run of consecutive FTRL rounds a
\emph{block}. By the telescoping bound \eqref{eq:potential-block-budget},
the regret contribution of a block is at most the sum of the accumulated
stability, the possible crossover charge, and the terminal support
deficit, and Proposition~\ref{prop:tuning} charges the crossover once
across all blocks. Lemma~\ref{lem:switch-accounting} combines the block
bounds using the switching tests, which follow FlipFlop
\citep[Section~3.2]{derooij2014follow}: \(A\) sums FTL's mixability gaps,
and \(c+M\) replaces the cumulative gap on AdaHedge rounds. Each completed
FTRL block more than doubles \(c+M\), so the last completed block and any
unfinished block control the final running maximum.
Appendix~\ref{app:single-safe} gives the proofs.

\paragraph{The slack rate without self-bounding.}
Combining part~\ref{item:switch-counter} with the FTL slack bound of
Proposition~\ref{prop:ftl-log-p} gives \(U_T\le N_p(1+\Delta_T)\), where
\(N_p=1+C_p^{\FTL}/2=O(p/\log p)\). This bounds the FTRL rounds'
shifted loss by quantities independent of the learner's own regret.
Rescaling the profile by
\(N_p\) costs at most a factor \(\max\{N_p^{1/r},\sqrt{m_pN_p}\}\), and
part~\ref{item:switch-rate} then gives the slack bound in Theorem~\ref{thm:main} for the
switching extension, with a finite coefficient depending only on \(p\)
(Appendix~\ref{app:switch-constants}). The sharper analysis in
Appendix~\ref{app:calibrated-standalone} gives \((2+o(1))p/\log p\).

On the paths of Proposition~\ref{prop:standalone-separation},
\(\Reg_t^{\FTL}=1\) at every horizon, so part~\ref{item:switch-comparison}
gives \(\Reg_t\le2\) for the switching extension,
whereas standalone FTRL has regret at least \(\sqrt T/16\) or of order
\(\log T\). The known loss-norm bound is essential here: over losses of
arbitrary norms, Proposition~\ref{prop:unknown-width} rules out small-loss
adaptation together with a uniform FTL comparison.

\section{Sharp dependence on \texorpdfstring{\(p\)}{p}}
\label{sec:p-dependence}

We now determine the coefficient in Theorem~\ref{thm:main} as
\(p\to\infty\), uniformly over dimensions and horizons. A mean-zero loss
construction gives the lower bound; FTL's slack bound and radial tracking
give the matching upper bound for both algorithms.

\subsection{A universal lower bound}

Recall \(C_{\mathsf L}(p)\) from Section~\ref{sec:olo-setting}.
Its \emph{leading constant} is
\(\liminf_{p\to\infty}(\log p/p)C_{\mathsf L}(p)\).
This constant concerns guarantees uniform in \(d\). At each fixed \(d\), the
norm-perturbation bound of Proposition~\ref{prop:general-prior} gives
\(\Reg_T\le O(d)[1+W_p(\Delta_T,d)]\) with an implied constant
independent of \(p\), since \(W_p(\Delta_T,d)\ge\sqrt{\Delta_T}\) for
\(\Delta_T\ge1\) and the additive term absorbs smaller slack. The lower
bound below therefore concerns the uniform coefficient, and
fixed-dimension constants remain open.

\begin{theorem}[Universal leading constant]\label{thm:universal-lower}
\(\inf_{\mathsf L}C_{\mathsf L}(p)\ge[2-O(1/\log p)]\,p/\log p\), the infimum
being over all causal learners. Hence no learner has leading constant
below two, and, with Theorem~\ref{thm:main},
\(\lim_{p\to\infty}(\log p/p)\inf_{\mathsf L}C_{\mathsf L}(p)=2\).
\end{theorem}

Near the cube, losses can nearly share a minimiser without becoming
predictable: the lower construction has \(T+1\) mean-zero unit losses with
\(T=\lceil2p\rceil\), in dimension \(O(p^6)\), but expected slack only
\(\log p+O(1)\). It costs about \(\log p/p\) in slack to reverse a
coordinate near the cube. Appendix~\ref{app:power-dependence} turns this
into the matching coefficient.
The construction lies on the power branch; a matching leading constant
for the square-root branch alone remains open.

\subsection{Matching upper bounds}

Before the first quadratic action, Lemma~\ref{lem:radial-tracking}
bounds the cost of the radial action relative to FTL, so the shifted loss
that either algorithm uses for tuning is at most
\((3+\Reg_T^{\FTL}+2\Delta_T)/(1-\varepsilon_p)\). For sufficiently large
\(p\) and slack at most \(e^{\sqrt p}\), a coordinatewise potential bounds
FTL's regret by \(4+2d+2\sqrt p\,\Delta_T\)
(Corollary~\ref{cor:ftl-low-dim}). Together, these bounds keep both
algorithms radial on the small-slack power branch, where the leading
constant is determined; there, \(\varepsilon_p\to0\) removes the damping
cost, and both algorithms inherit FTL's leading coefficient two. In the
remaining regimes, the same bounds force
\(d\le O(\log p/\sqrt p)\Delta_T\) at the first crossover, which makes the
additive cost of the implicit quadratic bound lower order.
Appendix~\ref{app:calibrated-standalone} gives the details, uniformly in
dimension and horizon.

\section{Cancellation and fast stochastic rates}
\label{sec:stochastic-fast}

For i.i.d.\ losses, we bound the expected regret against the best realised
comparator, keeping cancellation and the norm of the mean separate, and
identify the exact cost of quadratic smoothing.

The losses \(\theta_t\) are i.i.d.\ with
\(h(\theta_t)\le1\) and nonzero mean. Write
\[
 \mu=\E\theta_1,\qquad \gamma=\E h(\theta_1)-h(\mu).
\]
The population minimiser is \(-\nabla h(\mu)\). The time profile is the generalised
harmonic sum
\[
 \mathcal H_r(T)=\sum_{n=1}^{T-1}n^{-r/2},\qquad \mathcal H_r(1)=0.
\]

\subsection{Adaptation to stochastic signal}

\begin{theorem}[Stochastic adaptation]\label{thm:stochastic-signal}
Under the preceding assumptions, both Algorithms~\ref{alg:one-clock} and~\ref{alg:lp-hybrid} satisfy
\begin{equation}
 \E\Reg_T\le A_p\min\left\{
 1+W_p(\gamma T,d),\
 \frac{d^{1-2/p}}{h(\mu)}
 +\frac{d^{1-r/2}\gamma^{r/2}}{h(\mu)^{r-1}}\mathcal H_r(T)
 \right\}
 \label{eq:stochastic-cancellation-rate}
\end{equation}
against the best fixed action in hindsight. The constant \(A_p\) depends
only on \(p\); it is bounded as \(p\downarrow2\) but grows with \(p\).
\end{theorem}

The algorithms need neither \(h(\mu)\), \(\gamma\), nor the horizon.
For \(\bar\theta_T=T^{-1}\sum_{t\le T}\theta_t\), cancellation has the
Bregman-information representation \citep[Section~3.1]{banerjee2005clustering}
\[
 \Delta_T=\sum_{t\le T}D_h(\theta_t,\bar\theta_T),\qquad
 \gamma=\E D_h(\theta_1,\mu).
\]
The exact sample decomposition gives
\begin{equation}
 T\gamma-\E\Delta_T
 =T\E D_h(\bar\theta_T,\mu)
 =\E h\left(\sum_{t=1}^T\theta_t\right)-T h(\mu).
 \label{eq:stochastic-bregman-information}
\end{equation}
The nonnegative Jensen gap \(T\E D_h(\bar\theta_T,\mu)\) is the
hindsight correction: the expected cost of comparing with the best
realised action rather than the population minimiser. Expected regret
is this correction plus the pseudo-regret
\(\sum_{t\le T}\E[h(\mu)+\ip\mu{x_t}]\)
by \eqref{eq:stochastic-hindsight-decomposition}.
The first branch of \eqref{eq:stochastic-cancellation-rate} follows from
Theorem~\ref{thm:main} and \(\E\Delta_T\le\gamma T\); the second bounds the
pseudo-regret and the hindsight correction separately
(Appendix~\ref{app:stochastic}).

With \(\alpha=1-r/2=(p-2)/(2(p-1))\), integral comparison gives
\begin{equation}
 \mathcal H_r(T)=
 \begin{cases}
 (T^\alpha-1)/\alpha+O(1),&p>2,\\
 \log T+O(1),&p=2,
 \end{cases}
 \label{eq:stochastic-profile-limit}
\end{equation}
where the remainder lies in \([0,1]\). For fixed \(p>2\), this yields
horizon dependence \(O_p(T^\alpha)\); as \(p\downarrow2\),
\((T^\alpha-1)/\alpha\to\log T\), recovering
\(O(\min\{1+\sqrt{\gamma T},(1+\gamma\log(eT))/h(\mu)\})\).
Sharp joint dependence on \(d\), \(h(\mu)\), and \(\gamma\) remains open,
including the dimension dependence of the additive term.
For \(p>2\), the horizon exponent is necessary in dimension two
\citep[Theorem~C.1]{bhaskara2020imperfect}.

\subsection{Local curvature and the cost of smoothing}
\label{sec:augmentation-geometry}

The support function is smooth near a mean with no zero coordinates
(and near every nonzero mean at \(p=2\)), giving logarithmic rates;
local curvature also underlies \citet{tsuchiya2024fast}.
\begin{proposition}[Local stochastic asymptotics]\label{prop:local-stochastic}
Under the assumptions of Theorem~\ref{thm:stochastic-signal}, suppose
\(p=2\) or every coordinate of \(\mu\) is nonzero. Write
\(\Sigma=\E[(\theta_1-\mu)(\theta_1-\mu)^\top]\).
Then \(\E\Reg_T^{\FTL}/\log T\to\tfrac12\operatorname{tr}(\nabla^2h(\mu)\Sigma)\),
and Algorithm~\ref{alg:one-clock} satisfies
\begin{equation}
 \lim_{T\to\infty}
 \frac{\E\Reg_T}{\log T}
 =\frac12\operatorname{tr}(\nabla^2h(\mu)\Sigma)
  +\frac{m_p\gamma}{2\bar F_2}\nabla h(\mu)^\top\nabla^2h(\mu)\nabla h(\mu).
 \label{eq:stochastic-local-rate}
\end{equation}
\end{proposition}

For expected regret divided by \(\log T\), FTL has the variance coefficient
and standalone FTRL adds smoothing bias. By Theorem~\ref{thm:main},
Algorithm~\ref{alg:lp-hybrid} instead satisfies
\(\limsup_{T\to\infty}\E\Reg_T/\log T\le\operatorname{tr}(\nabla^2h(\mu)\Sigma)\),
independently of the augmentation-bias term.

Since \(\nabla^2h(\mu)\mu=0\), only transverse error contributes.
For standalone FTRL, the action \(x_{n+1}\) is the radial action at the
effective dual point \(Y_n-\beta_2(S_n)x_{n+1}\)
(Lemma~\ref{lem:effective-dual-point}). At scale \(n^{-1/2}\), this point
displaces the empirical mean by \(-\sqrt{m_p\gamma/\bar F_2}\,\nabla h(\mu)\)
\eqref{eq:stochastic-local-shift}. At \(p=2\), \(\nabla h(\mu)\parallel\mu\), so
this displacement has zero curvature cost. For \(p>2\), it can tilt the
leader and add regret even without transverse statistical fluctuations.
The bias term vanishes exactly when \(\gamma=0\) or
\(\nabla h(\mu)\parallel\mu\). Appendix~\ref{app:persistent-gap} proves the
regret limits.

\section{Discussion}
\label{sec:olo-open}
\label{sec:computation}
We have provided two algorithms for small-loss online linear
optimisation on \(\ell_p\) balls, \(2\le p<\infty\), whose regret adapts
to the cancellation slack. The first, standalone FTRL, changes once from
a bounded radial regulariser to its quadratic augmentation. On every loss
sequence, and without knowing the horizon or the slack, it attains the
minimax rate for each fixed \(p\), with leading coefficient
\((2+o(1))p/\log p\) uniformly over dimensions and horizons; we have
shown that no learner improves the constant two. Because standalone FTRL
can pay a cost growing with the horizon on paths where FTL has regret
one, we have also given a switching extension, which keeps the sharp
coefficient and has at most twice the regret of FTL. For i.i.d.\ losses,
both algorithms adapt separately to cancellation and to the norm of
the mean. At a mean with no zero coordinates, we have computed the exact
logarithmic coefficient of standalone FTRL, which adds an augmentation
bias to FTL's variance term; the switching extension's coefficient is at
most twice FTL's, with no bias term.

Our sharp coefficient describes the limit \(p\to\infty\), and the upper
bounds we prove approach it slowly. With the radius cutoff
\(\bar R=p/\log^2(e+p)\) used in our proofs, the small-slack coefficient
is at least \(b_p(\bar R)\), which exceeds \(2p/\log p\) by the factor
\(1+(2+o(1))\log\log p/\log p\). Attaining the
sharp worst-case coefficient also does not remove the cost of smoothing
relative to FTL: the quadratic scale \(\sqrt{m_p}\) of our tuning enters
the stochastic analysis, where its square multiplies the
augmentation-bias coefficient.

Both algorithms need only explicit formulas and scalar root-finding. The
radial action is explicit, and Appendix~\ref{app:oracle-errors} finds
the quadratic action by solving one scalar equation, with a separate
coordinate equation at each evaluation. We have not analysed inexact
actions: a finite-precision guarantee would need error bounds for the
actions and support evaluations, rigorous stopping rules for the root
solver, and control of errors in the switching tests. The bit complexity
remains open, including when actions lie exponentially close to the
boundary.

It remains open whether one fixed regulariser shape with an adaptive
scalar temperature can attain the anytime worst-case rate for \(p>2\),
beyond the classes excluded by \citet{johnson2025necessity}. We also do
not know a direct FTRL tuning rule that combines the sharp small-loss
bound with a uniform comparison to FTL; our separations apply only to
the present standalone learner. Tuning a single FTRL potential from accumulated
mixability gaps, as in AdaHedge \citep{derooij2014follow}, may help
control the cost of smoothing, but preserving both small-loss branches
without persistent bias requires a new analysis.

Beyond \(\ell_p\) balls, we do not know whether computable bounded
regularisers with shifted-loss stability improve the small-loss bounds
known for general symmetric convex bodies. Our argument uses the strict
convexity of \(B_p^d\); on polytopes, losses can share a minimising
vertex without being aligned, and the leader can jump between vertices.
Our analysis also uses full information. Under bandit feedback, scalar
observations do not directly reveal the per-round shift
\(h(\theta_t)\). Conditionally unbiased loss estimates \(\hat\theta_t\)
give an unbiased estimate of the cumulative loss vector, but not of the
shift. Since \(h\) is convex, the plug-in shift \(h(\hat\theta_t)\)
exceeds \(h(\theta_t)\) in conditional expectation by the expected
divergence \(D_h(\hat\theta_t,\theta_t)\), which depends on the
dispersion of the estimates rather than on the slack. Adapting our tuning
to bandit feedback therefore requires either an estimate of the shift
whose cumulative bias is at most of order \(1+\Delta_T\), or a tuning rule
that does not use the shift.

\section*{AI disclosure}
OpenAI's Codex and Anthropic's Claude assisted with proof auditing,
revisions, experiments, numerical verification, literature checks, and
editing, including code and suggested mathematical arguments. The author
has verified all mathematical arguments by hand, line by line, has checked
all other content, and bears full responsibility for the correctness,
completeness, and novelty of the content.

\bibliography{triangle_gap_olo}

\appendix
\counterwithin{theorem}{section}
\counterwithin{equation}{section}
\section*{Guide to the appendices}

\paragraph{Additional notation.}
The appendices also write \(e_1,\ldots,e_d\) for the standard basis of
\(\R^d\), \(\interior(\mathcal K)\) for the interior of \(\mathcal K\), and
\(\iota_{\mathcal K}\) for its indicator, zero on \(\mathcal K\) and
\(+\infty\) outside, so that \(\iota_{\mathcal K}^*=h\).
\begin{center}
\small
\begin{tabular}{@{}>{\raggedright\arraybackslash}p{0.2\textwidth}>{\raggedright\arraybackslash}p{\dimexpr0.62\textwidth-4\tabcolsep\relax}>{\raggedright\arraybackslash}p{0.18\textwidth}@{}}
\toprule
Symbol & Meaning & Where \\
\midrule
\(G^*,D_G\) & conjugate, Bregman divergence & \S\ref{sec:olo-setting} \\
\(h,f\) & support function \(\norm\cdot_r\), its gradient selection & \S\ref{sec:olo-setting} \\
\(Y_t,H_t,\Delta_t\) & negative cumulative loss, cumulative loss norm, slack & \S\ref{sec:olo-setting} \\
\(\ell^+,S_t,U_t\) & shifted loss; its sum over all rounds, over FTRL rounds & \S\ref{sec:olo-setting}, \S\ref{sec:lp-switching} \\
\(W_p(n,d)\) & \(\min\{n^{1/r},d^{1/2-1/p}\sqrt n\}\) & \eqref{eq:width-profile} \\
\(g,F_1,F_2,\bar F_2\) & radial profile, regularisers, cap of \(F_2\) & \S\ref{sec:lp-algorithm} \\
\(F^{b,q},\Phi^{b,q}\) & augmented regulariser and potential & \S\ref{sec:one-clock} \\
\(\bar\kappa_p,m_p,d_*,\varepsilon_p\) & quadratic stability constant and tuning & \eqref{eq:calibrated-profile} \\
\(c,A_t,Q_t,M_t\) & \(2^{1/r}\); accumulated regrets and running maximum & \S\ref{sec:lp-switching} \\
\(\Upsilon_0,\Upsilon,J_*,\mathcal C\) & stability coefficients, crossover charge, and cap & App.~\ref{app:one-clock} \\
\(b_p(\bar R),C_p^{\FTL}\) & FTL slack coefficients & \S\ref{sec:one-clock} \\
\(C_{\mathsf L}(p)\) & least uniform coefficient of a learner & \S\ref{sec:olo-setting} \\
\(\mu,\gamma\) & mean, expected cancellation & \S\ref{sec:stochastic-fast} \\
\(\Pi_a\) & projection transverse to \(\nabla h(a)\), along \(a\) & Lemma~\ref{lem:support-curvature} \\
\bottomrule
\end{tabular}
\end{center}

For the sharp upper bounds, follow Lemma~\ref{lem:scalar-potential}
and Corollary~\ref{cor:ftl-low-dim} to Appendix~\ref{app:calibrated-standalone},
which treats both algorithms.
For the lower bound, Proposition~\ref{prop:universal-finite} combines
mean-zero losses with Jensen's inequality; a horizon of order \(p\)
gives Theorem~\ref{thm:universal-lower}.

\section{Geometry and stability}
\label{app:lp-proofs}

We establish the local estimates used in the regret analysis:
bounded regularisation, support-function curvature, radial and quadratic
stability, and the effective dual point.

\subsection{Bounded regularisers}

The approximation estimate below is the bounded-penalty estimate used
in max-function smoothing \citep[Section~2]{nesterov2005smooth}.
Here differentiability of the conjugate is part of the definition.

\begin{lemma}[Bounded regularisers]\label{lem:bounded-regulariser}
Let \(F\) be a bounded regulariser on compact set \(\mathcal K\) with cap \(C\),
and let \(\Phi_\beta=(\beta F)^*\). Then, for every \(\beta>0\) and
\(y\in\R^d\),
\[
 \nabla\Phi_\beta(y)\in \mathcal K,\qquad
 \Phi_\beta(0)=0,\qquad
 0\le h(y)-\Phi_\beta(y)\le C\beta.
\]
\end{lemma}

\begin{proof}
Fix \(\beta>0\) and \(y\in\R^d\). Lower semicontinuity of \(F\) on
compact \(\mathcal K\) gives a maximiser \(x_y\in\mathcal K\) of
\(x\mapsto\ip{x}{y}-\beta F(x)\). For every \(z\in\R^d\),
\[
 \Phi_\beta(z)\ge\ip{x_y}{z}-\beta F(x_y)=\Phi_\beta(y)+\ip{x_y}{z-y},
\]
so \(x_y\in\partial\Phi_\beta(y)\), and differentiability gives
\(\nabla\Phi_\beta(y)=x_y\in\mathcal K\). Since
\(h(y)=\max_{x\in\mathcal K}\ip xy\), the bounds \(0\le F\le C\) on
\(\mathcal K\) give the support deficit, and \(F\ge0=F(0)\) gives
\(\Phi_\beta(0)=0\).
\end{proof}

\subsection{Transverse geometry and local curvature}
\label{app:support-curvature}

Homogeneity separates displacement along a ray from changes of direction.
The exact normalisation below holds in every nonzero direction; the
quadratic expansion additionally requires local smoothness.

\begin{lemma}[Support geometry]\label{lem:support-curvature}
Let \(a\ne0\) and \(\Pi_a=I-a\nabla h(a)^\top/h(a)\).
\begin{enumerate}[label=(\alph*),nosep,leftmargin=*]
\item The map \(\Pi_a\) projects onto \(\{z:\ip{\nabla h(a)}z=0\}\) along \(a\),
and \(h(\Pi_a z)\le2h(z)\). For \(A=\ip{\nabla h(a)}z>0\),
\begin{equation}
 \begin{aligned}
 D_h(z,a)&=A D_h\left(\frac zA,\frac a{h(a)}\right),\\
 D_h(a,z)&=h(a)D_h\left(\frac a{h(a)},\frac zA\right).
 \end{aligned}
 \label{eq:support-normalisation}
\end{equation}
\item If \(p=2\) or all coordinates of \(a\) are nonzero, then
\begin{equation}
 \nabla^2h(a)=\frac{r-1}{h(a)}
 \left\{\operatorname{diag}\!\left(\left|\frac{a_i}{h(a)}\right|^{r-2}\right)
       -\nabla h(a)\nabla h(a)^\top\right\},
 \label{eq:norm-local-curvature}
\end{equation}
where \(0^0=1\) at \(p=2\). The Hessian is positive semidefinite
with \(\ker\nabla^2h(a)=\operatorname{span}\{a\}\).
Consequently, \(\nabla h(a)^\top\nabla^2h(a)\nabla h(a)=0\) at \(p=2\); for \(p>2\), it
vanishes exactly when all \(|a_i|\) are equal.
For \(c>0\), \(\nabla^2h(ca)=c^{-1}\nabla^2h(a)\),
\(\nabla^2h(-a)=\nabla^2h(a)\), and
\begin{equation}
 \nabla^2h(a)=\Pi_a^\top\nabla^2h(a)\Pi_a.
 \label{eq:transverse-curvature}
\end{equation}
As \(u\to0\),
\begin{equation}
 \begin{aligned}
 D_h(a+u,a)&=\tfrac12u^\top\nabla^2h(a)u+O(\norm u_2^3),\\
 D_h(a,a+u)&=\tfrac12u^\top\nabla^2h(a)u+O(\norm u_2^3),
 \end{aligned}
 \label{eq:support-taylor}
\end{equation}
with remainders locally uniform in \(a\).
\end{enumerate}
\end{lemma}

\begin{proof}
\emph{Projection.} Since \(\ip{\nabla h(a)}a=h(a)\), \(\Pi_a\) annihilates
\(a\), fixes every \(z\) with \(\ip{\nabla h(a)}z=0\), and maps into
that hyperplane. For every \(z\),
\begin{align*}
 h(\Pi_a z)
 &\le h(z)+|\ip{\nabla h(a)}z|
 \tag{triangle inequality}\\
 &\le2h(z).
 \tag{H\"older, \(\norm{\nabla h(a)}_p=1\)}
\end{align*}

\emph{Normalisation.} Let \(A=\ip{\nabla h(a)}z>0\). Then
\(\ip{\nabla h(a)}{z/A}=1\) and \(z/A=a/h(a)+\Pi_a z/A\).
The gradient \(\nabla h\) is invariant under positive scaling, and
\(D_h(b',b)=h(b')-\ip{\nabla h(b)}{b'}\) for \(b\ne0\). Hence
\begin{align*}
 D_h(z,a)
 &=h(z)-\ip{\nabla h(a)}z\\
 &=A\Bigl\{h\Bigl(\frac zA\Bigr)
   -\Bigl\langle\nabla h\Bigl(\frac a{h(a)}\Bigr),\frac zA\Bigr\rangle\Bigr\}
 \tag{homogeneity}\\
 &=A\,D_h\Bigl(\frac zA,\frac a{h(a)}\Bigr),\\
 D_h(a,z)
 &=h(a)-\ip{\nabla h(z)}a\\
 &=h(a)\Bigl\{h\Bigl(\frac a{h(a)}\Bigr)
   -\Bigl\langle\nabla h\Bigl(\frac zA\Bigr),\frac a{h(a)}\Bigr\rangle\Bigr\}
 \tag{homogeneity}\\
 &=h(a)\,D_h\Bigl(\frac a{h(a)},\frac zA\Bigr).
\end{align*}
This proves \eqref{eq:support-normalisation}.

\emph{Local curvature.} Differentiating
\(\nabla h(a)_i=\operatorname{sgn}(a_i)|a_i/h(a)|^{r-1}\) gives
\eqref{eq:norm-local-curvature}. For every \(z\),
\begin{align*}
 (\nabla h(a)^\top z)^2
 &\le\Bigl(\sum_i\Bigl|\frac{a_i}{h(a)}\Bigr|^{r-1}|z_i|\Bigr)^2\\
 &\le\Bigl(\sum_i\Bigl|\frac{a_i}{h(a)}\Bigr|^r\Bigr)
       \Bigl(\sum_i\Bigl|\frac{a_i}{h(a)}\Bigr|^{r-2}z_i^2\Bigr)
 \tag{Cauchy--Schwarz}\\
 &=\sum_i\Bigl|\frac{a_i}{h(a)}\Bigr|^{r-2}z_i^2,
 \tag{\(\norm a_r=h(a)\)}
\end{align*}
with equality exactly when \(z\) is parallel to \(a\).
By \eqref{eq:norm-local-curvature}, \(z^\top\nabla^2h(a)z\) is
\((r-1)/h(a)\) times the difference of the two sides, so
\(\nabla^2h(a)\succeq0\) with kernel \(\operatorname{span}\{a\}\).
Hence \(\nabla h(a)^\top\nabla^2h(a)\nabla h(a)=0\) exactly when
\(\nabla h(a)\) is parallel to \(a\). At \(p=2\), \(\nabla h(a)=a/\norm a_2\).
For \(p>2\), parallelism is equivalent to the ratios
\(|a_i|^{r-2}/h(a)^{r-1}\) being equal, hence to equal \(|a_i|\).
Homogeneity and evenness give the Hessian scaling and reflection identities.
Also
\begin{align*}
 \Pi_a^\top\nabla^2h(a)\Pi_a
 &=\Pi_a^\top\nabla^2h(a)
 \tag{\(\nabla^2h(a)a=0\)}\\
 &=\nabla^2h(a),
 \tag{symmetry}
\end{align*}
which is \eqref{eq:transverse-curvature}.

\emph{Expansions.} On a neighbourhood of a point satisfying
the assumptions, \(h\) has
bounded third derivatives. Taylor's theorem gives the first expansion in
\eqref{eq:support-taylor}; expanding both \(h(a+u)\) and \(\nabla h(a+u)\)
in \(D_h(a,a+u)\) gives the second. The derivative bounds are uniform
on a sufficiently small compact neighbourhood, proving the stated uniformity.
\end{proof}

\subsection{Radial stability}
\label{app:power}

The proof separates changes in the norm from changes in direction.
The intermediate lemmas are stated for general norms where the argument
permits this.

\subsubsection{Power estimates and normalised increments}

For \(1<r\le2\), write \(\varphi(a)=|a|^r/r\) on \(\R\) and
\(P(y)=\sum_{i=1}^d\varphi(y_i)\) on \(\R^d\).

\begin{lemma}[Scalar power bounds and dilation]\label{lem:scalar-power}
Let \(1<r\le2\). For all \(a,b,w\in\R\), the following hold:
\begin{enumerate}[label=(\alph*),ref=(\alph*),nosep,leftmargin=*]
\item\label{item:scalar-power-derivative}
\(|\varphi'(a)-\varphi'(b)|\le2^{2-r}|a-b|^{r-1}\).
\item\label{item:scalar-power-divergence}
\(D_\varphi(a+w,a)\le2^{2-r}|w|^r/r\).
\item\label{item:scalar-power-dilation}
For every \(t\in[0,1]\),
\begin{equation}
 D_\varphi(a+tw,a)\le 2^{2-r}t^rD_\varphi(a+w,a).
 \label{eq:scalar-power-comparison}
\end{equation}
\end{enumerate}
For every \(d\ge1\), both divergence bounds extend by summation to
\(P\) on \(\R^d\), with
\(|w|^r\) replaced by \(\norm{w}_r^r\).
\end{lemma}

\begin{proof}
The vector bounds follow by summing the scalar ones. At \(r=2\), all
three scalar estimates are equalities, so assume \(1<r<2\).

\emph{Power bounds.}
We have \(\varphi'(a)=\operatorname{sign}(a)|a|^{r-1}\). If \(ab\ge0\),
subadditivity gives
\(|\varphi'(a)-\varphi'(b)|\le|a-b|^{r-1}\).
If \(ab<0\), then \(|a-b|=|a|+|b|\). Since \(0<r-1<1\), the map
\(s\mapsto s^{r-1}\) is concave on \([0,\infty)\), so its midpoint
inequality gives
\begin{align*}
 |\varphi'(a)-\varphi'(b)|
 &=|a|^{r-1}+|b|^{r-1}\\
 &\le2\left(\frac{|a|+|b|}{2}\right)^{r-1}
 =2^{2-r}|a-b|^{r-1}.
\end{align*}
This proves part~\ref{item:scalar-power-derivative}. By the fundamental theorem of calculus,
we can integrate along the segment from \(a\) to \(a+w\) and apply this bound:
\begin{align*}
 D_\varphi(a+w,a)
 &=\varphi(a+w)-\varphi(a)-\varphi'(a)w\\
 &=\int_0^1\{\varphi'(a+\sigma w)-\varphi'(a)\}w\dif\sigma\\
 &\le |w|\int_0^1|\varphi'(a+\sigma w)-\varphi'(a)|\dif\sigma\\
 &\le 2^{2-r}|w|^r\int_0^1\sigma^{r-1}\dif\sigma
 \tag{from part~\ref{item:scalar-power-derivative}}\\
 &=\frac{2^{2-r}}r|w|^r.
\end{align*}

\emph{Dilation, part~\ref{item:scalar-power-dilation}.}
The cases \(t=0\) and \(w=0\) are immediate. Otherwise, let
\(q=-a/w\) and \(I_r(q)=D_\varphi(1-q,-q)\).
The homogeneity identity
\(D_\varphi(\lambda u,\lambda v)=|\lambda|^rD_\varphi(u,v)\)
gives
\begin{align*}
 D_\varphi(a+w,a)&=|w|^rI_r(q),\\
 D_\varphi(a+tw,a)&=t^r|w|^rI_r(q/t).
 \tag{homogeneity}
\end{align*}
Thus it suffices to prove \(I_r(q/t)\le2^{2-r}I_r(q)\).
Since \(\varphi''(s)=(r-1)|s|^{r-2}\) for \(s\ne0\), Taylor's
integral formula yields
\[
 I_r(q)=(r-1)\int_0^1(1-v)|v-q|^{r-2}\dif v.
\]
This remains valid when the segment crosses zero because \(r-2>-1\),
so the singularity is integrable.

\emph{Outside \([0,1]\).}
If \(q\le0\), then \(q/t\le q\le v\) for every \(v\in[0,1]\).
If \(q\ge1\), then \(q/t\ge q\ge v\).
In either case, \(|v-q/t|\ge|v-q|\), so the integral formula gives
\[
 I_r(q/t)\le I_r(q)\le2^{2-r}I_r(q).
 \tag{\(r-2<0\)}
\]
It remains to consider \(0<q<1\).

\emph{Shape of \(I_r\) on \((0,1)\).}
Splitting the integral at \(v=q\) and evaluating the two pieces gives
\[
 I_r(q)=\left(q^{r-1}-\frac{q^r}{r}\right)+\frac{(1-q)^r}{r}.
\]
Define \(J_r(q)=q+q^{2-r}(1-q)^{r-1}\). Differentiation gives
\[
 I_r'(q)=(r-1)q^{r-2}-q^{r-1}-(1-q)^{r-1}
 =q^{r-2}\{(r-1)-J_r(q)\}.
\]
For \(0<q\le1/2\),
\begin{align*}
 J_r'(q)
 &=1+(2-r)\left(\frac{1-q}{q}\right)^{r-1}
   -(r-1)\left(\frac q{1-q}\right)^{2-r}\\
 &\ge1-(r-1)=2-r>0.
 \tag{\(q/(1-q)\le1\)}
\end{align*}
Since \(J_r(0+)=0\), \(J_r(1/2)=1\), and \(0<r-1<1\), there is a
unique \(\hat q\in(0,1/2)\) with \(J_r(\hat q)=r-1\).
For \(1/2\le q<1\),
\[
 J_r(q)=q+(1-q)\left(\frac q{1-q}\right)^{2-r}
 \ge q+(1-q)=1>r-1.
 \tag{\(q/(1-q)\ge1\)}
\]
Thus \(I_r\) increases up to \(\hat q\) and decreases from \(\hat q\)
to \(1\). It also decreases for \(q\ge1\): increasing \(q\) increases
\(|v-q|\) in the integral formula. By continuity, \(I_r\) is
nonincreasing on \([\hat q,\infty)\).

\emph{Conclusion.}
If \(q\ge\hat q\), then \(q/t\ge q\ge\hat q\), so
\[
 I_r(q/t)\le I_r(q)\le2^{2-r}I_r(q).
 \tag{monotonicity}
\]
If \(0<q<\hat q\), then \(I_r(q)\ge I_r(0)=1/r\).
Applying part~\ref{item:scalar-power-divergence} with base point \(-q/t\)
and increment \(1\) gives
\begin{align*}
 I_r(q/t)
 &=D_\varphi(1-q/t,-q/t)\le\frac{2^{2-r}}r
 \tag{from part~\ref{item:scalar-power-divergence}}\\
 &\le2^{2-r}I_r(q).
 \tag{\(I_r(q)\ge1/r\)}
\end{align*}
The homogeneity identities now give \eqref{eq:scalar-power-comparison}.
\end{proof}

\paragraph{Normalising a norm increment.}
Let \(h\) be a norm on \(\R^d\), let \(q\ge1\), and write \(P=h^q/q\).
Fix \(y,\theta\in\R^d\) with \(R=h(y)>0\) and \(s=h(\theta)>0\), and a
selection \(a\in\partial h(y)\). Normalise the two directions as
\(u=y/R\) and \(v=-\theta/s\), and put \(t=s/(R+s)\) and
\(m=(1-t)u+tv\), so that \(y-\theta=(R+s)m\). The divergences \(D_h(\cdot,y)\),
\(D_h(\cdot,u)\), and \(D_P(\cdot,u)\) are linearised with \(a\). Write
\(\tau=D_h(y-\theta,y)\), \(\delta=R+s-h(y-\theta)\), and
\(z_f=s+\ip a\theta\); when \(a=f(y)\), these are FTL's regret increment, the slack increment,
and FTL's shifted loss on the step from \(y\) to \(y-\theta\).

\begin{lemma}[Normalisation of norm increments]\label{lem:norm-increment}
In this setting, \(a\in\partial h(u)\cap\partial P(u)\), and the
following hold:
\begin{enumerate}[label=(\alph*),ref=(\alph*),nosep,leftmargin=*]
\item\label{item:norm-increment-regret}
\(\tau=(R+s)D_h(m,u)\le(R+s)D_P(m,u)\);
\item\label{item:norm-increment-loss}
\(\delta+\tau=z_f=sD_h(v,u)=sD_P(v,u)\);
\item\label{item:norm-increment-contraction}
if \(D_P(m,u)\le Ct^\gamma D_P(v,u)\) for some \(C\ge0\) and
\(\gamma>0\), then \(\tau\le Ct^{\gamma-1}z_f\).
\end{enumerate}
\end{lemma}

\begin{proof}
Positive homogeneity of \(h\) gives \(\partial h(y)=\partial h(u)\) and
\(\ip au=h(u)=1\). The chain rule gives
\(\partial P(u)=h(u)^{q-1}\partial h(u)=\partial h(u)\), so
\(a\in\partial P(u)\). Hence, for every \(b\in\R^d\),
\[
 D_h(b,u)=h(b)-\ip ab,\qquad
 D_P(b,u)-D_h(b,u)=\frac{h(b)^q-1}{q}-\{h(b)-1\}\ge0,
\]
where the inequality is the tangent bound at \(1\) for the convex map
\(x\mapsto x^q/q\).

\emph{Part~\ref{item:norm-increment-regret}.}
\begin{align*}
 \tau
 &=h(y-\theta)-\ip a{y-\theta}
 \tag{\(\ip ay=h(y)\)}\\
 &=(R+s)\{h(m)-\ip am\}
 \tag{\(y-\theta=(R+s)m\)}\\
 &=(R+s)D_h(m,u)
 \tag{\(D_h(m,u)=h(m)-\ip am\)}\\
 &\le(R+s)D_P(m,u).
 \tag{\(D_P\ge D_h\)}
\end{align*}

\emph{Part~\ref{item:norm-increment-loss}.}
The definitions of \(\delta\) and \(\tau\) give
\begin{align*}
 \delta+\tau
 &=R+s-\ip a{y-\theta}\\
 &=s+\ip a\theta
 \tag{\(\ip ay=R\)}\\
 &=s(1-\ip av)
 \tag{\(\theta=-sv\)}\\
 &=sD_h(v,u)
 \tag{\(h(v)=1\)}\\
 &=sD_P(v,u),
 \tag{\(h(v)=1\)}
\end{align*}
and the second line is \(z_f\).

\emph{Part~\ref{item:norm-increment-contraction}.}
If \(D_P(m,u)\le Ct^\gamma D_P(v,u)\), then
\begin{align*}
 \tau
 &\le(R+s)D_P(m,u)
 \tag{part~\ref{item:norm-increment-regret}}\\
 &\le C(R+s)t^\gamma D_P(v,u)
 \tag{hypothesis}\\
 &=Ct^{\gamma-1}z_f.
 \tag{\((R+s)t=s\), part~\ref{item:norm-increment-loss}}
\end{align*}
\end{proof}

\subsubsection{Radial decomposition and control}

For a radial potential \(\Phi=\psi\circ h\), the next identity separates
the scalar divergence of the radii from the divergence of the norm.
It is the composition-remainder identity used by
\citet[proof of Proposition~3.8]{sprung2019bounds}.
The divergences \(D_h(\cdot,b)\) and \(D_\Phi(\cdot,b)\) use a selected
\(f(b)\in\partial h(b)\) and \(\psi'(h(b))f(b)\), respectively.

\begin{lemma}[Bregman composition identity]\label{lem:radial-decoupling}
Let \(h\colon\R^d\to[0,\infty)\) be finite and convex, and let
\(\psi\colon[0,\infty)\to\R\) be convex, nondecreasing, and continuously
differentiable. For every \(a,b\in\R^d\) and selection
\(f(b)\in\partial h(b)\), we have
\(\psi'(h(b))f(b)\in\partial\Phi(b)\) and
\begin{equation}
 D_\Phi(a,b)=D_\psi(h(a),h(b))+\psi'(h(b))D_h(a,b).
 \label{eq:power-decomposition}
\end{equation}
If, in addition, \(h\) is a norm differentiable away from zero and
\(\psi'(0)=0\), then \(\Phi\) is continuously differentiable, with
\(\nabla\Phi(b)=\psi'(h(b))f(b)\) and \(\nabla\Phi(0)=0\).
\end{lemma}

\begin{proof}
For every \(a\in\R^d\),
\begin{align*}
 \Phi(a)
 &\ge\psi(h(b))+\psi'(h(b))\{h(a)-h(b)\}
 \tag{convexity of \(\psi\)}\\
 &\ge\Phi(b)+\psi'(h(b))\ip{f(b)}{a-b},
 \tag{\(\psi'\ge0\), \(f(b)\in\partial h(b)\)}
\end{align*}
so \(\psi'(h(b))f(b)\in\partial\Phi(b)\). With this subgradient,
\begin{align*}
 D_\Phi(a,b)
 &=\psi(h(a))-\psi(h(b))-\psi'(h(b))\ip{f(b)}{a-b}\\
 &=D_\psi(h(a),h(b))+\psi'(h(b))\{h(a)-h(b)-\ip{f(b)}{a-b}\}\\
 &=D_\psi(h(a),h(b))+\psi'(h(b))D_h(a,b).
\end{align*}

For a norm \(h\) differentiable away from zero, the chain rule gives the
stated gradient there, and it is continuous. At zero, \(\psi'(0)=0\)
gives \(\Phi(y)-\Phi(0)=o(h(y))\), so \(\nabla\Phi(0)=0\) by equivalence
of norms. Since norm subgradients are bounded and \(\psi'(h(y))\to0\),
the gradient is continuous at zero as well.
\end{proof}

\paragraph{Controlling the radial term.}
For the radial application, let \(h\) be a norm and select
\(f(y)\in\partial h(y)\), with \(f(0)=0\). For a continuously
differentiable profile \(\psi\) and \(\Phi=\psi\circ h\), write
\(R=h(y)\), \(R'=h(y-\theta)\), and \(s=h(\theta)\).
The action is \(x=\alpha f(y)\), with radial scale \(\alpha=\psi'(R)\).
Its shifted loss is \(z=s+\ip{x}{\theta}\), and FTL's is
\(z_f=s+\ip{f(y)}{\theta}\). These satisfy
\(z=(1-\alpha)s+\alpha z_f\). The exponential profile is
\[
 \psi_\beta(R)=R+\beta e^{-R/\beta}-\beta,\qquad \beta>0.
\]

\begin{lemma}[Radial control]\label{lem:radial-control}
Let \(h\) be any norm on \(\R^d\), and let
\(\psi\colon[0,\infty)\to\R\) be convex and continuously differentiable
with \(0\le\psi'\le1\). For all \(y,\theta\in\R^d\),
\(0\le z\le2s\) and \(D_\Phi(y-\theta,y)\le z\).
For \(\psi=\psi_\beta\) and every \(\beta>0\), we also have
\begin{equation}
 D_{\psi_\beta}(R',R)\le\frac{s z}{2\beta}
 \qquad\text{for all }y,\theta\in\R^d.
 \label{eq:radial-control}
\end{equation}
\end{lemma}

\begin{proof}
\emph{General profile.}
Since \(f(y)\in\partial h(y)\), H\"older's inequality gives
\(|\ip{f(y)}{\theta}|\le s\), so \(z_f\in[0,2s]\). With \(\alpha\in[0,1]\),
\[
 z=s+\alpha\ip{f(y)}{\theta}=(1-\alpha)s+\alpha z_f\in[0,2s].
\]
With the subgradient \(x\) of Lemma~\ref{lem:radial-decoupling},
\begin{align*}
 D_\Phi(y-\theta,y)
 &=\psi(R')-\psi(R)+\ip x\theta\\
 &\le|R'-R|+\ip x\theta
 \tag{\(0\le\psi'\le1\)}\\
 &\le s+\ip x\theta=z.
 \tag{reverse triangle inequality}
\end{align*}

\emph{Exponential profile.}
Now take \(\psi=\psi_\beta\). Then
\(\alpha=\psi_\beta'(R)=1-e^{-R/\beta}\), so
\(z\ge(1-\alpha)s=e^{-R/\beta}s\). Also \(|R'-R|\le s\), and
\(\psi_\beta''(\rho)=e^{-\rho/\beta}/\beta\) is decreasing and at most
\(1/\beta\). If \(R'\ge R\), Taylor's theorem gives
\begin{align*}
 D_{\psi_\beta}(R',R)
 &\le\tfrac12\psi_\beta''(R)(R'-R)^2
 \tag{\(\psi_\beta''\) decreasing}\\
 &\le\frac{e^{-R/\beta}s^2}{2\beta}
 \tag{\(|R'-R|\le s\)}\\
 &\le\frac{s z}{2\beta}.
 \tag{\(z\ge e^{-R/\beta}s\)}
\end{align*}
If \(R'<R\), then
\(0\le D_h(y-\theta,y)=R'-R+\ip{f(y)}{\theta}\) gives
\(\ip{f(y)}{\theta}\ge R-R'>0\), so \(z\ge s\) and
\begin{align*}
 D_{\psi_\beta}(R',R)
 &\le\frac{s^2}{2\beta}
 \tag{\(\psi_\beta''\le1/\beta\), \(|R'-R|\le s\)}\\
 &\le\frac{s z}{2\beta}.
 \tag{\(z\ge s\)}
\end{align*}
\end{proof}

\subsubsection{Proof of radial stability}

\begin{proof}[Proof of Lemma~\ref{lem:power}]
Write
\(\Phi_\beta=(\beta F_1)^*\), which is \(\Phi^{\beta,0}\).
We compute the conjugate, control its angular term, and combine this
with the radial estimate above.

\emph{Conjugacy and regularisation cost.}
Let \(R=h(y)\). Since \(g'(\rho)=-\log(1-\rho)\) is increasing, the
scalar problem below has the unique maximiser \(\rho=1-e^{-R}\), and
\begin{align*}
 F_1^*(y)
 &=\sup_{0\le\rho\le1}\{\rho R-g(\rho)\}
 \tag{\(\max_{\norm x_p=\rho}\ip xy=\rho R\)}\\
 &=R-\rho=R+e^{-R}-1.
 \tag{\(\rho=1-e^{-R}\)}
\end{align*}
Scaling gives \(\Phi_\beta(y)=\beta F_1^*(y/\beta)=\psi_\beta(h(y))\).
Lemma~\ref{lem:radial-decoupling} gives continuous differentiability and
\(\nabla\Phi_\beta(y)=(1-e^{-h(y)/\beta})f(y)\), including at zero.
Since \(g\) is convex and nondecreasing on \([0,1]\), \(F_1\) is convex
and lower semicontinuous, with \(F_1(0)=0\) and
\(0\le F_1\le g(1)=1\) on \(\mathcal K\). Thus \(F_1\) is a bounded
regulariser with cap \(1\).

\emph{Angular control.}
Use the notation of Lemma~\ref{lem:norm-increment} with \(q=r\) and
\(a=f(y)\); then \(z_f\in[0,2s]\).
For \(P=h^r/r=\sum_i\varphi(\cdot_i)\), Lemma~\ref{lem:scalar-power}
gives the hypothesis of Lemma~\ref{lem:norm-increment}\ref{item:norm-increment-contraction}
with \(C=2^{2-r}\) and \(\gamma=r\), so
\begin{equation}
 \tau\le2^{2-r}\left(\frac{s}{R+s}\right)^{r-1}z_f.
 \label{eq:power-angular}
\end{equation}

\emph{Shifted-loss stability.}
Let \(R'=h(y-\theta)\), \(\alpha=1-e^{-R/\beta}\),
\(x=\nabla\Phi_\beta(y)=\alpha f(y)\), and \(z=s+\ip{x}{\theta}\).
By \eqref{eq:stability}, we must show
\(\beta^{r-1}D_{\Phi_\beta}(y-\theta,y)\le\tfrac52s^{r-1}z\).
A zero loss makes both sides vanish. If \(s>\beta\),
Lemma~\ref{lem:radial-control} gives
\(\beta^{r-1}D_{\Phi_\beta}(y-\theta,y)\le\beta^{r-1}z\le s^{r-1}z\).
Suppose now \(0<s\le\beta\). By \eqref{eq:power-decomposition},
\[
 D_{\Phi_\beta}(y-\theta,y)=D_{\psi_\beta}(R',R)+\alpha\tau.
\]
For the radial term,
\begin{align*}
 \beta^{r-1}D_{\psi_\beta}(R',R)
 &\le\tfrac12\beta^{r-2}s z
 \tag{Lemma~\ref{lem:radial-control}}\\
 &\le\tfrac12s^{r-1}z.
 \tag{\((s/\beta)^{2-r}\le1\)}
\end{align*}
For the angular term, \(\alpha\tau=0\) at \(R=0\). For \(R>0\), we have
\(z\ge(1-\alpha)s>0\), \(z\ge(1+\alpha)z_f/2\) since \(z_f\le2s\), and
\(\alpha\le\min\{R/\beta,1\}\). Hence
\begin{align*}
 \frac{\beta^{r-1}\alpha\tau}{s^{r-1}z}
 &\le2^{2-r}\Bigl(\frac\beta R\Bigr)^{r-1}\frac{\alpha z_f}{z}
 \tag{by \eqref{eq:power-angular}, \(R+s\ge R\)}\\
 &\le2^{2-r}\Bigl(\frac\beta R\Bigr)^{r-1}\frac{2\alpha}{1+\alpha}
 \tag{\(z\ge(1+\alpha)z_f/2\)}\\
 &\le2^{2-r}\Bigl(\frac\beta R\Bigr)^{r-1}
       \min\Bigl\{\frac{2R}\beta,1\Bigr\}
 \tag{\(\alpha\le\min\{R/\beta,1\}\)}\\
 &\le2^{2-r}2^{r-1}=2.
 \tag{\(\min\{v,1\}\le v^{r-1}\)}
\end{align*}
Adding the two terms gives
\(\beta^{r-1}D_{\Phi_\beta}(y-\theta,y)\le(\tfrac12+2)s^{r-1}z\).
Thus \(F_1\) is \((r,5/2)\)-shifted-loss stable.
\end{proof}

\subsection{Quadratic stability}
\label{app:quad}

We prove Theorem~\ref{thm:quadratic-log}. Lemma~\ref{lem:quad-reduction}
reduces shifted-loss stability to a bound on inverse curvature. Strong
convexity of the squared dual norm gives such a bound with a constant
below \(p\), and scalar estimates near the boundary improve the constant
to \(\bar\kappa_p\). A two-dimensional witness shows that no smaller
constant works asymptotically.

\subsubsection{Reduction to a curvature bound}
Write \(\Phi_\beta=(\beta F_2)^*\). For a nonzero
\(x\in\interior(\mathcal K)\) and \(w\in\R^d\) with \(h(w)=1\), write
\(x=\rho u\) and \(\omega=1-\ip uw\) as in
Section~\ref{sec:lp-algorithm}, and put
\[
 \eta=1-\rho,\qquad v=\nabla\norm{x}_p,\qquad J=\nabla^2F_2(x).
\]
Then \(h(v)=1\) and \(\omega=D_h(w,v)\). For a loss \(\theta\) with
\(s=h(\theta)>0\) and \(w=-\theta/s\),
\(\ell^+(x,\theta)=s(\eta+\rho\omega)\). We call
\(w^\top J^{-1}w/(\eta+\rho\omega)\) the curvature ratio.

\begin{lemma}[Reduction to a curvature bound]\label{lem:quad-reduction}
The regulariser \(F_2\) is bounded with cap \(\bar F_2\). If \(\kappa>0\)
and \(w^\top J^{-1}w\le2\kappa(\eta+\rho\omega)\) for every nonzero
\(x\in\interior(\mathcal K)\) and every \(w\) with \(h(w)=1\), then
\(F_2\) is \((2,\kappa)\)-shifted-loss stable.
\end{lemma}

\begin{proof}
\emph{Bounded regulariser.} The profile \(g\) is continuous on \([0,1]\), takes values in \([0,1]\),
and satisfies \(g'(\rho)=-\log(1-\rho)\ge0\) and
\(g''(\rho)=(1-\rho)^{-1}>0\) for \(0\le\rho<1\).
Thus \(F_2\) is nonnegative, lower semicontinuous, and Euclidean
1-strongly convex on its compact convex domain. Its conjugate has a
unique maximiser and a globally \(1/\beta\)-Lipschitz gradient
\citep[Theorem~3]{kakade2012regularization};
Lemma~\ref{lem:effective-dual-point}, with \(b=q=\beta\), places the
maximiser in the interior. Since \(F_2(0)=0\) and
\(\norm{x}_2\le d^{1/2-1/p}\) on \(\mathcal K\), its cap is
\(\bar F_2=1+d^{1-2/p}/2\).

\emph{Integration.} Since \(g(\rho)=O(\rho^2)\),
\(\nabla F_2(0)=0\), so the action vanishes exactly at \(y=0\).
Away from zero, \(F_2\) is twice continuously differentiable with positive
definite Hessian; the inverse function theorem gives
\(\beta\nabla^2\Phi_\beta(y)=J^{-1}\).
Fix \(y\) and a loss \(\theta\) with \(s=h(\theta)>0\), and put
\(x=\nabla\Phi_\beta(y)\) and \(w=-\theta/s\).
Along the segment \(y-t\theta\), the derivative of
\(q(t)=\Phi_\beta(y-t\theta)\) is Lipschitz and hence absolutely continuous,
and \(z(t)=s-q'(t)=\ell^+(\nabla\Phi_\beta(y-t\theta),\theta)\ge0\).
Off the possible crossing of zero, with \(J,\eta,\rho,\omega\)
evaluated at the action \(\nabla\Phi_\beta(y-t\theta)\),
\begin{align*}
 \beta q''(t)
 &=\theta^\top J^{-1}\theta
 \tag{\(\beta\nabla^2\Phi_\beta=J^{-1}\)}\\
 &=s^2w^\top J^{-1}w\\
 &\le2\kappa s^2(\eta+\rho\omega)=2\kappa s\,z(t).
 \tag{hypothesis}
\end{align*}
Convexity makes \(q'\) nondecreasing, so \(0\le z(t)\le z(0)\). Hence
\begin{align*}
 \beta D_{\Phi_\beta}(y-\theta,y)
 &=\beta\int_0^1(1-t)q''(t)\dif t
 \tag{\(q'\) absolutely continuous}\\
 &\le2\kappa s\int_0^1(1-t)z(t)\dif t\\
 &\le\kappa s\,\ell^+(x,\theta).
 \tag{\(z(t)\le z(0)\)}
\end{align*}
The case \(\theta=0\) is immediate.
\end{proof}

\subsubsection{Interior estimate}
For \(1<r\le2\), write \(h=\norm{\cdot}_r\) on \(\R^d\) and
\(G=h^2/2\). The first claim below is the \((r-1)\)-strong convexity of
\(G\) with respect to \(h\); see \citet{ball1994sharp} and
\citet[Lemma~9]{kakade2012regularization}.

\begin{lemma}[Strong convexity of the squared \(\ell_r\) norm]\label{lem:dual-convexity}
For every \(1<r\le2\) and all \(v,v'\in\R^d\),
\(D_G(v',v)\ge\tfrac{r-1}{2}h(v'-v)^2\).
In particular, if \(h(v)=h(v')=1\), then
\begin{equation}
 D_h(v',v)\ge\frac{r-1}{2}h(v'-v)^2
 \ge\frac{r-1}{2}\norm{v'-v}_2^2.
 \label{eq:dual-convexity}
\end{equation}
\end{lemma}

\begin{proof}
The first claim restates the cited strong convexity for the
differentiable function \(G\). On the unit sphere,
\(D_G(v',v)=D_h(v',v)\); also \(h\ge\norm{\cdot}_2\) because \(r\le2\).
\end{proof}

Fix a nonzero \(x\in\interior(\mathcal K)\) and \(w\) with \(h(w)=1\).
The chain rule gives
\begin{equation}
 J=I+\frac{vv^\top}{1-\rho}
       -\log(1-\rho)\nabla^2\norm{x}_p
 \succeq I+\frac{vv^\top}{1-\rho}.
 \label{eq:hessian}
\end{equation}
Hence \(J^{-1}\preceq I\). With \(\xi=J^{-1}v\),
\begin{align*}
 \eta^{-1}(v^\top J^{-1}v)^2
 &=\eta^{-1}(v^\top\xi)^2\\
 &\le\xi^\top J\xi
 \tag{\(J\succeq vv^\top/\eta\)}\\
 &=v^\top J^{-1}v,
\end{align*}
so \(v^\top J^{-1}v\le\eta\). Therefore
\begin{align*}
 \sqrt{w^\top J^{-1}w}
 &\le\sqrt{(w-v)^\top J^{-1}(w-v)}+\sqrt{v^\top J^{-1}v}
 \tag{triangle inequality}\\
 &\le\norm{w-v}_2+\sqrt\eta
 \tag{\(J^{-1}\preceq I\)}\\
 &\le\sqrt{2(p-1)\omega}+\sqrt\eta.
 \tag{Lemma~\ref{lem:dual-convexity}}
\end{align*}
Squaring and applying Cauchy--Schwarz with weights \(2(p-1)/\rho\) and \(1\) gives
\[
 w^\top J^{-1}w
 \le\left(1+\frac{2(p-1)}\rho\right)(\eta+\rho\omega).
\]
Also \(w^\top J^{-1}w\le\norm w_2^2\le h(w)^2=1\), so
\begin{equation}
 \frac{w^\top J^{-1}w}{\eta+\rho\omega}
 \le\min\left\{\frac1\eta,\,1+\frac{2(p-1)}\rho\right\}
 \le2\widetilde\kappa_p,
 \qquad \widetilde\kappa_p=\frac{p+\sqrt{p^2-1}}2.
 \label{eq:uniform-quad-curvature}
\end{equation}
The two bounds in the minimum are respectively increasing and decreasing
in \(\rho\), and meet at \(\rho=1-p+\sqrt{p^2-1}\), where both equal
\(2\widetilde\kappa_p\). With Lemma~\ref{lem:quad-reduction}, this
makes \(F_2\) \((2,\widetilde\kappa_p)\)-shifted-loss stable, and
\(\widetilde\kappa_p<p\).

\subsubsection{Scalar and matrix estimates}
Put \(\nu=1/(p-1)=r-1\), and let
\(\varphi(t)=|t|^r/r=|t|^{1+\nu}/(1+\nu)\) be the scalar power of
Appendix~\ref{app:power}. We use the convention \(0^0=1\).

\Needspace{13\baselineskip}
The scalar weight is
\[
 \varsigma_r(t)=\frac{D_\varphi(1,t)}{(1-t)^2}\quad(0\le t<1),
 \qquad \varsigma_r(1)=\frac\nu2,
\]
where the endpoint value is the continuous extension.

\begin{lemma}[Weighted scalar Bregman cost]\label{lem:weighted-scalar}
Let \(1<r\le2\). The function \(\varsigma_r\) is nonincreasing on \([0,1]\)
(constant, equal to \(1/2\), when \(r=2\)).
For all \(v,w\in[-1,1]\),
\(D_\varphi(w,v)\ge \varsigma_r(|v|)(w-v)^2\ge\frac\nu2(w-v)^2\).
Consequently, for all \(v,w\in[-1,1]\),
\(b\ge0\), and \(t\in(0,1)\),
\begin{equation}
 \frac{(w-v)^2}{1+b|v|^{2-r}}
 \le\max\left\{\frac1{\varsigma_r(t)},\,
          \frac{2}{\nu(1+bt^{2-r})}\right\}D_\varphi(w,v).
 \label{eq:scalar-weighted}
\end{equation}
\end{lemma}

\begin{proof}
The case \(v=w\) is immediate. Otherwise, Taylor's integral formula
and \(|v+\tau(w-v)|\le(1-\tau)|v|+\tau\) give
\begin{align*}
 \frac{D_\varphi(w,v)}{(w-v)^2}
 &=\nu\int_0^1(1-\tau)|v+\tau(w-v)|^{\nu-1}\dif\tau
 \tag{\(\varphi''(t)=\nu|t|^{\nu-1}\)}\\
 &\ge\nu\int_0^1(1-\tau)((1-\tau)|v|+\tau)^{\nu-1}\dif\tau
 \tag{\(\nu\le1\), \(|w|\le1\)}\\
 &=\varsigma_r(|v|).
 \tag{definition of \(\varsigma_r\)}
\end{align*}
The nonpositive exponent \(\nu-1\) makes the last integral nonincreasing in \(|v|\),
and its value at one is \(\nu/2\). The possible singularity at a zero
of the affine segment is integrable because \(\nu>0\).
For \eqref{eq:scalar-weighted}, if \(|v|\le t\), then
\begin{align*}
 \frac{(w-v)^2}{1+b|v|^{2-r}}
 &\le(w-v)^2\\
 &\le\frac{D_\varphi(w,v)}{\varsigma_r(t)}.
 \tag{\(\varsigma_r(|v|)\ge\varsigma_r(t)\)}
\end{align*}
If \(|v|\ge t\), then
\begin{align*}
 \frac{(w-v)^2}{1+b|v|^{2-r}}
 &\le\frac{(w-v)^2}{1+bt^{2-r}}\\
 &\le\frac{2D_\varphi(w,v)}{\nu(1+bt^{2-r})}.
 \tag{\(\varsigma_r(|v|)\ge\nu/2\)}
\end{align*}
\end{proof}

For the normalised vectors \(u,v\), write
\[
 \Pi=\operatorname{diag}(|u_i|^{p-2})
   =\operatorname{diag}(|v_i|^{2-r}),\qquad
 b=\frac{(p-1)\log(1/\eta)}{\rho},\qquad \bar J=I+b\Pi,
\]
so that \(\Pi=I\) when \(p=2\). The Hessian has the rank-one decomposition
\begin{equation}
 J=\bar J+(\eta^{-1}-b)vv^\top
 =I+b(\Pi-vv^\top)+\eta^{-1}vv^\top.
 \label{eq:log-full-hessian}
\end{equation}

\begin{lemma}[Diagonal comparison]\label{lem:diagonal-comparison}
For every \(\zeta\in\R^d\),
\(\zeta^\top J^{-1}\zeta\le\zeta^\top\bar J^{-1}\zeta+\eta h(\zeta)^2\).
\end{lemma}

\begin{proof}
If \(\eta^{-1}\ge b\), then \(J\succeq\bar J\), which proves more.
Suppose \(\eta^{-1}<b\), and put \(c=b-\eta^{-1}\in(0,b)\), so that
\(J=\bar J-cvv^\top\). Since \(v_i=0\) whenever \(\Pi_{ii}=0\), and
\(|v_i|=|u_i|^{p-1}\),
\begin{align*}
 v^\top\bar J^{-1}v
 &\le\sum_{i:v_i\ne0}\frac{v_i^2}{b\Pi_{ii}}
 \tag{\(\bar J_{ii}\ge b\Pi_{ii}\)}\\
 &=\frac1b\sum_i|u_i|^p=\frac1b.
\end{align*}
Hence \(1-cv^\top\bar J^{-1}v\ge1-c/b=\eta^{-1}/b\), and
\begin{align*}
 \zeta^\top J^{-1}\zeta
 &=\zeta^\top\bar J^{-1}\zeta
 +\frac{c\,(v^\top\bar J^{-1}\zeta)^2}{1-c\,v^\top\bar J^{-1}v}
 \tag{Sherman--Morrison}\\
 &\le\zeta^\top\bar J^{-1}\zeta+\eta\,(b\,v^\top\bar J^{-1}\zeta)^2.
 \tag{\(c\le b\)}
\end{align*}
For every coordinate,
\(b|v_i|/(1+b\Pi_{ii})
=b|u_i|^{p-1}/(1+b|u_i|^{p-2})\le|u_i|\). Therefore
\begin{align*}
 |b\,v^\top\bar J^{-1}\zeta|
 &\le\sum_i|u_i\zeta_i|\\
 &\le h(\zeta),
 \tag{H\"older, \(\norm{u}_p=1\)}
\end{align*}
and substitution proves the claim.
\end{proof}

\subsubsection{Boundary estimate and the stability constant}
Suppose the diagonal term satisfies
\[
 (w-v)^\top\bar J^{-1}(w-v)\le M\omega
 \qquad\text{for some }M\ge0.
\]
Then
\begin{align*}
 \sqrt{w^\top J^{-1}w}
 &\le\sqrt{(w-v)^\top J^{-1}(w-v)}+\sqrt{v^\top J^{-1}v}
 \tag{triangle inequality}\\
 &\le\sqrt{M\omega+\eta h(w-v)^2}+\sqrt\eta
 \tag{Lemma~\ref{lem:diagonal-comparison}, \(v^\top J^{-1}v\le\eta\)}\\
 &\le\sqrt{M\omega}+3\sqrt\eta.
 \tag{\(h(w-v)\le2\)}
\end{align*}
Squaring and applying Cauchy--Schwarz with weights \(M/\rho\) and \(9\)
gives the geometric bound
\[
 w^\top J^{-1}w
 \le\left(9+\frac M\rho\right)(\eta+\rho\omega).
\]
The remaining task is to obtain \(M\) from the scalar estimate near the
boundary and balance it against the interior bound \(1/\eta\).

Choose a boundary cutoff and a coordinate threshold for
Lemma~\ref{lem:weighted-scalar}:
\[
 \bar\eta_p=\frac{\log p}{p},\qquad t_p=\min\{1/2,2/p\}.
\]
At these thresholds, the two scalar bounds give
\[
 M_p=\max\left\{
 \frac{(1-t_p)^2}{D_\varphi(1,t_p)},\,
 \frac{2}{\nu\{1+(p-1)\log(1/\bar\eta_p)t_p^{2-r}\}}\right\}.
\]
The curvature ratio is bounded by the larger of \(\bar\eta_p^{-1}\) and
\(9+M_p/(1-\bar\eta_p)\). Half this bound gives a stability constant;
we compare it with the uniform choice
\(\widetilde\kappa_p\):
\begin{equation}
 \bar\kappa_p=\min\left\{\widetilde\kappa_p,\,
 \frac12\max\left\{\bar\eta_p^{-1},\,9+\frac{M_p}{1-\bar\eta_p}\right\}\right\}.
 \label{eq:log-quad-constant}
\end{equation}
All quantities depend only on \(p\), and
\(\bar\kappa_p\ge\min\{\widetilde\kappa_p,9/2\}\ge1\).

\begin{proof}[Proof of Theorem~\ref{thm:quadratic-log}: the upper bound]
Lemma~\ref{lem:quad-reduction} gives the cap and reduces the claim to
bounding the curvature ratio by \(2\bar\kappa_p\), and
\(\bar\kappa_p\le\widetilde\kappa_p<p\) by \eqref{eq:log-quad-constant}.
By \eqref{eq:uniform-quad-curvature}, the ratio is at most
\(2\widetilde\kappa_p\); it remains to bound it by
\(\max\{\bar\eta_p^{-1},9+M_p/(1-\bar\eta_p)\}\).

\emph{Boundary regime.} If \(\eta\ge\bar\eta_p\), the interior
bound is at most \(\bar\eta_p^{-1}\). If \(\eta\le\bar\eta_p\), then
\(\rho\ge1-\bar\eta_p\) and \(b\ge(p-1)\log(1/\bar\eta_p)\).
For \(P=h^r/r\), the unit norms of \(v,w\) give
\(\sum_iD_\varphi(w_i,v_i)=D_P(w,v)=\omega\), and every coordinate of
\(v,w\) lies in \([-1,1]\). Hence
\begin{align*}
 (w-v)^\top\bar J^{-1}(w-v)
 &=\sum_i\frac{(w_i-v_i)^2}{1+b|v_i|^{2-r}}\\
 &\le\sum_i\max\Bigl\{\frac1{\varsigma_r(t_p)},
      \frac2{\nu(1+bt_p^{2-r})}\Bigr\}D_\varphi(w_i,v_i)
 \tag{by \eqref{eq:scalar-weighted}}\\
 &\le M_p\omega.
 \tag{\(b\ge(p-1)\log(1/\bar\eta_p)\)}
\end{align*}
The boundary estimate with \(M=M_p\) then gives
\begin{align*}
 \frac{w^\top J^{-1}w}{\eta+\rho\omega}
 &\le9+\frac{M_p}\rho\\
 &\le9+\frac{M_p}{1-\bar\eta_p}.
 \tag{\(\rho\ge1-\bar\eta_p\)}
\end{align*}
In both regimes the curvature ratio is at most
\(\max\{\bar\eta_p^{-1},9+M_p/(1-\bar\eta_p)\}\). With
\eqref{eq:uniform-quad-curvature}, it is at most \(2\bar\kappa_p\), and
Lemma~\ref{lem:quad-reduction} proves that \(F_2\) is
\((2,\bar\kappa_p)\)-shifted-loss stable.

\emph{Asymptotics.} For large \(p\), \(t_p=2/p\) and
\(\nu=1/(p-1)\). Expanding \(D_\varphi(1,t_p)\) gives
\begin{align*}
 D_\varphi(1,t_p)
 &=\frac{1-t_p^\nu}{1+\nu}
  -\frac{\nu t_p^\nu(1-t_p)}{1+\nu}\\
 &=\frac{1-e^{-\nu\log(p/2)}}{1+\nu}+O(1/p)
 \tag{\(\nu=O(1/p)\)}\\
 &\sim\nu\log(p/2)\sim\frac{\log p}{p}.
 \tag{\(\nu\log(p/2)\to0\)}
\end{align*}
Also, \(t_p^{2-r}=(2/p)e^{\nu\log(p/2)}\sim2/p\) and
\(\log(1/\bar\eta_p)=\log p-\log\log p\sim\log p\). Therefore
\begin{align*}
 \frac{(1-t_p)^2}{D_\varphi(1,t_p)}&\sim\frac p{\log p},
 \tag{\(t_p\to0\)}\\
 \frac{2}{\nu\{1+(p-1)\log(1/\bar\eta_p)t_p^{2-r}\}}
 &\sim\frac{2(p-1)}{2\log p}\sim\frac p{\log p}.
 \tag{\(t_p^{2-r}\sim2/p\)}
\end{align*}
Thus \(M_p\sim p/\log p\). Since
\(\bar\eta_p^{-1}=p/\log p\), \(\bar\eta_p\to0\), and
\(\widetilde\kappa_p\sim p\), both entries in the boundary maximum in
\eqref{eq:log-quad-constant} are asymptotic to \(p/\log p\), and its
boundary bound is eventually smaller than \(\widetilde\kappa_p\).
Consequently, \(\bar\kappa_p\sim p/(2\log p)\).
\end{proof}

\subsubsection{A matching lower bound}

The asymptotic constant cannot be reduced: a two-dimensional action near
the boundary gives a matching lower bound, and the example embeds in
every higher dimension. For \(d\ge2\), let \(\kappa^*_{p,d}\) be the least
\(\kappa\) such that \(F_2\) on \(B_p^d\) is \((2,\kappa)\)-shifted-loss
stable. For \(p>e\), the lower-bound witness uses \(t=1/\{p(\log p)^2\}\) and
\[
 J_{11}=1+p\log p\,t^{\frac{p-2}{p-1}}+(p-p\log p)t^2
 \quad\text{and}\quad
 \underline\kappa_p=\frac1{2J_{11}\{1-(1-1/p)t^{1/(p-1)}\}}.
\]
\begin{proposition}[Lower quadratic stability constant]
\label{prop:quad-constant-lower}
For every \(p>e\) and \(d\ge2\),
\(\kappa^*_{p,d}\ge \underline\kappa_p\), and
\(\underline\kappa_p\sim p/(2\log p)\). Consequently, with the upper
bound in Theorem~\ref{thm:quadratic-log},
\(\sup_{d\ge2}\bigl|\kappa^*_{p,d}\,(2\log p)/p-1\bigr|\to0\) as \(p\to\infty\).
\end{proposition}

\begin{proof}
\emph{Two-dimensional witness.}
Put
\(v=(t,(1-t^r)^{1/r})\), \(u=\nabla h(v)\), \(\rho=1-1/p\), and
\(x=\rho u\), so that \(u_1=t^{r-1}=t^{1/(p-1)}\),
\(\eta=1/p\), and \(b=p\log p\) in \eqref{eq:log-full-hessian}.
Take \(\beta=1\), \(y=\nabla F_2(x)\), and the loss
\(\theta=-\vartheta e_1\) with \(\vartheta>0\). Then \(s=\vartheta\) and
\(\ell^+(x,\theta)=\vartheta(1-\rho u_1)\). With \(J=\nabla^2F_2(x)\),
\(\nabla^2\Phi^{1,1}(y)=J^{-1}\), and \eqref{eq:log-full-hessian} gives
the displayed \(J_{11}\). Hence
\begin{align*}
 \kappa^*_{p,2}
 &\ge\lim_{\vartheta\downarrow0}
   \frac{D_{\Phi^{1,1}}(y+\vartheta e_1,y)}{\vartheta^2(1-\rho u_1)}
 \tag{by \eqref{eq:stability}}\\
 &=\frac{(J^{-1})_{11}}{2(1-\rho u_1)}
 \tag{Taylor}\\
 &\ge\frac1{2J_{11}(1-\rho u_1)}=\underline\kappa_p.
 \tag{\((e_1^\top Je_1)(e_1^\top J^{-1}e_1)\ge1\)}
\end{align*}

\emph{Higher dimensions.}
Embed \(y\) and the loss in \(\R^2\times\{0\}\).
Since \(g\) is nondecreasing, setting the other action coordinates to zero
preserves the linear objective and cannot increase \(F_2\). Thus \(\Phi^{1,1}\)
restricts to its two-dimensional version, with the same optimiser and
stability ratio. Hence \(\kappa^*_{p,d}\ge \underline\kappa_p\).

\emph{Asymptotics.}
Since \(-\log t=\log p+2\log\log p\sim\log p\),
\begin{align*}
 1-\rho u_1
 &=\frac1p+\left(1-\frac1p\right)\bigl(1-t^{1/(p-1)}\bigr)\\
 &=\frac1p+\frac{-\log t}{p}\,(1+o(1))
 \tag{\(-\log t=o(p)\)}\\
 &\sim\frac{\log p}{p}.
\end{align*}
The second term of \(J_{11}\) is
\(p\log p\,t^{(p-2)/(p-1)}=t^{-1/(p-1)}/\log p\sim1/\log p\),
and the last is \(O(p\log p\,t^2)=o(1)\). Hence \(J_{11}\to1\) and
\(\underline\kappa_p\sim p/(2\log p)\).
Finally, \(\underline\kappa_p\le\kappa^*_{p,d}\le\bar\kappa_p\) uniformly
in \(d\ge2\), and both endpoints are asymptotic to \(p/(2\log p)\).
This proves the uniform limit, the last claim of
Theorem~\ref{thm:quadratic-log}.
\end{proof}

\subsection{The effective dual point}
\label{app:mixed-potentials}

The quadratic stability proof, the pathwise comparisons with FTL, and the
stochastic analysis use the effective dual point.
Write \(\mathcal Q(x)=\tfrac12\norm{x}_2^2\) on \(\R^d\).

\begin{lemma}[Moreau envelope and quadratic augmentation]\label{lem:effective-dual-point}
Let \(\Omega_0\) be proper, convex, and lower semicontinuous, with
\(P=\Omega_0^*\) finite and differentiable on \(\R^d\).
For \(q\ge0\), put \(\Omega_q=\Omega_0+q\mathcal Q\) and
\(\Phi_q=\Omega_q^*\). Then \(\Phi_q\) is differentiable. Fix \(y\),
and write \(x=\nabla\Phi_q(y)\) and \(w=y-qx\). We have \(x=\nabla P(w)\),
and, for \(q>0\),
\begin{equation}
 \Phi_q(y)=\min_{u\in\R^d}
 \left\{P(u)+\frac{\norm{y-u}_2^2}{2q}\right\},
 \label{eq:mixed-moreau-envelope}
\end{equation}
with unique minimiser \(w\).
\end{lemma}

\begin{proof}[Proof of Lemma~\ref{lem:effective-dual-point}]
Since \(q\mathcal Q\) is finite everywhere, the conjugate of
\(\Omega_0+q\mathcal Q\) is the infimal convolution of \(P\) with
\((q\mathcal Q)^*=\norm{\cdot}_2^2/(2q)\)
\citep[Theorem~16.4]{rockafellar1970convex}, which is
\eqref{eq:mixed-moreau-envelope}. The minimand is coercive and strictly
convex, so its minimiser \(w\) is unique and satisfies
\(y=w+q\nabla P(w)\). Strong convexity of \(\Omega_q\) makes \(\Phi_q\)
differentiable, and the Moreau envelope has gradient
\(\nabla\Phi_q(y)=(y-w)/q\), so
\(x=\nabla P(w)\). At \(q=0\), the assertions reduce to \(\Phi_0=P\) and
\(w=y\).
\end{proof}

For \(\Omega_0=bF_1\), so that \(\Phi_q=\Phi^{b,q}\),
\eqref{eq:families} gives \(\nabla P(w)=(1-e^{-h(w)/b})f(w)\). The
augmented action is therefore the radial action at its effective dual
point:
\begin{equation}
 x=\bigl(1-e^{-h(w)/b}\bigr)f(w),\qquad w=y-qx.
 \label{eq:effective-radial-action}
\end{equation}

\section{Tuning by the shifted loss}
\label{app:regret-proofs}

We turn the local stability estimates of Appendix~\ref{app:lp-proofs}
into a regret bound in the learner's shifted loss, and bound FTL's regret by
the slack. Appendix~\ref{app:calibrated-standalone} then proves the
standalone guarantee in the slack and its sharp coefficient.
Throughout, \(m_p\) and \(d_*\) are as in
\eqref{eq:calibrated-profile}.

\subsection{Tuning from observed shifted loss}
\label{app:one-clock}

For the action \eqref{eq:monotone-action}, with \(c=2^{1/r}\), define the stability coefficient, crossover charge,
and regularisation-cost bound
\[
 \Upsilon_0=\max\Bigl\{5\cdot3^{\frac12-\frac1p}\varepsilon_p^{-\frac1{p-1}},\
      \sqrt6\,\frac{\bar\kappa_p}{m_p}\Bigr\},
\]
\[
 J_*=\varepsilon_p(d_*+2)^{1/r}\one\{p>2\},\qquad
 \Upsilon=\Upsilon_0+3\varepsilon_p\one\{p>2\},
\]
\[
 \mathcal C(u)=c+\frac1{\sqrt2}+\Bigl(1+\frac1{\sqrt2}\Bigr)W_p(u,d_*).
\]
At \(p=2\), the potential is nonincreasing even at crossover, so no
jump charge is needed.

\begin{proposition}[Tuning with one change of regulariser]\label{prop:tuning}
Suppose the learner plays
\(x_t=x(U_{t-1},Y_{t-1})\) on a causally chosen
set of rounds and any \(x_t\in\mathcal K\) on the remaining rounds.
Here \(U_t\) sums the played shifted losses on selected rounds \(i\le t\),
\(Y_t\) includes every loss, and
\(\Phi_t=\Phi^{\beta_1(U_{t-1}),\beta_2(U_{t-1})}\).
On every selected round,
\begin{equation}
 D_{\Phi_t}(Y_t,Y_{t-1})\le\Upsilon_0[W_p(U_t,d_*)-W_p(U_{t-1},d_*)],
 \qquad 0\le h-\Phi_t\le \mathcal C(U_t).
 \label{eq:complexity-increment}
\end{equation}
Along selected rounds, the potential is nonincreasing except possibly
at the first change to \(F_2\); that increase is at most \(J_*\) pointwise.
Consequently:
\begin{enumerate}[label=(\roman*),nosep,leftmargin=*]
\item if every round is selected, then \(U=S\) and
\(\Reg_T\le\Upsilon W_p(S_T,d_*)+\mathcal C(S_T)\);
\item if the selected rounds are the FTRL rounds of
Algorithm~\ref{alg:lp-hybrid} (Section~\ref{sec:lp-switching}), then
\[
 \Reg_T\le5[2\Upsilon W_p(U_T,d_*)+3\mathcal C(U_T)]+2-c.
\]
\end{enumerate}
\end{proposition}

\paragraph{Stability, crossover, and support deficit.}
For the selected-round sum \(U_t\) in Proposition~\ref{prop:tuning}, set
\[
 E_t=\Upsilon_0W_p(U_t,d_*)+J_*\one\{U_t\ge d_*\}.
\]
This sequence is nonnegative and nondecreasing, with \(E_0=0\).
The step in \(E_t\) pays for the only possible positive change of
potential. Since \(J_*\le3\varepsilon_p\one\{p>2\}W_p(u,d_*)\) for
\(u\ge d_*\) (proved below), \(E_t\le\Upsilon W_p(U_t,d_*)\).

On a consecutive run \([a,b]\) of selected rounds, the decomposition
behind \eqref{eq:one-clock-telescope} gives
\begin{align}
 \Reg_b-\Reg_{a-1}
 &=\sum_{t=a}^bD_{\Phi_t}(Y_t,Y_{t-1})
   +\sum_{t=a}^{b-1}(\Phi_{t+1}-\Phi_t)(Y_t)
   -(h-\Phi_a)(Y_{a-1})+(h-\Phi_b)(Y_b)\notag\\
 &\le\Upsilon_0[W_p(U_b,d_*)-W_p(U_{a-1},d_*)]
   +J_*\one\{U_{a-1}<d_*\le U_b\}+\mathcal C(U_b)
 \tag{by \eqref{eq:complexity-increment}}\\
 &\le E_b-E_{a-1}+\mathcal C(U_b).
 \label{eq:potential-block-budget}
\end{align}
In the second line, the stability terms telescope by
\eqref{eq:complexity-increment}; the potential changes are positive only
at the first change to \(F_2\), by at most \(J_*\); and the support
deficits lie in \([0,\mathcal C(U_b)]\). These are the per-round claims of
Proposition~\ref{prop:tuning}, proved below. The charge is made once across the entire
execution, even when the selected rounds form many FTRL blocks. If every
round is selected, \([a,b]=[1,T]\) gives \(\Reg_T\le E_T+\mathcal C(U_T)\).

\paragraph{Concave increments.}
For a continuous increasing concave profile \(W\) on \([0,\infty)\) with
\(W(0)=0\), and for \(u\ge0\) and \(0\le z\le2\),
\begin{align}
 zW'_+(u+2)
 &\le\int_u^{u+z}W'_+(v)\dif v
 \tag{\(W'_+\) nonincreasing, \(u+z\le u+2\)}\\
 &=W(u+z)-W(u).
 \label{eq:concave-budget}
\end{align}

\begin{proof}[Proof of Proposition~\ref{prop:tuning}]
Write \(W=W_p(\cdot,d_*)\), fix a selected round \(t\), and put
\(u=U_{t-1}\), \(z=\ell^+(x_t,\theta_t)\), so \(U_t=u+z\).
By \eqref{eq:concave-budget}, it suffices to bound the stability coefficient by
\(\Upsilon_0 W'_+(u+2)\); the increment inequality then gives
\begin{equation}
 D_{\Phi_t}(Y_t,Y_{t-1})\le\Upsilon_0 zW'_+(u+2)
 \le\Upsilon_0[W(U_t)-W(U_{t-1})].
 \label{eq:tuning-stability}
\end{equation}
We verify this coefficient bound for each temperature regime below, using
\(h(\theta_t)\le1\). The two branches of \(W\) meet at \(d_*\).
Its right derivative is \(W'_+(v)=r^{-1}v^{-1/p}\) for \(0<v<d_*\),
and \(W'_+(v)=d_*^{1/2-1/p}/(2\sqrt v)\) for \(v\ge d_*\).

\emph{Radial rounds.} If \(u<d_*\), the temperatures are
\((\varepsilon_p(u+2)^{1/r},0)\), and
\begin{align*}
 D_{\Phi_t}(Y_t,Y_{t-1})
 &\le\tfrac52\bigl(\varepsilon_p(u+2)^{1/r}\bigr)^{-(r-1)}z
 \tag{Lemma~\ref{lem:power}}\\
 &=\tfrac52\varepsilon_p^{-1/(p-1)}(u+2)^{-1/p}z.
 \tag{\(r-1=1/(p-1)\)}
\end{align*}
If \(u+2<d_*\), then \(W'_+(u+2)=r^{-1}(u+2)^{-1/p}\), and the
coefficient ratio is
\(\tfrac52r\varepsilon_p^{-1/(p-1)}\le5\varepsilon_p^{-1/(p-1)}\).
If \(u+2\ge d_*\), then
\begin{align*}
 \frac{\tfrac52\varepsilon_p^{-1/(p-1)}(u+2)^{-1/p}}{W'_+(u+2)}
 &=5\varepsilon_p^{-\frac1{p-1}}\Bigl(\frac{u+2}{d_*}\Bigr)^{\frac12-\frac1p}
 \tag{\(W'_+(v)=\tfrac{d_*^{1/2-1/p}}{2\sqrt v}\)}\\
 &\le5\cdot3^{\frac12-\frac1p}\varepsilon_p^{-\frac1{p-1}}.
 \tag{\(u+2<d_*+2\le3d_*\)}
\end{align*}
Both ratios are at most \(\Upsilon_0\).

\Needspace{6\baselineskip}
\emph{Quadratic rounds.} If \(u\ge d_*\), then
\(\beta_1=\beta_2=\sqrt{m_p(u+2)/\bar F_2}\), and \((2,\bar\kappa_p)\)-shifted-loss
stability of \(F_2\) (Theorem~\ref{thm:quadratic-log}) bounds the divergence
by \(\bar\kappa_p z/\beta_2\).
The coefficient ratio is
\begin{align*}
 \frac{\bar\kappa_p/\beta_2}{W'_+(u+2)}
 &=\frac{2\bar\kappa_p\sqrt{\bar F_2}}{\sqrt{m_p}\,d_*^{1/2-1/p}}
 \tag{\(\beta_2=\sqrt{m_p(u+2)/\bar F_2}\)}\\
 &=\frac{2\bar\kappa_p}{m_p}\frac{\sqrt{\bar F_2}}{d^{1/2-1/p}}
 \tag{\(d_*^{1-2/p}=m_pd^{1-2/p}\)}\\
 &\le\sqrt6\,\bar\kappa_p/m_p\le\Upsilon_0.
 \tag{\(\bar F_2\le\tfrac32d^{1-2/p}\)}
\end{align*}
This completes the proof of \eqref{eq:tuning-stability}.

\emph{Regularisation cost.}
We bound both parts of the cap at \(u=U_{t-1}\).
On a radial round, \(\beta_2=0\) and
\(\beta_1\le(u+2)^{1/r}\le c+u^{1/r}=c+W(u)\), by subadditivity.
On a quadratic round, \(u\ge d_*\ge1\) gives
\(\sqrt{u+2}\le\sqrt u+1/\sqrt{d_*}\), and \(W(u)=d_*^{1/2-1/p}\sqrt u\). Hence
\begin{align*}
 \beta_2=\sqrt{m_p/\bar F_2}\,\sqrt{u+2}
 &\le\sqrt{m_p/\bar F_2}\,\bigl(\sqrt u+1/\sqrt{d_*}\bigr)\\
 &\le\frac{W(u)+1}{d^{1/2-1/p}\sqrt{\bar F_2}}
 \tag{\(\sqrt{m_p}\,d^{1/2-1/p}=d_*^{1/2-1/p}\le\sqrt{d_*}\)}\\
 &\le1+W(u)\le c+W(u).
 \tag{\(d^{1-2/p}\bar F_2\ge1\)}
\end{align*}
Since \(\beta_1=\beta_2\) on quadratic rounds,
\(\beta_1\le c+W(u)\) in both regimes. Also
\(d^{1-2/p}<2\bar F_2\). Therefore
\begin{align*}
 0\le h-\Phi_t
 &\le\beta_1+\frac{\beta_2}2d^{1-2/p}
 \tag{caps of \(F_1\) and \(F_2\)}\\
 &\le c+W(u)+\frac{1+W(u)}{\sqrt{\bar F_2}}\cdot\frac{d^{1/2-1/p}}2
 \tag{bounds on \(\beta_1,\beta_2\)}\\
 &\le c+W(u)+\frac{1+W(u)}{\sqrt2}
 \tag{\(d^{1-2/p}<2\bar F_2\)}\\
 &=\mathcal C(u)\le\mathcal C(U_t).
\end{align*}

\emph{Changes of potential.} Within each phase, the temperature is
nondecreasing, so the potential is nonincreasing. At the transition,
write \(b\) for the preceding radial temperature and
\(\Phi_{\mathrm{old}},\Phi_{\mathrm{new}}\) for the two potentials.
At every \(y\),
\begin{align*}
 \Phi_{\mathrm{new}}(y)-\Phi_{\mathrm{old}}(y)
 &\le h(y)-\Phi_{\mathrm{old}}(y)
 \tag{\(\Phi_{\mathrm{new}}\le h\)}\\
 &\le b<\varepsilon_p(d_*+2)^{1/r}.
 \tag{cap of \(bF_1\)}
\end{align*}
For \(p>2\), the right side is \(J_*\), and for \(u\ge d_*\ge1\),
\begin{align*}
 J_*=\varepsilon_p(d_*+2)^{1/r}
 &\le3\varepsilon_p d_*^{1/r}
 \tag{\(d_*+2\le3d_*\)}\\
 &\le3\varepsilon_p W(u).
 \tag{\(W(d_*)=d_*^{1/r}\), \(u\ge d_*\)}
\end{align*}
This proves the charge and the bound on \(E_t\) when \(p>2\).

At \(p=2\), no charge is needed. Indeed,
\(\log(1-\rho)\le-\rho\) gives \(g(\rho)\le\rho^2\), hence
\(F_2\ge\tfrac32F_1\) on \(B_2^d\).
The new temperature is \(q=\sqrt{2(u+2)/3}\) with \(u\ge d_*\), so
\begin{align*}
 qF_2
 &\ge\tfrac32qF_1
 \tag{\(F_2\ge\tfrac32F_1\)}\\
 &\ge\sqrt{d_*+2}\,F_1\ge bF_1.
 \tag{\(\tfrac32q\ge\sqrt{u+2}\), \(u\ge d_*\)}
\end{align*}
Thus the new potential is at most the old one.

\emph{Regret bounds.} If every round is selected, \(U=S\), and
\eqref{eq:potential-block-budget} with \([a,b]=[1,T]\) gives
\(\Reg_T\le E_T+\mathcal C(S_T)\le\Upsilon W(S_T)+\mathcal C(S_T)\).
For switching, \eqref{eq:potential-block-budget} holds on each block,
and Lemma~\ref{lem:switch-accounting} proves (ii), as detailed in
Appendix~\ref{app:single-safe}.
\end{proof}

\subsection{Temperature growth, the implicit inequality, and radial tracking}
\label{app:constants}

For Algorithm~\ref{alg:one-clock}, write
\(b_t=\beta_1(S_{t-1})\) and \(q_t=\beta_2(S_{t-1})\).
Since \(S_n\le2n\), its rule gives, for \(n\ge1\),
\begin{equation}
 q_{n+1}\le2\sqrt{m_p n/\bar F_2},\qquad
 b_{n+1}\le2\sqrt{m_p}(d+2)^{1/r-1/2}\sqrt n.
 \label{eq:standalone-temperature-growth}
\end{equation}
Indeed, \(q_{n+1}^2\le m_p(S_n+2)/\bar F_2\) and \(S_n+2\le4n\).
If \(S_n\ge d_*\), then
\(b_{n+1}=q_{n+1}\), and \(\bar F_2\ge1\) gives the second bound.
Otherwise \(b_{n+1}=\varepsilon_p(S_n+2)^{1/r}\) with \(\varepsilon_p\le1\), and
\begin{align*}
 (S_n+2)^{1/r}
 &=(S_n+2)^{1/r-1/2}\sqrt{S_n+2}\\
 &\le(d_*+2)^{1/r-1/2}\sqrt{S_n+2}
 \tag{\(S_n<d_*\)}\\
 &\le\sqrt{m_p}(d+2)^{1/r-1/2}\sqrt{S_n+2}.
 \tag{\(d_*+2\le\tfrac{d_*}d(d+2)\)}
\end{align*}
The last step also uses \((d_*/d)^{1/r-1/2}=(d_*/d)^{1/2-1/p}=\sqrt{m_p}\).

\begin{lemma}[An implicit quadratic regret inequality]\label{lem:exact-closure}
If \(A,b,\Delta\ge0\) and \(R\le A+b\sqrt{\Delta+R}\) with
\(\Delta+R\ge0\), then \(R\le\tfrac32(A+b^2)+b\sqrt\Delta\).
\end{lemma}

\begin{proof}
If \(R\le A\), the claim is immediate. Otherwise, squaring
\(R-A\le b\sqrt{\Delta+R}\) gives \((R-A)^2\le b^2(\Delta+R)\), and
\begin{align*}
 R&\le A+\tfrac12b^2+b\sqrt{\Delta+A+b^2/4}
 \tag{upper root}\\
  &\le A+b^2+b\sqrt\Delta+b\sqrt A
 \tag{subadditivity of \(\sqrt\cdot\)}\\
  &\le\tfrac32(A+b^2)+b\sqrt\Delta.
 \tag{\(b\sqrt A\le(A+b^2)/2\)}
\end{align*}
\end{proof}

\begin{lemma}[Radial tracking on selected rounds]\label{lem:radial-tracking}
Fix \(T\), \(I\subseteq\{1,\ldots,T\}\), and \(B>0\). Suppose the learner
plays \(f(Y_{t-1})\) outside \(I\), and
\(x_t=(1-e^{-h(Y_{t-1})/\beta_t})f(Y_{t-1})\) on \(I\), with
\(0<\beta_t\le B\). Then
\begin{equation}
 \Reg_T\le\Reg_T^{\FTL}+1+\Delta_T+B.
 \label{eq:radial-tracking}
\end{equation}
\end{lemma}

\begin{proof}
Put \(s_t=h(\theta_t)\le1\) and \(\Delta=\Delta_T\). Then
\begin{align*}
 \Reg_T-\Reg_T^{\FTL}
 &=-\sum_{t\in I}e^{-h(Y_{t-1})/\beta_t}\ip{f(Y_{t-1})}{\theta_t}\\
 &\le\sum_{t\in I}s_te^{-h(Y_{t-1})/\beta_t}
 \tag{H\"older}\\
 &\le\sum_{t=1}^T s_t e^{-(H_{t-1}-\Delta)_+/B}
 \tag{\(h(Y_{t-1})=H_{t-1}-\Delta_{t-1}\), \(\beta_t\le B\)}\\
 &\le1+\int_0^\infty e^{-(v-\Delta)_+/B}\dif v
 \tag{Riemann sum}\\
 &=1+\Delta+B.
\end{align*}
The third line also uses \(\Delta_{t-1}\le\Delta\) and \(h\ge0\).
In the fourth, the sum is a left Riemann sum, over the points
\(H_0\le H_1\le\cdots\), of a decreasing function with values in
\([0,1]\). Each interval has length \(s_t\le1\), so its excess over the
integral is at most the drop in the integrand; these drops telescope
to at most one.
\end{proof}

\subsection{Radius-dependent FTL stability}
\label{app:radius-ftl}
Lemma~\ref{lem:radius-ftl} combines a contraction implied by
approximate Bregman symmetry with Lemma~\ref{lem:norm-increment}. The contraction requires no norm structure
or homogeneity, and approximate symmetry is needed only along the segment.
The conclusion is the fixed-base case of the triangle-scaling inequality
of \citet[Definition~2]{hanzely2021accelerated}; here we derive it from
approximate symmetry.

\begin{lemma}[Segment contraction from approximate symmetry]
\label{lem:symmetry-contraction}
Let \(P\colon\mathcal D\to\R\) be differentiable and convex on an open
convex set \(\mathcal D\subseteq\R^d\). Fix \(u,v\in\mathcal D\) and
\(A>0\), and write \(u_t=u+t(v-u)\).
If \(D_P(u,u_t)\ge A^{-1}D_P(u_t,u)\) for every \(0<t\le1\), then
\[
 D_P(u_t,u)\le t^{1+1/A}D_P(v,u),\qquad 0\le t\le1.
\]
\end{lemma}

\begin{proof}
Set \(B(t)=D_P(u_t,u)\). Since \(u_t-u=t(v-u)\), for \(0<t\le1\),
\begin{align*}
 tB'(t)-B(t)
 &=\ip{\nabla P(u_t)-\nabla P(u)}{u_t-u}-D_P(u_t,u)\\
 &=D_P(u,u_t)
 \tag{definition of \(D_P\)}\\
 &\ge B(t)/A.
 \tag{approximate symmetry}
\end{align*}
Thus
\(\tfrac{\dif}{\dif t}\{t^{-1-1/A}B(t)\}
=t^{-2-1/A}\{tB'(t)-(1+1/A)B(t)\}\ge0\), so
\(t^{-1-1/A}B(t)\le B(1)=D_P(v,u)\) for \(0<t\le1\). At \(t=0\), both
sides vanish.
\end{proof}

\begin{proof}[Proof of Lemma~\ref{lem:radius-ftl}]
Use the notation of Lemma~\ref{lem:norm-increment} with \(q=r\) and
\(a=f(y)\), and put \(\lambda=t^{1/(2p-2)}\in(0,1)\).
For \(P=h^r/r\), the approximate
symmetry bound of \mbox{\citet[Theorem~1]{kindermann2019symmetry}} gives the
hypothesis of Lemma~\ref{lem:symmetry-contraction} with \(A=2p-2\).
Since \(m=u+t(v-u)\), Lemma~\ref{lem:symmetry-contraction} then gives
\[
 D_P(m,u)\le t^{1+1/(2p-2)}D_P(v,u).
\]
With \(C=1\) and \(\gamma=1+1/(2p-2)\),
Lemma~\ref{lem:norm-increment}\ref{item:norm-increment-loss}
and~\ref{item:norm-increment-contraction} turn this into
\(\tau\le\lambda z_f=\lambda(\delta+\tau)\). Rearranging,
\begin{align*}
 \tau
 &\le\frac{\lambda}{1-\lambda}\,\delta
 \tag{\(\lambda<1\)}\\
 &=\frac{\delta}{(1+R/s)^{1/(2p-2)}-1}.
\end{align*}
\end{proof}

\begin{proof}[Proof of Proposition~\ref{prop:ftl-log-p}]
Splitting after the last step from within radius \(\bar R\) balances an
initial cost of order \(\bar R\) against the later cost per unit of
slack. Let \(j\) be the last round with \(h(Y_{j-1})\le\bar R\); it exists since
\(Y_0=0\). Then \(h(Y_j)\le h(Y_{j-1})+1\le\bar R+1\).

\emph{Rounds up to \(j\).} By \eqref{eq:protocol},
\begin{align*}
 \Reg_j^{\FTL}
 &=\sum_{t\le j}\ip{x_t}{\theta_t}+h(Y_j)\\
 &\le H_j+h(Y_j)
 \tag{\(x_t\in\mathcal K\)}\\
 &=\Delta_j+2h(Y_j)\le\Delta_j+2(\bar R+1).
\end{align*}

\emph{Rounds after \(j\).} For \(t>j\), \(h(Y_{t-1})>\bar R\).
Lemma~\ref{lem:radius-ftl}, \(h(\theta_t)\le1\), and
\eqref{eq:ftl-increments} give
\(\Reg_t^{\FTL}-\Reg_{t-1}^{\FTL}\le b_p(\bar R)(\Delta_t-\Delta_{t-1})\);
a zero loss contributes zero to both sides. Summing,
\begin{align*}
 \Reg_T^{\FTL}
 &\le2(\bar R+1)+\Delta_j+b_p(\bar R)(\Delta_T-\Delta_j)\\
 &\le2(\bar R+1)+\max\{1,b_p(\bar R)\}\Delta_T.
 \tag{\(0\le\Delta_j\le\Delta_T\)}
\end{align*}
This proves \eqref{eq:ftl-radius-cut}. The bound with \(C_p^{\FTL}\)
follows as \(2\bar R+2\ge1\).

\emph{Asymptotics.} At the chosen radius,
\(\log(1+\bar R)\sim\log p=o(p)\), so
\[
 b_p(\bar R)=\frac1{e^{\log(1+\bar R)/(2p-2)}-1}
 \sim\frac{2p-2}{\log(1+\bar R)}\sim\frac{2p}{\log p},
 \qquad 2\bar R+2\sim\frac{2p}{\log p}.
\]
\end{proof}

\section{The switching extension}
\label{app:switching}

We prove Proposition~\ref{prop:standalone-separation} and
Lemma~\ref{lem:single-safe}, with explicit constants, and show that the
known loss-norm bound is needed.

\subsection{Pathwise comparison with FTL}
\label{app:ftl-comparison}

The effective dual point and the deterministic temperature estimates
\eqref{eq:standalone-temperature-growth} suffice for the two
separations. They share with the stochastic limits of
Appendix~\ref{app:persistent-gap} a deterministic consistency argument,
which needs only empirical convergence of the losses.

\begin{lemma}[Empirical limits of the adaptive learner]\label{lem:empirical-limits}
Let \(h(\theta_t)\le1\), and suppose
\(\bar\theta_n=n^{-1}\sum_{t\le n}\theta_t\to\mu\ne0\) and
\(n^{-1}\sum_{t\le n}h(\theta_t)\to m\).
Put \(\gamma=m-h(\mu)\ge0\).
Algorithm~\ref{alg:one-clock} satisfies
\begin{equation}
 x_{n+1}\to-\nabla h(\mu),\qquad S_n/n\to\gamma,\qquad
 q_{n+1}^2/n\to m_p\gamma/\bar F_2,
 \label{eq:empirical-clock-limits}
\end{equation}
and \(1-\norm{x_{n+1}}_p=O(e^{-c\sqrt n})\) for some \(c>0\).
\end{lemma}

\begin{proof}
\emph{Action.}
The deterministic bounds \eqref{eq:standalone-temperature-growth} give
\(q_{n+1},b_{n+1}=O_{p,d}(\sqrt n)\). Feasibility implies
\(a_n:=\bar\theta_n+q_{n+1}x_{n+1}/n\to\mu\).
Lemma~\ref{lem:effective-dual-point}, at the effective point
\(Y_n-q_{n+1}x_{n+1}=-na_n\), and symmetry give
\[
 x_{n+1}=-\alpha_{n+1}f(a_n),\qquad
 \alpha_{n+1}=1-e^{-nh(a_n)/b_{n+1}}.
\]
Since \(h(a_n)\to h(\mu)>0\) and \(b_{n+1}=O(\sqrt n)\),
\(1-\alpha_{n+1}=O(e^{-c\sqrt n})\),
and continuity of \(f\) away from zero gives \(x_{n+1}\to-\nabla h(\mu)\).

\emph{Shifted loss.}
Since \(S_n=\sum_{t\le n}\{h(\theta_t)+\ip{x_t}{\theta_t}\}\),
\begin{align*}
 \Bigl|S_n-\sum_{t\le n}\{h(\theta_t)-\ip{\nabla h(\mu)}{\theta_t}\}\Bigr|
 &=\Bigl|\sum_{t\le n}\ip{x_t+\nabla h(\mu)}{\theta_t}\Bigr|\\
 &\le\sum_{t\le n}\norm{x_t+\nabla h(\mu)}_p
 \tag{H\"older, \(h(\theta_t)\le1\)}\\
 &=o(n).
 \tag{\(x_t\to-\nabla h(\mu)\)}
\end{align*}
The empirical assumptions and \(\ip{\nabla h(\mu)}\mu=h(\mu)\) therefore
give \(S_n/n\to m-h(\mu)=\gamma\).
If \(\gamma>0\), the quadratic term is eventually active and
\(q_{n+1}^2=m_p(S_n+2)/\bar F_2\); if \(\gamma=0\), use
\(q_{n+1}^2\le m_p(S_n+2)/\bar F_2\).
\end{proof}

\begin{lemma}[Pathwise collinear bias]\label{lem:collinear-bias}
Let \(h(\eta)=1\), and suppose \(p=2\) or every coordinate of \(\eta\)
is nonzero. Take \(\theta_t=\xi_t\eta\), with \(|\xi_t|\le1\), and suppose
\(n^{-1}\sum_{t\le n}\xi_t\to\rho>0\) and
\(n^{-1}\sum_{t\le n}|\xi_t|\to m\).
Put \(\gamma=m-\rho\).
Algorithm~\ref{alg:one-clock} satisfies
\begin{equation}
 \frac{\Reg_T}{\log T}\longrightarrow
 \frac{m_p\gamma}{2\rho\bar F_2}\nabla h(\eta)^\top\nabla^2h(\eta)\nabla h(\eta).
 \label{eq:pathwise-collinear-bias}
\end{equation}
\end{lemma}

\begin{proof}
\emph{Regret asymptotics.}
Put \(V_n=\sum_{t\le n}\xi_t\), \(q_{n+1}=\beta_2(S_n)\), and
\(b_{n+1}=\beta_1(S_n)\). Eventually \(V_n>0\), and
Lemma~\ref{lem:effective-dual-point} gives
\[
 a_n=\eta+\frac{q_{n+1}}{V_n}x_{n+1},\qquad
 x_{n+1}=-\alpha_{n+1}f(a_n),\qquad
 \alpha_{n+1}=1-e^{-V_nh(a_n)/b_{n+1}}.
\]
Lemma~\ref{lem:empirical-limits}, applied with mean \(\rho\eta\),
gives \(x_{n+1}\to-\nabla h(\eta)\), \(S_n/n\to\gamma\),
\(q_{n+1}^2/n\to m_p\gamma/\bar F_2\), and exponential damping.
Also \(a_n\to\eta\), since \(q_{n+1}/V_n\to0\).
Set \(d_t=1+\ip\eta{x_t}\). Since \(h(\eta)=1\),
\begin{align*}
 d_{n+1}
 &=(1-\alpha_{n+1})+\alpha_{n+1}\{1-\ip{f(a_n)}\eta\}\\
 &=(1-\alpha_{n+1})+\alpha_{n+1}D_h(\eta,a_n).
 \tag{\(D_h(\eta,a)=h(\eta)-\ip{f(a)}\eta\)}
\end{align*}
With \(a_n-\eta=(q_{n+1}/V_n)x_{n+1}\), \(x_{n+1}\to-\nabla h(\eta)\), and
\(nq_{n+1}^2/V_n^2\to m_p\gamma/(\rho^2\bar F_2)\), the reverse expansion
in \eqref{eq:support-taylor} and exponential damping give
\[
 nd_{n+1}\to\Gamma:=\frac{m_p\gamma}{2\rho^2\bar F_2}
       \nabla h(\eta)^\top\nabla^2h(\eta)\nabla h(\eta).
\]
For all sufficiently large \(T\), \(V_T>0\), so
\(h(Y_T)=V_T=\sum_{t\le T}\xi_t\). Hence
\begin{align*}
 \Reg_T
 &=\sum_{t\le T}\xi_t\ip{x_t}{\eta}+\sum_{t\le T}\xi_t
 =\sum_{t\le T}\xi_td_t
 \tag{by \eqref{eq:protocol}}\\
 &=\Gamma\sum_{t\le T}\frac{\xi_t}t+o(\log T)
 \tag{\(d_t=\Gamma/t+o(1/t)\), \(|\xi_t|\le1\)}\\
 &=\Gamma\rho\log T+o(\log T),
 \tag{Abel summation, \(V_n=\rho n+o(n)\)}
\end{align*}
which is \eqref{eq:pathwise-collinear-bias}.
\end{proof}

\begin{proof}[Proof of Proposition~\ref{prop:standalone-separation}\ref{item:radial-separation}]
Since \(T\ge4\), \(\lambda\le1/16\). All cumulative sums are positive. Writing \(R_n=|Y_n|\), we have
\[
 R_n=\lambda n+(1-\lambda)\one\{n\text{ odd}\},\qquad
 H_n=(1-\lambda)n+\lambda\one\{n\text{ odd}\}.
\]
In particular, \(R_n\ge\lambda n\), and FTL plays \(-1\) after its
initial zero action, giving regret exactly one at every horizon.

For Algorithm~\ref{alg:one-clock}, write \(x_t=-\alpha_t\), with
\(\alpha_1=0\). At \(p=2\), the stability constant is
\(\bar\kappa_2=(2+\sqrt3)/2\), so in dimension one
\(d_*=7/4+\sqrt3\in(3.45,3.5)\), and \(m_2=\varepsilon_2=1\).
Each pair of rounds adds
\begin{equation*}
 \Reg_{2k}-\Reg_{2k-2}
 =-\alpha_{2k-1}+(1-2\lambda)\alpha_{2k}+2\lambda
 =2\lambda(1-\alpha_{2k})+\alpha_{2k}-\alpha_{2k-1}.
\end{equation*}
It therefore suffices to show \(\alpha_t\le1/2\) and
\(\alpha_{2k}\ge\alpha_{2k-1}\) for all rounds: then every pair adds at
least \(\lambda\), and summing over \(T/2\) pairs gives
\(\Reg_T\ge\lambda T/2=\sqrt T/16\).

\emph{Rounds one to four.}
Before crossover, \(\alpha_{t+1}=1-e^{-R_t/\sqrt{S_t+2}}\).
Put \(c_0=1-e^{-1/\sqrt3}<0.45\). Since \(S_1=1\) and \(R_1=1\),
\begin{align*}
 \alpha_2&=c_0,\\
 S_2&=1+(1-2\lambda)(1+c_0)\in[15/8,\,2+c_0],
 \tag{\(\lambda\le1/16\)}\\
 \alpha_3&=1-e^{-2\lambda/\sqrt{S_2+2}}<1/10,
 \tag{\(2\lambda\le1/8\), \(S_2\ge15/8\)}\\
 S_3&=S_2+1-\alpha_3\le3+c_0<d_*.
\end{align*}
Thus round four is radial, with
\(\alpha_4=1-e^{-(1+2\lambda)/\sqrt{S_3+2}}\in(1/10,1/2)\), since
\(15/8\le S_3<d_*<3.5\) and \(1\le1+2\lambda\le9/8\).
Hence \(\alpha_t\le1/2\) through round four, and
\(\alpha_2\ge\alpha_1\), \(\alpha_4\ge\alpha_3\). The first two pairs
add nonnegative regret, so
\(S_4\ge\Delta_4=4-8\lambda\ge3.5>d_*\), and the quadratic term is
active from round five. There, both temperatures equal
\(q(u)=\sqrt{2(u+2)/3}\), where \(u\) is the cumulative shifted loss.

\emph{Bounded radius.} Suppose \(\alpha_t\le1/2\) through round
\(4\le n<T\). Each shifted loss is \(s_t(1\pm\alpha_t)\ge s_t/2\), so
\(S_n\ge H_n/2\ge3n/8\). The scalar first-order condition reads
\(R_n=q(S_n)\{-\log(1-\alpha_{n+1})+\alpha_{n+1}\}\). Therefore
\begin{align*}
 \alpha_{n+1}
 &\le\frac{R_n}{2q(S_n)}
 \tag{\(-\log(1-\alpha)\ge\alpha\)}\\
 &\le\frac{\lambda n+1}{2\sqrt{2/3}\sqrt{3n/8+2}}
 \tag{\(S_n\ge3n/8\)}\\
 &\le\frac{\sqrt{13/24}}{2\sqrt{2/3}}=\frac{\sqrt{13}}8<\frac12,
 \tag{see below}
\end{align*}
where \(\lambda\sqrt n\le1/8\) and Cauchy--Schwarz give
\begin{align*}
 \frac{(\lambda n+1)^2}{3n/8+2}
 &\le\frac{(\sqrt n/8+1)^2}{3n/8+2}\\
 &\le\frac{n/64}{3n/8}+\frac12=\frac{13}{24}.
 \tag{\(\frac{(a+b)^2}{A+B}\le\frac{a^2}A+\frac{b^2}B\)}
\end{align*}

\emph{Increasing radius within pairs.} Let \(k\ge3\). Both actions of
the pair are quadratic. The odd-round loss increases \(R\) by one and the
cumulative shifted loss by \(1-\alpha_{2k-1}\le1\). Since
\(q'(u)=1/(3q(u))\) and \(q(u)\ge\sqrt2\), the temperature increases by
\(0\le\Delta q\le1/(3\sqrt2)\). At the old radius
\(\alpha=\alpha_{2k-1}\le1/2\), the first-order expression
\(q\{-\log(1-\alpha)+\alpha\}\) increases by
\begin{align*}
 \Delta q\,\{-\log(1-\alpha)+\alpha\}
 &\le\frac{\log2+1/2}{3\sqrt2}
 \tag{\(\alpha\le\tfrac12\)}\\
 &<1.
\end{align*}
The input radius increases by one, and the expression increases in
\(\alpha\), so \(\alpha_{2k}\ge\alpha_{2k-1}\).
\end{proof}

\begin{proof}[Proof of Proposition~\ref{prop:standalone-separation}\ref{item:nonradial-separation}]
Write
\(\theta_t=\xi_t\eta\), where \(\xi_t=1\) at odd rounds and
\(\xi_t=-1/2\) at even rounds. The positive cumulative coefficient is
\[
 R_n=\sum_{t\le n}\xi_t=n/4+(3/4)\one\{n\text{ odd}\}.
\]
Thus \(h(Y_n)=R_n\ge n/4\), FTL plays \(-\nabla h(\eta)\) after its initial
zero action, and its regret is exactly one at every horizon.

For standalone FTRL, Lemma~\ref{lem:collinear-bias} applies with
\(\rho=1/4\), \(m=3/4\), and \(\gamma=1/2\); its coefficient is
\(m_p\gamma/(2\rho\bar F_2)=m_p/\bar F_2\), giving the stated limit.
Lemma~\ref{lem:support-curvature} makes it positive because the positive
coordinates of \(\eta\) are unequal.
Near \(p=2\), \(\nabla h(\eta)-\eta=O(p-2)\) and \(\nabla^2h(\eta)\) is
bounded. Therefore
\begin{align*}
 \nabla h(\eta)^\top\nabla^2h(\eta)\nabla h(\eta)
 &=(\nabla h(\eta)-\eta)^\top \nabla^2h(\eta)(\nabla h(\eta)-\eta)
 \tag{\(\nabla^2h(\eta)\eta=0\)}\\
 &=O((p-2)^2).
 \tag{\(\nabla h(\eta)-\eta=O(p-2)\)}
\end{align*}
The logarithmic coefficient vanishes at the Euclidean endpoint; the
bounded comparison is not uniform in the horizon as \(p\downarrow2\).
\end{proof}

\subsection{Switching regret in terms of slack}
\label{app:single-safe}

\subsubsection{Running maximum and comparison with FTL}

\begin{proof}[Proof of Lemma~\ref{lem:single-safe}]
\emph{Controlling the running maximum.}
Immediately before every FTRL round \(t\),
\(2(c+M_{t-1})\le A_{t-1}\).
Indeed, on entry from FTL the stronger test
\(A_{t-1}>4(c+M_{t-1})\) has just passed.
If the preceding round was already FTRL, its return test failed, giving
the required non-strict inequality.

On an FTRL round,
\begin{align*}
 Q_t-Q_{t-1}
 &=h(Y_t)-h(Y_{t-1})+\ip{x_t}{\theta_t}\\
 &\le h(\theta_t)+h(\theta_t)\le2.
 \tag{triangle inequality, \(x_t\in\mathcal K\)}
\end{align*}
Since \(Q_{t-1}\le M_{t-1}\),
\begin{align*}
 M_t=\max\{M_{t-1},Q_t\}
 &\le M_{t-1}+2\\
 &\le A_{t-1}/2+2-c
 \tag{\(2(c+M_{t-1})\le A_{t-1}\)}\\
 &=A_t/2+2-c.
 \tag{\(A\) fixed on FTRL rounds}
\end{align*}
On FTL rounds, \(M\) is unchanged and \(A\) is nondecreasing, so this
bound persists. It also holds initially, since \(A_0=M_0=0\) and
\(1<c<2\). Thus, at every horizon,
\[
 Q_T\le M_T\le A_T/2+2-c.
\]

\emph{Part~\ref{item:switch-counter}: the shifted loss of the FTRL rounds.}
On FTRL rounds, \eqref{eq:ftl-increments} reads
\(\ell^+(x_t,\theta_t)=(Q_t-Q_{t-1})+(\Delta_t-\Delta_{t-1})\), and
\(Q\) changes only on FTRL rounds. Summing over those rounds,
\begin{align*}
 U_T
 &=Q_T+\sum_{\substack{t\le T\\t\text{ is FTRL}}}(\Delta_t-\Delta_{t-1})\\
 &\le Q_T+\Delta_T
 \tag{slack increments are nonnegative}\\
 &\le\Delta_T+A_T/2+2-c
 \tag{bound on \(M_T\)}\\
 &\le\Delta_T+\Reg_T^{\FTL}/2+1.
 \tag{\(A_T\le\Reg_T^{\FTL}\), \(2-c<1\)}
\end{align*}

\emph{Part~\ref{item:switch-comparison}: comparison with FTL.}
If there are no FTRL rounds, the learner plays FTL throughout and
\(\Reg_T=\Reg_T^{\FTL}\ge0\), which proves~\ref{item:switch-comparison}.
Otherwise, the first entry into FTRL requires
\(A>4(c+M)\ge4c\). Since \(A\) never decreases,
\(A_T>4c\), so \(2-c<1<A_T/2\).
Hence
\begin{align*}
 \Reg_T=A_T+Q_T
 &\le\tfrac32A_T+2-c
 \tag{bound on \(M_T\)}\\
 &\le2A_T
 \tag{\(2-c<A_T/2\)}\\
 &\le2\Reg_T^{\FTL}.
 \tag{\(A_T\le\Reg_T^{\FTL}\)}
\end{align*}
This proves~\ref{item:switch-comparison}.

\emph{Part~\ref{item:switch-rate}: the tuned regret bound.}
Lemma~\ref{lem:calibrated-cost} and its coefficient estimates in
Appendix~\ref{app:switch-constants} give part~\ref{item:switch-rate}.
\end{proof}

\subsubsection{Controlling the FTRL blocks}

The switching tests convert a bound on each FTRL block into a bound on
the final running maximum.

\begin{lemma}[Switching from a block bound]\label{lem:switch-accounting}
Run the switching tests of Algorithm~\ref{alg:lp-hybrid} with arbitrary
feasible FTRL actions. Let \(E_t,\mathcal C_t\) be nonnegative
nondecreasing sequences, with \(E_0=0\) and \(\mathcal C_t\ge c\).
Suppose every prefix \([a,b]\) of an FTRL block satisfies
\begin{equation}
 Q_b-Q_{a-1}\le E_b-E_{a-1}+\mathcal C_b.
 \label{eq:switch-block-certificate}
\end{equation}
Then, at every horizon,
\begin{equation}
 c+M_T\le2E_T+3\mathcal C_T,
 \label{eq:block-record}
\end{equation}
and \(\Reg_T\le5(2E_T+3\mathcal C_T)+2-c\).
\end{lemma}

\begin{proof}
\emph{FTL balance.} Before each FTL round, \(A\le4(c+M)\):
initially \(A=0\), after a return \(A<2(c+M)\), and otherwise the
entry test failed. The round increases \(A\) by at most two, and \(M\)
is fixed. During FTRL rounds \(A\) is fixed and \(M\) cannot decrease.
Hence, at every horizon, \(A_T\le4(c+M_T)+2\), and
\begin{align}
 \Reg_T=A_T+Q_T
 &\le A_T+M_T
 \tag{\(Q_T\le M_T\)}\\
 &\le4(c+M_T)+2+M_T\notag\\
 &=5(c+M_T)+2-c.
 \label{eq:single-safe-balance}
\end{align}

\emph{Growth of the running maximum.} A block ends only when \(Q\) attains a new maximum:
\(A\) is fixed during the block, so the return test can first pass only
when \(M\) increases; then \(Q=M\). Since FTL rounds leave \(Q,M\)
unchanged, every block starts at its running maximum,
\(Q_{a-1}=M_{a-1}\), including the first block. For every
\(a\le b'\le b\) in a block \([a,b]\),
\begin{align*}
 Q_{b'}
 &\le M_{a-1}+E_{b'}-E_{a-1}+\mathcal C_{b'}
 \tag{by \eqref{eq:switch-block-certificate}, \(Q_{a-1}=M_{a-1}\)}\\
 &\le M_{a-1}+E_b-E_{a-1}+\mathcal C_b,
 \tag{monotonicity}
\end{align*}
so \(M_b-M_{a-1}\le E_b-E_{a-1}+\mathcal C_b\).
If the block ends at \(b\), the entry and return tests give
\(2(c+M_b)>A>4(c+M_{a-1})\), hence
\begin{align*}
 c+M_b
 &<2(M_b-M_{a-1})
 \tag{\(c+M_b>2(c+M_{a-1})\)}\\
 &\le2[E_b-E_{a-1}+\mathcal C_b].
\end{align*}

\emph{The final running maximum.} Let \(b\) be the latest return, if any.
Any later growth of the running maximum comes from one unfinished block
starting after \(b\). Hence
\begin{align*}
 c+M_T
 &\le c+M_b+E_T-E_b+\mathcal C_T
 \tag{unfinished block}\\
 &\le2E_b+2\mathcal C_b+E_T-E_b+\mathcal C_T
 \tag{completed block, \(E_{a-1}\ge0\)}\\
 &\le2E_T+3\mathcal C_T.
 \tag{monotonicity}
\end{align*}
Before the first return, at most one FTRL block has occurred, so
\(c+M_T\le c+E_T+\mathcal C_T\), which implies the same bound since
\(\mathcal C_T\ge c\). Substituting in \eqref{eq:single-safe-balance}
proves the regret bound.
\end{proof}

\begin{proof}[Proof of Proposition~\ref{prop:tuning}(ii)]
Use \(A_t,Q_t,M_t,U_t\) from Section~\ref{sec:lp-switching}, \(E_t\) from
Appendix~\ref{app:one-clock}, and the nondecreasing sequence
\(\mathcal C_t=\mathcal C(U_t)\ge c\). Recall that
\(E_t\le\Upsilon W_p(U_t,d_*)\).

The block bound \eqref{eq:potential-block-budget} therefore gives
\begin{equation}
 Q_b-Q_{a-1}\le E_b-E_{a-1}+\mathcal C(U_b)
 \label{eq:observed-block-bound}
\end{equation}
for every FTRL-block prefix \([a,b]\). Lemma~\ref{lem:switch-accounting}
now proves (ii).
\end{proof}

\emph{Zero slack.} If \(\Delta_T=0\), equality holds in the triangle inequality for the
strictly convex norm \(\norm{\cdot}_r\), so all nonzero losses up to round
\(T\) are positive multiples of one direction. FTL then pays only at its
first nonzero loss, played at \(Y=0\), so
\(A_t\le\Reg_t^{\FTL}\le1<4c\) for \(t\le T\): the entry test never
passes, the learner remains in FTL, and \(\Reg_T=\Reg_T^{\FTL}\le1\).

\subsubsection{Constants and the slack bound}
\label{app:switch-constants}

With \(\Upsilon\) from Proposition~\ref{prop:tuning}, the additive term has coefficient
\(A_{\mathrm{sw}}=2-c+15(c+1/\sqrt2)\),
and the profile term has coefficient
\(B_{\mathrm{sw}}=5[2\Upsilon+3(1+1/\sqrt2)]\).

\begin{lemma}[Explicit switching bound]\label{lem:calibrated-cost}
Algorithm~\ref{alg:lp-hybrid} satisfies
\(\Reg_T\le A_{\mathrm{sw}}+B_{\mathrm{sw}}W_p(U_T,d_*)\).
\end{lemma}

\begin{proof}
Expanding \(\mathcal C(U_T)\) in Proposition~\ref{prop:tuning}(ii),
\begin{align*}
 \Reg_T&\le5
  [2\Upsilon W_p(U_T,d_*)+3\mathcal C(U_T)]+2-c
 \tag{Proposition~\ref{prop:tuning}(ii)}\\
 &=A_{\mathrm{sw}}+B_{\mathrm{sw}}W_p(U_T,d_*).
 \tag{definition of \(\mathcal C\)}
\end{align*}
\end{proof}

\paragraph{Uniform constants.}
For \(p\ge3\), \(\bar\kappa_p/m_p=1\); for \(2\le p<3\),
\(\bar\kappa_p/m_p=\bar\kappa_p^{3-p}\le p<3\).
Also \(3^{1/2-1/p}\le\sqrt3\).
To bound the radial-scale factor, we show
\(\varepsilon_p^{-1/(p-1)}\le e^{1/8}\).
This is immediate for \(p\le e\), when \(\varepsilon_p=1\).
For \(p>e\), it is equivalent to \(\log\log p\le(p-1)/8\).
The difference is concave and maximised where \(p\log p=8\), at a point
in \((4.9,5)\); there it is below \(\log\log5-3.9/8<0\).

Consequently,
\(\Upsilon\le\max\{5\sqrt3\,e^{1/8},3\sqrt6\}+3=5\sqrt3\,e^{1/8}+3\),
including the crossover charge. In the explicit coefficients,
\begin{align*}
 A_{\mathrm{sw}}
 &=2+14c+15/\sqrt2\\
 &<5\bigl(6+3/\sqrt2\bigr)
 \tag{\(1<c<2\)}\\
 &<45,\\
 B_{\mathrm{sw}}
 &\le5\bigl(10\sqrt3\,e^{1/8}+9+3/\sqrt2\bigr)
 \tag{bound on \(\Upsilon\)}\\
 &<155.
\end{align*}
This is claim~\ref{item:switch-rate}.

\paragraph{Passing from the FTRL rounds' shifted loss to slack.}
Put \(N_p=1+C_p^{\FTL}/2\). Then
\begin{align*}
 U_T
 &\le\Delta_T+\Reg_T^{\FTL}/2+1
 \tag{part~\ref{item:switch-counter}}\\
 &\le\Delta_T+C_p^{\FTL}(1+\Delta_T)/2+1
 \tag{Proposition~\ref{prop:ftl-log-p}}\\
 &\le N_p(1+\Delta_T).
\end{align*}
By \eqref{eq:calibrated-profile}, branch by branch,
\begin{equation}
 W_p(qv,d_*)\le\max\{q^{1/r},\sqrt{m_pq}\}\,W_p(v,d),\qquad q,v\ge0.
 \label{eq:profile-scaling}
\end{equation}
Also \(W_p(1+n,d)\le1+W_p(n,d)\), since \(W_p(\cdot,d)\) is concave
with \(W_p(0,d)=0\) and \(W_p(1,d)=1\): its increment over an interval
of length one is largest at zero. Therefore
\begin{align*}
 W_p(U_T,d_*)
 &\le W_p(N_p(1+\Delta_T),d_*)
 \tag{monotonicity}\\
 &\le\max\{N_p^{1/r},\sqrt{m_pN_p}\}\,W_p(1+\Delta_T,d)
 \tag{by \eqref{eq:profile-scaling}}\\
 &\le\max\{N_p^{1/r},\sqrt{m_pN_p}\}\,[1+W_p(\Delta_T,d)].
 \tag{concavity}
\end{align*}
Since \(A_{\mathrm{sw}}<45\) and \(B_{\mathrm{sw}}<155\),
claim~\ref{item:switch-rate} gives the switching guarantees in
Theorem~\ref{thm:main} with coefficient
\(155(1+\max\{N_p^{1/r},\sqrt{m_pN_p}\})\).
At \(p=2\), the crossover charge vanishes, \(m_2=1\), and
\(W_2(u,d)=W_2(u,d_*)=\sqrt u\). Also \(\Upsilon=5\), since
\(\sqrt6\,\bar\kappa_2<5\). This gives
\(A_{\mathrm{sw}}<32.5\) and \(B_{\mathrm{sw}}<75.7\); also
\(C_2^{\FTL}<3.55\). Hence
\begin{align*}
 \Reg_T
 &\le A_{\mathrm{sw}}+B_{\mathrm{sw}}\sqrt{(1+C_2^{\FTL}/2)(1+\Delta_T)}
 \tag{Lemma~\ref{lem:calibrated-cost}, bound on \(U_T\)}\\
 &\le\Bigl(A_{\mathrm{sw}}+B_{\mathrm{sw}}\sqrt{1+C_2^{\FTL}/2}\Bigr)
   \bigl[1+\sqrt{\Delta_T}\bigr]
 \tag{\(\sqrt{1+\Delta}\le1+\sqrt\Delta\)}\\
 &<159\,[1+W_2(\Delta_T,d)].
 \tag{\(32.5+75.7\sqrt{2.775}<159\)}
\end{align*}
This proves the stated endpoint coefficient.
Here \(N_p=O(p/\log p)\) (Proposition~\ref{prop:ftl-log-p}) and
\(m_p\le\bar\kappa_p=O(p/\log p)\), so the coefficient is
\(O(p/\log p)\). Appendix~\ref{app:calibrated-standalone} sharpens it to
\((2+o(1))p/\log p\).

\subsection{Loss normalisation and the FTL comparison}
\label{app:unknown-width}%
The results assume \(h(\theta_t)\le1\). For a known bound
\(h(\theta_t)\le\bar h\), \(\bar h>0\), run either learner on \(\theta_t/\bar h\). The
main rate becomes
\(O_p(\bar h+\min\{\bar h^{1/p}\Delta_T^{1/r},d^{1/2-1/p}\sqrt{\bar h\Delta_T}\})\), and
the FTL factor is unchanged. Without such a bound,
Proposition~\ref{prop:unknown-width} shows that no deterministic learner
can both compete with FTL and attain the sharp rate.

\begin{proposition}[Incompatibility without a loss norm bound]\label{prop:unknown-width}
If a deterministic causal learner satisfies
\(\Reg_T\le K\Reg_T^{\FTL}\), for a fixed finite \(K\), on every
finite loss sequence of arbitrary norms, then it plays \(f(Y_t)\) after every
realisable history with \(Y_t\ne0\), and it does not attain
\eqref{eq:explicit} on all paths with loss norms at most one.
\end{proposition}

\begin{proof}
If \(x\ne f(Y)\) is played at \(Y\ne0\), then \(\ip{x}{Y}<h(Y)\), as the face
exposed by \(Y\) is a singleton. The loss \(\theta=-LY/h(Y)\) keeps the state
on its ray. FTL's regret increment is zero, whereas the learner's is
\(L(1-\ip{x}{Y}/h(Y))>0\). The past regrets are fixed, so letting
\(L\to\infty\) contradicts the assumed comparison.

For the second claim, take oblivious pairs
\(\sigma_je_1,-\sigma_je_1\) with independent uniform signs. Each pair
returns to \(Y=0\). The first loss of pair \(j\) is independent of the
learner's action, and at the second the learner must play
\(f(-\sigma_je_1)=-\sigma_je_1\). Hence, at even \(T\),
\begin{align*}
 \E\Reg_T
 &=\sum_{j\le T/2}\bigl(\E\ip{x_{2j-1}}{\sigma_je_1}
   +\ip{-\sigma_je_1}{-\sigma_je_1}\bigr)+h(Y_T)\\
 &=T/2,
 \tag{\(\E[\sigma_j\mid x_{2j-1}]=0\), \(Y_T=0\)}
\end{align*}
whereas \(\Delta_T=T\) and \(W_p(T,d)=d^{1/2-1/p}\sqrt T\) for \(T\ge d\).
Some sign sequence therefore has \(\Reg_T\ge T/2\), which exceeds every
bound of the form \eqref{eq:explicit} for large \(T\).
\end{proof}

\section{Minimax bounds and the optimal leading constant}
\label{app:p-dependence}

We prove Proposition~\ref{prop:minimax}, Theorem~\ref{thm:universal-lower},
and the coefficients in Theorem~\ref{thm:main}; the last also use a
refinement of the FTL bound for small slack. Throughout, \(m_p\) and \(d_*\) are as in
\eqref{eq:calibrated-profile}.

The lower bounds use the martingale-difference principle of
\citet[Theorem~3]{sridharan2010convex}: conditionally mean-zero losses
give zero expected learner loss, leaving the expected dual norm of
their sum as regret. The constructions below control slack as well.

\subsection{Minimax regret}
\label{app:minimax}

The upper bound follows from the pathwise guarantee. For the lower
bound, random signs on coordinate losses give zero expected learner
loss; their distribution across coordinates produces the two branches
of \(W_p\). We adapt the coordinate-sign constructions in
\citet[Appendices~C.3--C.4]{johnson2025necessity} to simultaneous constraints on mass
and slack.

\begin{proof}[Proof of Proposition~\ref{prop:minimax}]
Put \(m=\min\{\bar H,T\}\) and \(n=\min\{\bar H,\bar\Delta,T\}\).
If \(m<1\), then \(m=\bar H\), since \(T\ge1\) is an integer.
Playing zero gives regret \(h(Y_T)\le H_T\le m\); one random-sign
coordinate loss of mass \(m\), followed by zeros, has slack zero and forces
expected regret \(m\). This proves the exact value, including \(m=0\).
We henceforth assume \(m\ge1\).

\emph{Upper bound.}
Every admissible path has \(\Delta_T\le\min\{H_T,\bar\Delta\}\le n\).
Either algorithm in Theorem~\ref{thm:main} gives
\({O_p(1+W_p(n,d))}\).

\emph{Lower bound.}
One unit coordinate loss with an independent uniform sign, followed by
zeros, forces expected regret one with slack zero. This suffices if
\(n<1\), when \(W_p(n,d)\le1\).
For \(n\ge1\), play \(j=\lfloor n\rfloor\ge n/2\) unit coordinate losses
with independent uniform signs, followed by zeros. Use distinct coordinates
if \(j\le d\); otherwise distribute them evenly, giving each coordinate
at least \(\lfloor j/d\rfloor\ge j/(2d)\) signs.
Every realisation has mass \(j\le m\) and slack at most \(j\le\bar\Delta\).
Independence of each sign from the learner's randomisation and past losses
gives expected learner loss zero, so expected regret is \(\E h(Y_T)\).

If \(j\le d\), then \(h(Y_T)=j^{1/r}\). If \(j>d\), each coordinate is
a sum \(X\) of independent signs with \(\E X^2\ge j/(2d)\), and
\begin{align*}
 \E|X|
 &\ge\frac{(\E X^2)^{3/2}}{(\E X^4)^{1/2}}
 \tag{H\"older}\\
 &\ge\sqrt{\frac{\E X^2}{3}}.
 \tag{\(\E X^4\le3(\E X^2)^2\)}
\end{align*}
Hence
\begin{align*}
 \E h(Y_T)
 &\ge d^{1/r-1}\sum_{i\le d}\E|(Y_T)_i|
 \tag{\(h\ge d^{1/r-1}\norm\cdot_1\)}\\
 &\ge d^{1/r}\sqrt{j/(6d)}=d^{1/2-1/p}\sqrt{j/6}.
 \tag{\(\E X^2\ge j/(2d)\)}
\end{align*}
In both cases, concavity of \(W_p(\cdot,d)\) and \(j\ge n/2\) give
\(\E h(Y_T)\ge W_p(j,d)/\sqrt6\ge W_p(n,d)/(2\sqrt6)\).
The two admissible constructions therefore give
\begin{align*}
 \inf_{\mathsf L}\sup_{\mathsf A\in\mathcal A}\E_{\mathsf L,\mathsf A}[\Reg_T]
 &\ge\max\{1,W_p(n,d)/(2\sqrt6)\}\\
 &\ge\frac{1+W_p(n,d)}{4\sqrt6}.
 \tag{maximum at least the average}
\end{align*}
\end{proof}

\subsection{Lower bounds on the dependence on \texorpdfstring{\(p\)}{p}}
\label{app:power-dependence}

Recall \(C_{\mathsf L}(p)\) from Section~\ref{sec:olo-setting}.
We construct unit losses with conditional mean zero and small expected
slack to prove Theorem~\ref{thm:universal-lower}.

\subsubsection{The finite-horizon construction}

For \(T\ge2\), set \(N_0=T+\lceil(pT(T+1))^2\rceil\) and
\(\pi_0=1/(1+N_0^{1/p})\). Define the slack bound
\[
 \mathcal U_p(T)=\frac Tp\Bigl[\tfrac12\log\bigl(1+2/\pi_0\bigr)
   +(1-\pi_0)\bigl(1+\log(1+T^r)\bigr)\Bigr]+\frac3p.
\]

\begin{proposition}[Finite-horizon lower bound]\label{prop:universal-finite}
Let \(p>2\), \(T\ge2\), and \(d\ge N_0\). Suppose a causal learner, which
may depend on \(p\), \(d\), and \(T\), satisfies
\(\E\Reg_{T+1}\le C[1+W_p(\Delta_{T+1},d)]\) for every deterministic
loss sequence of length \(T+1\) with \(h(\theta_t)\le1\), the expectation
being over its randomisation. Then
\[
 C\ge\frac{T+1-\mathcal U_p(T)}{1+\mathcal U_p(T)^{1/r}}.
\]
\end{proposition}

\paragraph{The push--reverse construction.}
Unreversed coordinates share a small amplitude. A push increases that
amplitude; a reversal moves one coordinate near unit magnitude and
removes it from future losses. The probabilities make the next loss
have conditional mean zero.

The losses use only the first \(N_0\) coordinates. Draw independent
uniform signs \(\sigma_1,\ldots,\sigma_{N_0}\) and, independently,
fresh coins at every round. Round one plays
\(\theta_1=-N_0^{-1/r}\sum_{j\le N_0}\sigma_je_j\). Afterwards each
coordinate \(j\le N_0\) is \emph{thin} or \emph{fat}. All are thin after
round one, and the thin coordinates of \(Y_t\) equal \(\mu_t\sigma_j\)
for a common \(\mu_t>0\), with \(\mu_1=N_0^{-1/r}\). At round
\(t\in\{2,\ldots,T+1\}\), let \(\mathcal S\) be the thin set,
\(\mathsf N=|\mathcal S|\), \(\mu=\mu_{t-1}\), and \(\pi=1/(1+\mathsf N^{1/p})\). With
probability \(\pi\) the round is a \emph{push},
\[
 \theta_t=-\mathsf N^{-1/r}\sum_{j\in\mathcal S}\sigma_je_j
 =-\frac{Y_{t-1,\mathcal S}}{h(Y_{t-1,\mathcal S})},
\]
which raises \(\mu\) by \(\mathsf N^{-1/r}\); here \(Y_{\mathcal S}\) is the
restriction of \(Y\) to \(\mathcal S\). Otherwise the round is a
\emph{reversal}: an index \(j\) uniform on \(\mathcal S\) is drawn,
\(\theta_t=\sigma_je_j\), and \(j\) becomes fat with
\(Y_{t,j}=-(1-\mu)\sigma_j\). Fat coordinates receive no further loss.
Every loss has \(h(\theta_t)=1\), so \(H_t=t\), and no loss depends on
the learner's actions.

\subsubsection{Expected slack and the finite-horizon bound}

\begin{proof}[Proof of Proposition~\ref{prop:universal-finite}]
We show that the losses have conditional mean zero and
\(\E\Delta_{T+1}\le\mathcal U_p(T)\), then apply the assumed regret bound.
At round \(t\ge2\), write \(R=h(Y_{t-1})\), and let \(R_f\) and
\(R_s=\mathsf N^{1/r}\mu\) be the norms on the fat and thin coordinates.
Thus \(R^r=R_f^r+R_s^r\), and the slack increment is
\(\delta_t=1-[h(Y_t)-R]\), with \(\delta_1=0\).

\emph{Sizes.}
Before any round \(t\le T+1\), at most \(T-1\) reversals have occurred,
so \(\mathsf N\ge N_0-T+1\) and \(R\le H_{t-1}\le T\).
The initial amplitude and each push increment are at most
\((N_0-T)^{-1/r}\). Hence, for \(t\le T+1\),
\begin{align*}
 \mu_t
 &\le(T+1)(N_0-T)^{-1/r}
 \tag{at most \(T\) pushes}\\
 &\le\frac{T+1}{pT(T+1)}=\frac1{pT}.
 \tag{\(N_0-T\ge[pT(T+1)]^2\), \(2/r\ge1\)}
\end{align*}

\emph{Conditional mean.}
Let \(\mathcal G_{t-1}\) be generated by the learner's randomisation and
the losses up to round \(t-1\). For \(t\ge2\), the thin set and its signs are
known given \(\mathcal G_{t-1}\), since the first loss reveals all signs.
The fresh push and reversal coins give
\begin{align*}
 \E[\theta_t\mid\mathcal G_{t-1}]
 &=\left(-\pi\mathsf N^{-1/r}+\frac{1-\pi}{\mathsf N}\right)
   \sum_{j\in\mathcal S}\sigma_je_j\\
 &=0.
 \tag{\(1-\pi=\pi\mathsf N^{1/p}\), \(1/p-1=-1/r\)}
\end{align*}
Round one also has mean zero, independently of the learner's first action.
Every causal learner therefore has expected cumulative loss zero, and
\[
 \E\Reg_{T+1}=\E h(Y_{T+1})=T+1-\E\Delta_{T+1}.
\]

\emph{Slack of a push.}
A push leaves \(R_f\) unchanged and raises \(R_s\) to \(R_s+1\).
The map \(u\mapsto(R_f^r+u^r)^{1/r}\) is convex with derivative
\((u/R)^{r-1}\) at \(u=R_s\), so \(h(Y_t)-R\ge(R_s/R)^{r-1}\). Hence
\begin{align}
 \delta_t
 &\le1-(R_s/R)^{r-1}\notag\\
 &=1-\exp\Bigl\{-\frac1p\log\Bigl(1+\frac{R_f^r}{R_s^r}\Bigr)\Bigr\}
 \tag{\((r-1)/r=1/p\)}\\
 &\le\frac1p\log\Bigl(1+\frac{R_f^r}{R_s^r}\Bigr).
 \label{eq:universal-push}
\end{align}
The last step uses \(1-e^{-x}\le x\).

\emph{Slack of a reversal.}
A reversal gives \(h(Y_t)^r=R^r+b\), where
\(b=(1-\mu)^r-\mu^r\). Bernoulli's inequality and \(\mu^r\le\mu\) give
\(1-3\mu\le b\le1\). The size bound gives \(3\mu\le3/(pT)<1\). Hence
\begin{align*}
 h(Y_t)-R
 &\ge\frac br(R^r+b)^{-1/p}
 \tag{concavity of \(u\mapsto u^{1/r}\)}\\
 &\ge\frac{1-3\mu}{r}(1+R^r)^{-1/p}.
 \tag{\(1-3\mu\le b\le1\)}
\end{align*}
Put \(x=\log(1+R^r)/p\). Then
\begin{align}
 \delta_t
 &\le1-\frac{1-3\mu}{r}e^{-x}\notag\\
 &\le(1-e^{-x})+\frac{e^{-x}}p+3\mu
 \tag{\(1/r=1-1/p\)}\\
 &\le\frac{1+\log(1+R^r)}p+3\mu
 \tag{\(1-e^{-x}\le x\)}\\
 &\le\frac{1+\log(1+T^r)}p+3\mu.
 \label{eq:universal-reversal}
\end{align}
The last step uses \(R\le T\).

\emph{Expected push slack.}
Fix \(t\ge2\). Of the \(q=t-2\) rounds after the first and before \(t\),
let \(q_+\) be pushes. Fat coordinates have modulus at most one, so
\(R_f^r\le q-q_+\le q\).
Initially \(R_s^r=1\); each push raises it by at least one, since
\((R_s+1)^r\ge R_s^r+1\), and each reversal lowers it by
\(\mu^r\le1/(pT)\). Hence
\begin{align*}
 R_s^r
 &\ge1+q_+-\frac q{pT}\\
 &\ge\frac{1+q_+}2,
 \tag{\(q<T\), \(p>2\)}
\end{align*}
and \(R_f^r/R_s^r\le2q/(1+q_+)\).
The conditional push probability is at most \(1/2\), and the push
decision at round \(t\) is independent of \(q_+\). Thus
\begin{align*}
 \E[\delta_t\one\{\text{push}\}]
 &\le\frac1{2p}\E\log\Bigl(1+\frac{2q}{1+q_+}\Bigr)
 \tag{by \eqref{eq:universal-push}}\\
 &\le\frac1{2p}\log\Bigl(1+2q\,\E\frac1{1+q_+}\Bigr).
 \tag{Jensen}
\end{align*}
Every push probability is at least \(\pi_0\), since \(\mathsf N\le N_0\).
Generate pushes by comparing independent uniform coins with their current
probabilities. The number of coins below \(\pi_0\) is a binomial variable
\(B\) with parameters \(q,\pi_0\), and is at most \(q_+\). Therefore
\begin{align*}
 \E\frac1{1+q_+}
 &\le\E\frac1{1+B}
 \tag{\(B\le q_+\)}\\
 &=\int_0^1(1-\pi_0+\pi_0v)^q\dif v
 \tag{\(\tfrac1{1+B}=\int_0^1v^B\dif v\)}\\
 &\le\frac1{(q+1)\pi_0}.
\end{align*}
Substitution gives
\(\E[\delta_t\one\{\text{push}\}]\le(2p)^{-1}\log(1+2/\pi_0)\).

\emph{Total slack and the regret bound.}
A reversal has conditional probability at most \(1-\pi_0\).
By \eqref{eq:universal-reversal} and \(\mu\le1/(pT)\),
\[
 \E[\delta_t\one\{\text{reversal}\}]
 \le\frac{(1-\pi_0)[1+\log(1+T^r)]}{p}+\frac3{pT}.
\]
Add the expected push slack and sum over \(2\le t\le T+1\).
Since \(\delta_1=0\), this gives \(\E\Delta_{T+1}\le\mathcal U_p(T)\).

Conditioning on all the adversary's signs and coins leaves a deterministic
loss sequence independent of the learner's randomisation. We may therefore
average the assumed regret bound over these sequences.
Every path has \(\Delta_{T+1}\le T+1\le d\), so
\(W_p(\Delta_{T+1},d)=\Delta_{T+1}^{1/r}\).
Here \(C\ge0\), as the guarantee on the zero loss sequence requires. Hence
\begin{align*}
 T+1-\mathcal U_p(T)
 &\le T+1-\E\Delta_{T+1}
 \tag{\(\E\Delta_{T+1}\le\mathcal U_p(T)\)}\\
 &=\E\Reg_{T+1}
 \tag{conditional mean zero}\\
 &\le C[1+\E\Delta_{T+1}^{1/r}]
 \tag{assumed guarantee}\\
 &\le C[1+\mathcal U_p(T)^{1/r}].
 \tag{Jensen}
\end{align*}
Dividing proves the claim.
\end{proof}

\subsubsection{Consequences for the coefficient}

\begin{proof}[Proof of Theorem~\ref{thm:universal-lower}]
Take a horizon of order \(p\) in Proposition~\ref{prop:universal-finite},
namely \(T=\lceil2p\rceil\). The bounds \(T\le5p/2\) and \(T+1\le3p\) give
\(pT(T+1)\le8p^3\), hence \(N_0\le65p^6\).
Thus \(N_0^{1/p}=1+O(\log p/p)\) and
\(\pi_0=\tfrac12+O(\log p/p)\), giving
\(\log(1+2/\pi_0)=\log5+O(\log p/p)\).
Also, \(\log(1+T^r)=\log p+O(1)\). Hence
\begin{align*}
 \mathcal U_p(T)
 &=(2+O(1/p))\bigl[\tfrac12\log5+(\tfrac12+o(1))(\log p+O(1))\bigr]+3/p
 \tag{\(T/p=2+O(1/p)\)}\\
 &=\log p+O(1).
\end{align*}
For all sufficiently large \(p\), \(2p>\mathcal U_p(T)\), and
\begin{align*}
 C
 &\ge\frac{T+1-\mathcal U_p(T)}{1+\mathcal U_p(T)^{1/r}}
 \tag{Proposition~\ref{prop:universal-finite}}\\
 &\ge\frac{2p-\mathcal U_p(T)}{2+\mathcal U_p(T)}
 \tag{\(T+1\ge2p\), \(\mathcal U_p(T)^{1/r}\le1+\mathcal U_p(T)\)}\\
 &=\bigl[2-O(1/\log p)\bigr]\frac p{\log p}.
 \tag{\(\mathcal U_p(T)=\log p+O(1)\)}
\end{align*}
This applies to every learner at \(d=N_0\) and horizon \(T+1\), proving the
first two claims. The upper bounds of Appendix~\ref{app:calibrated-standalone}
give
\(\inf_{\mathsf L}C_{\mathsf L}(p)\le(2+o(1))p/\log p\), which proves the limit.
\end{proof}

\subsection{An FTL refinement for small slack}
\label{app:ftl-low-dim}

We use the bounded potential
\(\Xi(y)=2\sum_i\varpi(y_i)\), where \(\varpi(x)=\min\{|x|,1\}^r\), to
improve the FTL bound below the slack scale \(e^{\sqrt p}\).
The scalar inequality below telescopes coordinatewise, costing at most
\(2d\) in Corollary~\ref{cor:ftl-low-dim}.
Put \(\nu=1/(p-1)\) and \(\varphi(x)=|x|^r/r\).

\begin{lemma}[A scalar potential inequality]\label{lem:scalar-potential}
Let \(p\ge21\), \(u,v\in[-1,1]\), \(R\ge1\), and \(0<s\le1\) with
\((R+s)^\nu\le11/10\). Then
\begin{equation}
 17(R+s)D_\varphi\Bigl(\frac{Ru+sv}{R+s},u\Bigr)
 -16sD_\varphi(v,u)\le2[\varpi(Ru+sv)-\varpi(Ru)].
 \label{eq:scalar-potential}
\end{equation}
\end{lemma}

\begin{proof}
By evenness we may assume \(u\ge0\), and \(v\ge0\) if \(u=0\).
Let \(E\) be the left side of \eqref{eq:scalar-potential}, put
\(t=s/(R+s)\le1/2\), and set
\(g(\tau)=(v-u)[\varphi'(u+\tau(v-u))-\varphi'(u)]\).
This derivative of \(D_\varphi(u+\tau(v-u),u)\) is nondecreasing,
with \(g(0)=0\). Since \((Ru+sv)/(R+s)=u+t(v-u)\) and \((R+s)t=s\),
\begin{align}
 \frac Es
 &=\frac{17}t\int_0^tg-16\int_0^1g\notag\\
 &=\frac{17-16t}{t}\int_0^t g-16\int_t^1g
 \label{eq:scalar-potential-split}\\
 &\le(17-16t)g(t)-16(1-t)g(t)=g(t).
 \tag{\(g\) nondecreasing}
\end{align}
We split according to the short segment from \(u\) to \(u+t(v-u)\).

\emph{Increasing: \(v\ge u\).}
If \(v=u\), then \(E=0\) and the potential is nondecreasing, so the
claim is immediate. Assume \(v>u\). Since \(\varphi'(x)=x^\nu\) for
\(x\ge0\),
\begin{align*}
 E
 &\le s\,g(t)
 \tag{by \eqref{eq:scalar-potential-split}}\\
 &\le s(v-u)\{t(v-u)\}^\nu
 \tag{\(|a^\nu-b^\nu|\le|a-b|^\nu\)}\\
 &\le s^r(v-u)^r\le(sv)^r,
 \tag{\(t\le s\), \(u\ge0\)}
\end{align*}
and \(E\le sg(t)\le1\), since \(s\le1\) and \(0<v-u\le1\).
For \(Ru\le1/2\): if \(Ru+sv\le1\), superadditivity of \(x\mapsto x^r\)
gives \(2[\varpi(Ru+sv)-\varpi(Ru)]\ge2(sv)^r\ge E\); otherwise the
potential increase is at least \(1-2^{-r}\ge1/2\ge E/2\).

If \(Ru>1/2\), then
\begin{align*}
 t\Bigl(\frac vu\Bigr)^{1-\nu}
 &\le\frac{sv}{Ru}
 \tag{\(v/u\ge1\), \(t\le s/R\)}\\
 &<2\le4(1-t).
 \tag{\(sv\le1<2Ru\), \(t\le\tfrac12\)}
\end{align*}
Since \(\varphi''(x)=\nu x^{\nu-1}\) is decreasing on \((0,\infty)\),
\begin{align*}
 \frac{g(t)}{g(1)-g(t)}
 &\le\frac{t}{1-t}\Bigl(\frac vu\Bigr)^{1-\nu}
 \tag{mean value theorem}\\
 &\le4.
\end{align*}
Concavity of \(g\) also gives
\(\int_t^1g\ge(1-t)[g(t)+g(1)]/2\). With \(\int_0^tg\le tg(t)\),
\eqref{eq:scalar-potential-split} yields
\begin{align*}
 \frac Es
 &\le(17-16t)g(t)-8(1-t)[g(t)+g(1)]\\
 &=g(t)-8(1-t)[g(1)-g(t)]\\
 &\le0.
 \tag{\(8(1-t)\ge4\)}
\end{align*}
The potential is nondecreasing here, so the claim follows.

\emph{Decreasing without crossing zero: \(v<u\), \(Ru+sv\ge0\).}
On \([0,t]\), \(u+\tau(v-u)\ge0\) and \(\varphi'\) is concave there, so
\(g\) is convex on \([0,t]\). Hence
\begin{align*}
 \frac Es
 &\le\frac{17-16t}2g(t)-16\int_t^1g
 \tag{\(\int_0^tg\le tg(t)/2\)}\\
 &\le9(1-t)g(t)-16\int_t^1g
 \tag{\(t\le\tfrac12\)}\\
 &\le-7\int_t^1g\le0.
 \tag{\(\int_t^1g\ge(1-t)g(t)\)}
\end{align*}
If \(v\ge0\), the potential is nondecreasing and we are done.
If \(v<0\), then \(g(\tau)\ge(u-v)u^\nu\) wherever
\(u+\tau(v-u)\le0\). This set is an interval in \([t,1]\) of length
\(|v|/(u-v)\), so \(\int_t^1g\ge|v|u^\nu\). Therefore
\begin{align*}
 E
 &\le-7s|v|u^\nu\\
 &\le-2r(Ru)^\nu s|v|
 \tag{\(2rR^\nu<7\)}\\
 &\le2[\varpi(Ru+sv)-\varpi(Ru)].
 \tag{\(\varpi'(x)\le rx^\nu\) on \([Ru+sv,Ru]\)}
\end{align*}

\emph{Crossing zero: \(Ru+sv<0\).}
Put \(x=Ru\) and \(y=-(Ru+sv)>0\).
Then \(x+y=s|v|\le1\), so the potential change is \(y^r-x^r\).
Expanding both divergences with \(\varphi'(u)=u^\nu\) and \(s|v|=x+y\),
\[
 rE=\frac{17y^r}{(R+s)^{r-1}}-16s|v|^r+\nu(17R+s)u^r+ru^\nu(y-16x).
\]
Since \(u^\nu\le x^\nu\) and \(u^\nu x=Ru^r\),
\begin{align*}
 rE
 &\le17y^r-16(x+y)^r+rx^\nu y+[\nu s-(16-\nu)R]u^r
 \tag{\(R+s\ge1\), \(s\le1\), \(r>1\)}\\
 &\le17y^r-16(x+y)^r+rx^\nu y
 \tag{\(R\ge1\ge s\), \(\nu\le1/20\)}\\
 &\le2y^r-(16-\nu)x^r
 \tag{superadditivity, Young}\\
 &\le2r(y^r-x^r).
 \tag{\(16-\nu\ge2r\)}
\end{align*}
The third step uses \((x+y)^r\ge x^r+y^r\) and Young's inequality
\(rx^\nu y\le\nu x^r+y^r\).
\end{proof}

\begin{corollary}[FTL below the scale \(e^{\sqrt p}\)]\label{cor:ftl-low-dim}
If \(p\ge130\) and \(\Delta_T\le e^{\sqrt p}\), then
\(\Reg_T^{\FTL}\le4+2d+2\sqrt p\,\Delta_T\).
\end{corollary}

\begin{proof}
Put \(\bar R=e^{\sqrt p}\ge e^{11}\) and \(R_t=h(Y_t)\).
For \(j\le j'\),
\begin{align}
 \Delta_{j'}-\Delta_j
 &=H_{j'}-H_j-(R_{j'}-R_j)\notag\\
 &\ge h(Y_j-Y_{j'})+R_j-R_{j'}
 \tag{triangle inequality}\\
 &\ge2(R_j-R_{j'}).
 \label{eq:gap-radius-drop}
\end{align}
The last step is the reverse triangle inequality.
Let \(j_1\le j_2\) be the last indices in \(\{0,\ldots,T\}\) with
\(R_{j_1}\le1\) and \(R_{j_2}\le\bar R\), respectively, and put
\(i_k=\min\{j_k+1,T\}\). We bound three ranges of rounds.

\emph{Initial rounds: \(t\le i_1\).}
Since \(R_{i_1}\le R_{j_1}+1\le2\),
\begin{align*}
 \Reg_{i_1}^{\FTL}
 &\le H_{i_1}+R_{i_1}
 \tag{feasibility}\\
 &=\Delta_{i_1}+2R_{i_1}\le\Delta_{i_1}+4.
\end{align*}

\emph{Intermediate radii: \(i_1<t\le i_2\).}
Here \(j_1<t-1\le j_2\), so
\begin{align*}
 1<R_{t-1}
 &\le R_{j_2}+\frac{\Delta_{j_2}-\Delta_{t-1}}2
 \tag{by \eqref{eq:gap-radius-drop}}\\
 &\le\bar R+\frac{e^{\sqrt p}}2=\tfrac32\bar R.
 \tag{\(\Delta_T\le e^{\sqrt p}\)}
\end{align*}
The scalar estimate applies because
\begin{align*}
 \nu\log(R_{t-1}+1)
 &\le\frac{\sqrt p+\log1.51}{p-1}
 \tag{\(\tfrac32\bar R+1\le1.51\bar R\)}\\
 &<\log1.1.
 \tag{\(p\ge130\)}
\end{align*}
The fraction decreases with \(p\) and is below \(0.0916\) at \(p=130\).
For a nonzero loss, put \(R=R_{t-1}\), \(s=h(\theta_t)\),
\(\tau=D_h(Y_t,Y_{t-1})\), and \(\delta=\Delta_t-\Delta_{t-1}\).
Apply Lemma~\ref{lem:scalar-potential} with
\((u,v)=((Y_{t-1})_i/R,-(\theta_t)_i/s)\); both coordinates lie in
\([-1,1]\) and \((R+s)^\nu\le(R+1)^\nu\le11/10\).
Write \(u,v,m\) for the vectors of Lemma~\ref{lem:norm-increment} and
\(P=h^r/r=\sum_i\varphi(\cdot_i)\), so that \(Ru+sv=Y_t\). Then
\begin{align*}
 \tau-16\delta
 &=17\tau-16(\tau+\delta)\\
 &\le17(R+s)D_P(m,u)-16sD_P(v,u)
 \tag{Lemma~\ref{lem:norm-increment}\ref{item:norm-increment-regret}, \ref{item:norm-increment-loss}}\\
 &\le\Xi(Y_t)-\Xi(Y_{t-1}).
 \tag{Lemma~\ref{lem:scalar-potential}, summed}
\end{align*}
For a zero loss both sides vanish. Summing over these rounds and using
\(0\le\Xi\le2d\) gives
\[
 \Reg_{i_2}^{\FTL}-\Reg_{i_1}^{\FTL}
 \le16(\Delta_{i_2}-\Delta_{i_1})+2d.
\]

\emph{Large radii: \(t>i_2\).}
Now \(R_{t-1}>\bar R\), so Lemma~\ref{lem:radius-ftl} bounds each FTL regret
increment by \(b_p(\bar R)\) times the corresponding slack increment, with
\begin{align*}
 b_p(\bar R)
 &\le\frac1{e^{\sqrt p/(2(p-1))}-1}
 \tag{\(1+\bar R\ge e^{\sqrt p}\)}\\
 &\le\frac{2(p-1)}{\sqrt p}<2\sqrt p.
 \tag{\(e^x-1\ge x\)}
\end{align*}
Add the three ranges. Their slack coefficients \(1\), \(16\), and
\(2\sqrt p\) are at most \(2\sqrt p\), proving the claim.
\end{proof}

\subsection{The leading constants}
\label{app:calibrated-standalone}

We prove the coefficients in Theorem~\ref{thm:main}: a finite
coefficient for Algorithm~\ref{alg:one-clock} at every \(p\), and the
upper asymptotic \((2+o(1))p/\log p\) for both algorithms.
Appendix~\ref{app:switch-constants} gives the finite coefficients of
Algorithm~\ref{alg:lp-hybrid}. Write \(\Delta=\Delta_T\), and let \(V_t\)
be the shifted loss used for tuning: \(V_t=S_t\) for
Algorithm~\ref{alg:one-clock}, and \(V_t=U_t\) for
Algorithm~\ref{alg:lp-hybrid}.

\begin{proof}[Proof of Theorem~\ref{thm:main}: the coefficients]
\emph{A finite coefficient for Algorithm~\ref{alg:one-clock}.}
Proposition~\ref{prop:tuning}(i) gives
\begin{equation}
 \Reg_T\le A+B W_p(S_T,d_*),\qquad S_T=\Delta+\Reg_T,
 \label{eq:calibrated-standalone-budget}
\end{equation}
where we may take
\[
 A=c+1/\sqrt2,\qquad
 B=\max\{5\cdot3^{1/2-1/p}\varepsilon_p^{-1/(p-1)},
              \sqrt6\,\bar\kappa_p/m_p\}+3\varepsilon_p\one\{p>2\}+1+1/\sqrt2.
\]
These constants depend only on \(p\), and \(A,B=O(1)\) as
\(p\to\infty\). The coefficient \(B\) includes the one crossover charge from
Proposition~\ref{prop:tuning}.

If \(\Delta\le d\), then, since \(W_p(S_T,d_*)\le S_T^{1/r}\),
\begin{align*}
 \Reg_T
 &\le A+BS_T^{1/r}
 \tag{by \eqref{eq:calibrated-standalone-budget}}\\
 &\le A+B\Delta^{1/r}+B(\Reg_T)_+^{1/r}.
 \tag{subadditivity of \(u^{1/r}\)}
\end{align*}
If \(\Reg_T>2(A+B\Delta^{1/r})\), then \(\Reg_T<2B\Reg_T^{1/r}\), so
\(\Reg_T<(2B)^{r/(r-1)}=(2B)^p\). Thus a coefficient depending only on
\(p\) suffices on the power branch. If \(\Delta\ge d\), then
\(W_p(S_T,d_*)\le d_*^{1/2-1/p}\sqrt{S_T}=\sqrt{m_pd^{1-2/p}S_T}\), so
\(\Reg_T\le A+B\sqrt{m_pd^{1-2/p}}\sqrt{\Delta+\Reg_T}\), and
Lemma~\ref{lem:exact-closure} gives
\begin{equation}
 \Reg_T\le\tfrac32(A+B^2m_pd^{1-2/p})+B\sqrt{m_pd^{1-2/p}\Delta}.
 \label{eq:calibrated-quadratic-closure}
\end{equation}
Since \(d^{1-2/p}\le d^{1-1/p}\le W_p(\Delta,d)=d^{1/2-1/p}\sqrt\Delta\), the coefficient
\(\tfrac32 A+\tfrac32B^2m_p+B\sqrt{m_p}\) suffices.
These finite coefficients remain bounded as \(p\downarrow2\).
At \(p=2\), \(m_2=1\), \(\sqrt6\,\bar\kappa_2<5\), and
\(W_2(S,d_*)=\sqrt S\). Hence \eqref{eq:calibrated-quadratic-closure}
holds for every \(\Delta\ge0\), with
\[
 A=\sqrt2+1/\sqrt2,\qquad B=6+1/\sqrt2.
\]
The coefficient \(\max\{\tfrac32(A+B^2),B\}=70.66\ldots<71\) therefore suffices.

\emph{Shifted loss on radial rounds.}
Take \(p\ge130\), so that \(\varepsilon_p<1/2\). We show that
\begin{equation}
 V_t+2\le\frac{3+\Reg_t^{\FTL}+2\Delta_t}{1-\varepsilon_p}
 \label{eq:calibrated-radial-counter}
\end{equation}
at every \(t\) for Algorithm~\ref{alg:lp-hybrid}, and whenever every
action up to \(t\) is radial for Algorithm~\ref{alg:one-clock}. For
Algorithm~\ref{alg:lp-hybrid},
Lemma~\ref{lem:single-safe}\ref{item:switch-counter} gives the stronger
\(U_t+2\le3+\Reg_t^{\FTL}/2+\Delta_t\). For
Algorithm~\ref{alg:one-clock}, the radial temperatures up to \(t\) are at
most \(\varepsilon_p(S_t+2)^{1/r}\), so
\begin{align*}
 S_t+2
 &=\Delta_t+\Reg_t+2\\
 &\le\Reg_t^{\FTL}+3+2\Delta_t+\varepsilon_p(S_t+2)^{1/r}
 \tag{Lemma~\ref{lem:radial-tracking}}\\
 &\le\Reg_t^{\FTL}+3+2\Delta_t+\varepsilon_p(S_t+2),
 \tag{\((S_t+2)^{1/r}\le S_t+2\)}
\end{align*}
and rearranging gives \eqref{eq:calibrated-radial-counter}.

\emph{FTL regret.}
Corollary~\ref{cor:ftl-low-dim} bounds FTL when
\(\Delta_t\le e^{\sqrt p}\). Above that scale, take
\(\bar R=e^{\sqrt p}-1\) in \eqref{eq:ftl-radius-cut}; then
\(b_p(\bar R)<2\sqrt p\) and \(2\bar R+2\le2\Delta_t\), so
\(\Reg_t^{\FTL}\le(2+2\sqrt p)\Delta_t\). Together these estimates give,
at every \(t\),
\begin{equation}
 \Reg_t^{\FTL}\le4+2d+(2+2\sqrt p)\Delta_t.
 \label{eq:calibrated-ftl-global}
\end{equation}

\emph{Entry into the quadratic phase.}
Suppose some FTRL action up to round \(T\) is quadratic. Then
\(V_n\ge d_*\) for some \(n<T\); let \(n\) be the first such index. Every
FTRL action up to round \(n\) is radial, so
\eqref{eq:calibrated-radial-counter} applies at \(n\), even if the
crossing overshoots \(d_*\), and
\begin{align*}
 (1-\varepsilon_p)d_*
 &\le(1-\varepsilon_p)(V_n+2)
 \tag{\(V_n\ge d_*\)}\\
 &\le3+\Reg_n^{\FTL}+2\Delta_n
 \tag{by \eqref{eq:calibrated-radial-counter}}\\
 &\le7+2d+(4+2\sqrt p)\Delta
 \tag{by \eqref{eq:calibrated-ftl-global}, \(\Delta_n\le\Delta\)}\\
 &\le9d+(4+2\sqrt p)\Delta.
 \tag{\(d\ge1\)}
\end{align*}
Consequently
\begin{equation}
 \Delta\ge L_p d,\qquad
 L_p=\frac{(1-\varepsilon_p)d_*/d-9}{4+2\sqrt p}
       \sim\frac{\bar\kappa_p}{2\sqrt p}\longrightarrow\infty.
 \label{eq:calibrated-entry-slack}
\end{equation}
Here \(m_p=\bar\kappa_p\), with \(\bar\kappa_p/\sqrt p\to\infty\), and
\(d_*/d=m_p^{p/(p-2)}\sim m_p\).

\emph{Small slack on the power branch.}
Suppose \(\Delta<\min\{d,e^{\sqrt p}\}\). For all sufficiently large
\(p\), \(L_p>1\), so \eqref{eq:calibrated-entry-slack} rules out quadratic
actions, and \eqref{eq:calibrated-radial-counter} holds at \(T\). Every
FTRL temperature is at most \(\varepsilon_p(V_T+2)^{1/r}\), and FTL rounds
play \(f(Y_{t-1})\). Hence
\begin{align*}
 \Reg_T
 &\le\Reg_T^{\FTL}+1+\Delta+\varepsilon_p(V_T+2)
 \tag{Lemma~\ref{lem:radial-tracking}}\\
 &\le\frac{\Reg_T^{\FTL}}{1-\varepsilon_p}
       +\frac{1+\varepsilon_p}{1-\varepsilon_p}\Delta
       +1+\frac{3\varepsilon_p}{1-\varepsilon_p}.
 \tag{by \eqref{eq:calibrated-radial-counter}}
\end{align*}
Take \(\bar R=p/\log^2(e+p)\) in \eqref{eq:ftl-radius-cut}.
Then \(2\bar R+2=o(p/\log p)\),
\(b_p(\bar R)\sim2p/\log p\), and
\(\Delta\le e^{1/\sqrt p}\Delta^{1/r}\).
Since \(\varepsilon_p\to0\), the coefficient of
\(1+\Delta^{1/r}\) is at most \((2+o(1))p/\log p\).
This estimate includes \(\Delta=0\).

\emph{The remaining slack.}
Suppose \(\Delta\ge\min\{d,e^{\sqrt p}\}\), so that \(\Delta\ge1\). If
\(\Delta<e^{\sqrt p}\), then \(d\le\Delta\), and
\eqref{eq:calibrated-ftl-global} gives
\(\Reg_T^{\FTL}\le(8+2\sqrt p)\Delta\); otherwise the radius cutoff above
gives \(\Reg_T^{\FTL}\le(2+2\sqrt p)\Delta\). Whenever
\eqref{eq:calibrated-radial-counter} holds at \(T\), with
\(\varepsilon_p<1/2\) and \(\Delta\ge1\) it gives
\[
 V_T\le2\bigl[3+(10+2\sqrt p)\Delta\bigr]\le q_p\Delta,
 \qquad q_p=26+4\sqrt p,
\]
and, by \eqref{eq:profile-scaling} and \(m_p\le p\),
\[
 W_p(V_T,d_*)\le\max\{q_p^{1/r},\sqrt{m_pq_p}\}\,W_p(\Delta,d)
 =O(p^{3/4})\,W_p(\Delta,d).
\]
This applies to Algorithm~\ref{alg:lp-hybrid}, whose regret is at most
\(A_{\mathrm{sw}}+B_{\mathrm{sw}}W_p(U_T,d_*)\) by
Lemma~\ref{lem:calibrated-cost}, and to Algorithm~\ref{alg:one-clock}
with no quadratic action up to \(T\), by
\eqref{eq:calibrated-standalone-budget}. After a quadratic action of
Algorithm~\ref{alg:one-clock}, \eqref{eq:calibrated-entry-slack} gives
\(\Delta\ge L_pd\ge d\) for large \(p\), so
\(W_p(\Delta,d)=d^{1/2-1/p}\sqrt\Delta\), and
\begin{align*}
 \frac{m_pd^{1-2/p}}{W_p(\Delta,d)}
 &=m_p\sqrt{d^{1-2/p}/\Delta}\\
 &\le\frac{m_p}{\sqrt{L_p}}
 \tag{\(d^{1-2/p}\le d\le\Delta/L_p\)}\\
 &=O(p^{1/4}\sqrt{\bar\kappa_p})=O(p^{3/4}).
 \tag{\(L_p\sim\bar\kappa_p/(2\sqrt p)\), \(\bar\kappa_p\le p\)}
\end{align*}
Equation~\eqref{eq:calibrated-quadratic-closure} then also gives a
coefficient \(O(p^{3/4})\). Every case of this regime therefore
contributes only \(o(p/\log p)\).

\emph{Conclusion.}
The cases exhaust all dimensions, horizons, and loss paths for both
algorithms. Every remainder depends only on \(p\), so the upper
asymptotic is uniform, and Theorem~\ref{thm:universal-lower} supplies the
matching lower bound.
\end{proof}

\section{Stochastic adaptation and local asymptotics}
\label{app:stochastic}

We separate population excess loss from the hindsight correction, then
control the empirical fluctuations and the quadratic perturbation.
Use the notation of Section~\ref{sec:stochastic-fast}, and write
\[
 \Pi_\mu=I-\frac{\mu\nabla h(\mu)^\top}{h(\mu)},\qquad
 \bar\theta_n=n^{-1}\sum_{t\le n}\theta_t=-Y_n/n.
\]
Lemma~\ref{lem:support-curvature}(a) identifies \(\Pi_\mu\) as the
oblique projection onto \(\{a:\ip{\nabla h(\mu)}a=0\}\) along \(\mu\), with
\(h(\Pi_\mu a)\le2h(a)\). Also \(\norm{\nabla h(\mu)}_p=1\) and \(\ip{\nabla h(\mu)}\mu=h(\mu)\).
Since \(\sum_{t\le T}(\theta_t-\bar\theta_T)=0\), summing the
Bregman definitions yields
\[
 \Delta_T=\sum_{t\le T}D_h(\theta_t,\bar\theta_T),\qquad
 \sum_{t\le T}D_h(\theta_t,\mu)
 =\Delta_T+T D_h(\bar\theta_T,\mu).
\]
Taking expectations gives \eqref{eq:stochastic-bregman-information},
using \(\E\bar\theta_T=\mu\) and \(h(T\bar\theta_T)=T h(\bar\theta_T)\).
For any causal learner, independence decomposes expected regret into
pseudo-regret and the hindsight correction:
\begin{equation}
 \E\Reg_T
 =\sum_{t=1}^T\E[h(\mu)+\ip\mu{x_t}]
  +\E h\left(\sum_{t=1}^T\theta_t\right)-T h(\mu).
 \label{eq:stochastic-hindsight-decomposition}
\end{equation}
Its last term is the forward remainder \(T\E D_h(\bar\theta_T,\mu)\),
linearised at \(\mu\), by \eqref{eq:stochastic-bregman-information}.
Throughout these proofs, \(b_t=\beta_1(S_{t-1})\) and
\(q_t=\beta_2(S_{t-1})\) denote the standalone temperatures.

\subsection{Stochastic signal and cancellation}
\label{app:stochastic-fast}

\begin{proof}[Proof of Theorem~\ref{thm:stochastic-signal}]
Constants implicit in \(\lesssim_p\) depend only on \(p\) and remain
bounded near \(p=2\).

\emph{Cancellation and transverse moments.}
For \(s=h(\theta)>0\), put \(u=\theta/s\). Since \(\Pi_\mu\mu=0\) and
\(D_h(u,\mu/h(\mu))=1-\ip{\nabla h(\mu)}u\),
\begin{align*}
 h(\Pi_\mu\theta)^2
 &=s^2h\bigl(\Pi_\mu(u-\mu/h(\mu))\bigr)^2
 \tag{\(\Pi_\mu\mu=0\)}\\
 &\le4s^2h\bigl(u-\mu/h(\mu)\bigr)^2
 \tag{Lemma~\ref{lem:support-curvature}(a)}\\
 &\le\frac{8s^2}{r-1}(1-\ip{\nabla h(\mu)}u)
 \tag{Lemma~\ref{lem:dual-convexity}}\\
 &\le\frac8{r-1}\{s-\ip{\nabla h(\mu)}\theta\}.
 \tag{\(s\le1\)}
\end{align*}
The same bound holds at \(s=0\). Define
\(V_\mu=\sum_i(\E|(\Pi_\mu\theta_1)_i|^2)^{r/2}\). Since
\(\E[h(\theta_1)-\ip{\nabla h(\mu)}{\theta_1}]=\gamma\),
\begin{align*}
 V_\mu
 &\le d^{1-r/2}\Bigl(\sum_i\E|(\Pi_\mu\theta_1)_i|^2\Bigr)^{r/2}
 \tag{concavity of \(a^{r/2}\)}\\
 &\le d^{1-r/2}\bigl(\E h(\Pi_\mu\theta_1)^2\bigr)^{r/2}
 \tag{\(\norm\cdot_2\le h\)}\\
 &\le d^{1-r/2}\Bigl(\frac{8\gamma}{r-1}\Bigr)^{r/2}.
 \tag{display above}
\end{align*}
Each coordinate of \(\Pi_\mu\bar\theta_n\) has mean zero and variance
\(n^{-1}\E|(\Pi_\mu\theta_1)_i|^2\), so
\(\E|Z|^r\le(\E|Z|^2)^{r/2}\), applied coordinatewise and summed, gives
\begin{equation}
 V_\mu\le d^{1-r/2}\left(\frac{8\gamma}{r-1}\right)^{r/2},\qquad
 \E h(\Pi_\mu\bar\theta_n)^r\le V_\mu n^{-r/2}.
 \label{eq:stochastic-transverse-moments}
\end{equation}

\emph{Norm remainders.}
Let \(P=h^r/r\), and let \(h(u)=1\). If \(h(a-u)\le1/2\), then
\(h(a)\ge1/2\) and
\begin{align*}
 h(a)^{r-1}D_h(u,a)
 &\le D_P(u,a)
 \tag{Lemma~\ref{lem:radial-decoupling}}\\
 &\le2^{2-r}h(a-u)^r/r.
 \tag{Lemma~\ref{lem:scalar-power}}
\end{align*}
Hence \(D_h(u,a)\le2h(a-u)^r\). If \(h(a-u)>1/2\), then
\(D_h(u,a)\le2<2^{r+1}h(a-u)^r\). In both cases
\(D_h(u,a)\le2^{r+1}h(a-u)^r\).
For \(A=\ip{\nabla h(\mu)}a>0\), the point \(a/A\) equals
\(\mu/h(\mu)+\Pi_\mu a/A\). Hence
\begin{align}
 D_h(\mu,a)
 &=h(\mu) D_h\left(\frac{\mu}{h(\mu)},\frac{\mu}{h(\mu)}+\frac{\Pi_\mu a}{A}\right)
 \tag{by \eqref{eq:support-normalisation}}\\
 &\le2^{r+1}h(\mu) A^{-r}h(\Pi_\mu a)^r.
 \label{eq:stochastic-norm-remainder}
\end{align}
The forward remainder satisfies
\(D_h(u+w,u)\le D_P(u+w,u)\le2^{2-r}h(w)^r/r\) when \(h(u)=1\), by the
same two lemmas.

\emph{The adaptive quadratic perturbation.}
Jensen's inequality gives \(\E h(Y_n)\ge nh(\mu)\), hence
\(\E\Delta_n\le\gamma n\). Since \(W_p(\Delta,d)\le\Delta^{1/r}\),
\begin{align}
 \E(S_n+2)
 &=\E\Delta_n+\E\Reg_n+2\notag\\
 &\le\gamma n+C_{\mathsf L}(p)\{1+(\gamma n)^{1/r}\}+2
 \tag{Theorem~\ref{thm:main} for Algorithm~\ref{alg:one-clock}, Jensen}\\
 &\lesssim_p1+\gamma n.
 \label{eq:stochastic-counter}
\end{align}
Set \(a_n=\bar\theta_n+q_{n+1}x_{n+1}/n\), \(n\ge1\).
Lemma~\ref{lem:effective-dual-point}, applied at \(Y_n=-n\bar\theta_n\),
gives the exact representation
\begin{equation}
 x_{n+1}=-\alpha_{n+1} f(a_n),\qquad
 \alpha_{n+1}=1-e^{-nh(a_n)/b_{n+1}}.
 \label{eq:stochastic-effective-mean}
\end{equation}
Here \(\alpha_{n+1}=0\) if \(a_n=0\), so the formula also covers that case.
Since \(h(x_{n+1})\le d^{1-2/p}\) and
\(q_{n+1}^2\le m_p(S_n+2)/\bar F_2\le2m_p(S_n+2)/d^{1-2/p}\),
\begin{align*}
 \E h\left(\Pi_\mu\frac{q_{n+1}x_{n+1}}n\right)^r
 &\le\frac{2^rd^{r(1-2/p)}}{n^r}\,\E q_{n+1}^r
 \tag{Lemma~\ref{lem:support-curvature}(a)}\\
 &\le\frac{2^rd^{r(1-2/p)}}{n^r}\Bigl(\frac{2m_p}{d^{1-2/p}}\Bigr)^{r/2}
   \bigl(\E(S_n+2)\bigr)^{r/2}
 \tag{concavity of \(a^{r/2}\)}\\
 &\lesssim_p d^{1-r/2}\{n^{-r}+\gamma^{r/2}n^{-r/2}\}.
 \tag{by \eqref{eq:stochastic-counter}, \(d^{r(1-2/p)/2}=d^{1-r/2}\)}
\end{align*}
Since \(h(\Pi_\mu a_n)^r\le2^{r-1}\{h(\Pi_\mu\bar\theta_n)^r
+h(\Pi_\mu q_{n+1}x_{n+1}/n)^r\}\), combining this with
\eqref{eq:stochastic-transverse-moments} gives
\begin{equation}
 \E h(\Pi_\mu a_n)^r
 \lesssim_p d^{1-r/2}\{n^{-r}+\gamma^{r/2}n^{-r/2}\}.
 \label{eq:stochastic-effective-moment}
\end{equation}

\emph{Population excess loss.}
Here the linearisation point is the effective mean \(a_n\).
Equation~\eqref{eq:stochastic-effective-mean} gives
\begin{align*}
 h(\mu)+\ip\mu{x_{n+1}}
 &=(1-\alpha_{n+1})h(\mu)+\alpha_{n+1}\{h(\mu)-\ip{f(a_n)}\mu\}\\
 &=(1-\alpha_{n+1})h(\mu)+\alpha_{n+1}D_h(\mu,a_n).
\end{align*}
Put \(B=\sqrt{m_p}{(d+2)^{1/r-1/2}}\), so \(B^2\le3m_pd^{1-2/p}\).
Hoeffding's inequality gives
\[
 \Pr(G_n^c)\le e^{-nh(\mu)^2/8}\quad\text{for}\quad
 G_n=\{\ip{\nabla h(\mu)}{\bar\theta_n}\ge h(\mu)/2\}.
\]
Equation~\eqref{eq:standalone-temperature-growth} gives
\(q_{n+1}\le2\sqrt{m_p n/\bar F_2}\) and \(b_{n+1}\le2B\sqrt n\). Thus
\begin{align*}
 h(q_{n+1}x_{n+1}/n)
 &\le\frac{2d^{1-2/p}}{n}\sqrt{\frac{m_pn}{\bar F_2}}
 \tag{\(h(x_{n+1})\le d^{1-2/p}\)}\\
 &\le\sqrt{8m_pd^{1-2/p}/n}.
 \tag{\(\bar F_2\ge d^{1-2/p}/2\)}
\end{align*}
Let \(n\ge N=\lceil128m_pd^{1-2/p}/h(\mu)^2\rceil\), so this is at most
\(h(\mu)/4\). On \(G_n\),
\begin{align*}
 h(a_n)\ge\ip{\nabla h(\mu)}{a_n}
 &\ge\ip{\nabla h(\mu)}{\bar\theta_n}-h(q_{n+1}x_{n+1}/n)
 \tag{H\"older}\\
 &\ge h(\mu)/2-h(\mu)/4=h(\mu)/4.
 \tag{\(G_n\), \(n\ge N\)}
\end{align*}
Hence, on \(G_n\), \(1-\alpha_{n+1}\le e^{-h(\mu)\sqrt n/(8B)}\), and
\eqref{eq:stochastic-norm-remainder} with \(A\ge h(\mu)/4\) gives
\(D_h(\mu,a_n)\le2^{3r+1}h(\mu)^{1-r}h(\Pi_\mu a_n)^r\). By
\eqref{eq:stochastic-effective-moment},
\[
 \E[(h(\mu)+\ip\mu{x_{n+1}})\one_{G_n}]
 \lesssim_p h(\mu)^{1-r}d^{1-r/2}\{n^{-r}+\gamma^{r/2}n^{-r/2}\}
       +h(\mu) e^{-h(\mu)\sqrt n/(8B)}.
\]
The excess is always at most \(2h(\mu)\). Including the initial excess
\(h(\mu)\), early rounds \(n<N\), and the events \(G_n^c\),
\begin{align}
 \sum_{t=1}^T\E[h(\mu)+\ip\mu{x_t}]
 &\le h(\mu)+2h(\mu)N+2h(\mu)\sum_{n\ge1}e^{-nh(\mu)^2/8}
   +\sum_{n=N}^{T-1}\E[(h(\mu)+\ip\mu{x_{n+1}})\one_{G_n}]\notag\\
 &\lesssim_p\frac{d^{1-2/p}}{h(\mu)}
   +h(\mu)^{1-r}d^{1-r/2}\sum_{n\ge1}n^{-r}
   +\frac{B^2}{h(\mu)}
   +h(\mu)^{1-r}d^{1-r/2}\gamma^{r/2}\mathcal H_r(T)
 \tag{see below}\\
 &\lesssim_p\frac{d^{1-2/p}}{h(\mu)}
       +h(\mu)^{1-r}d^{1-r/2}\gamma^{r/2}\mathcal H_r(T).
 \label{eq:stochastic-population-cost}
\end{align}
The second line uses \(h(\mu)\le1\le d^{1-2/p}\), \(\sum_ne^{-nh(\mu)^2/8}\le8/h(\mu)^2\),
and \(h(\mu)\sum_{n\ge1}e^{-h(\mu)\sqrt n/(8B)}\le128B^2/h(\mu)\). The
last uses \(\sum_{n\ge1}n^{-r}\le1+1/(r-1)\), \(d^{1-r/2}\le d^{1-2/p}\),
\(h(\mu)^{1-r}\le1/h(\mu)\), and \(B^2\le3m_pd^{1-2/p}\).
For FTL, use \(a_n=\bar\theta_n\), \(\alpha_{n+1}=1\) away from zero,
and zero action at zero. The same good-event argument with no quadratic
perturbation or damping proves \eqref{eq:stochastic-population-cost}
for FTL as well.

\emph{The hindsight correction.}
Write \(Z_T=\sum_{t\le T}(\theta_t-\mu)\) and \(z=T\mu+Z_T\). By
\eqref{eq:stochastic-hindsight-decomposition}, the hindsight correction
is \(\E D_h(z,\mu)=\E[h(z)-\ip{\nabla h(\mu)}z]\).
On \(G_T\), \(A=\ip{\nabla h(\mu)}z\ge Th(\mu)/2\) and
\(\Pi_\mu z=\Pi_\mu Z_T=T\Pi_\mu\bar\theta_T\), so
\begin{align*}
 D_h(z,\mu)
 &=A\,D_h\Bigl(\frac{\mu}{h(\mu)}+\frac{\Pi_\mu Z_T}A,\frac\mu{h(\mu)}\Bigr)
 \tag{by \eqref{eq:support-normalisation}}\\
 &\le\frac{2^{2-r}}rA^{1-r}h(\Pi_\mu Z_T)^r
 \tag{forward remainder}\\
 &\le\frac{2^{2-r}}r\Bigl(\frac{Th(\mu)}2\Bigr)^{1-r}T^r
   h(\Pi_\mu\bar\theta_T)^r.
 \tag{\(A\ge Th(\mu)/2\)}
\end{align*}
By \eqref{eq:stochastic-transverse-moments}, the contribution of \(G_T\)
is \(O(h(\mu)^{1-r}V_\mu T^{1-r/2})\).
On \(G_T^c\), \(D_h(z,\mu)=h(z)-Th(\mu)-\ip{\nabla h(\mu)}{Z_T}\le2h(Z_T)\), and
\begin{align*}
 \E[D_h(z,\mu)\one_{G_T^c}]
 &\le2\sqrt{\E h(Z_T)^2}\sqrt{\Pr(G_T^c)}
 \tag{Cauchy--Schwarz}\\
 &\le2\sqrt{d^{1-2/p}T}\,e^{-Th(\mu)^2/16}
 \tag{\(h^2\le d^{1-2/p}\norm\cdot_2^2\), \(\E\norm{Z_T}_2^2\le T\)}\\
 &=O(d^{1/2-1/p}/h(\mu)).
\end{align*}
For \(T\ge2\), \(\mathcal H_r(T)\ge(T-1)^{1-r/2}\ge T^{1-r/2}/2\).
This absorbs the hindsight correction into the second branch of
\eqref{eq:stochastic-cancellation-rate}; \(T=1\) is immediate.
Theorem~\ref{thm:main} transfers the FTL bound to the switching extension.

\emph{Conclusion.}
Finally, \(W_p(\cdot,d)\) is increasing and concave, so
Theorem~\ref{thm:main} and \(\E\Delta_T\le\gamma T\) give the first
branch by Jensen's inequality. Taking the smaller bound proves the
theorem. All constants used above depend on the small-loss coefficient,
\(m_p\), bounded functions of \(r\), and \(1/(r-1)=p-1\).
The finite-coefficient proof in Appendix~\ref{app:calibrated-standalone}
and \(m_p\to1\) ensure they remain bounded as \(p\downarrow2\).
\end{proof}

\subsection{Local curvature and stochastic limits}
\label{app:persistent-gap}

The variance term follows the second-order delta method
\citep[Sections~3.3 and~3.5]{vandervaart1998asymptotic}.
We give the moment and remainder estimates explicitly, and also track
the perturbation from adaptive regularisation.

\begin{proof}[Proof of Proposition~\ref{prop:local-stochastic}]
We apply second-order Taylor expansions to the empirical mean and its
regularisation perturbation. Put \(e_n=\bar\theta_n-\mu\). Bounded i.i.d.\ losses give
\(\E\norm{e_n}_2^k=O(n^{-k/2})\) for \(k=2,3,4\), and an exponentially
small probability of leaving any fixed neighbourhood of zero.

\emph{Hindsight correction.}
Since \(0\le D_h(\bar\theta_T,\mu)\le2h(e_T)\le4\), leaving the
neighbourhood contributes \(o(1)\), and
\begin{align*}
 \E h\left(\sum_{t\le T}\theta_t\right)-T h(\mu)
 &=T\E D_h(\bar\theta_T,\mu)
 \tag{by \eqref{eq:stochastic-bregman-information}}\\
 &=\tfrac T2\E[e_T^\top\nabla^2h(\mu)e_T]+O(T\E\norm{e_T}_2^3)+o(1)
 \tag{by \eqref{eq:support-taylor}}\\
 &=\tfrac12\operatorname{tr}(\nabla^2h(\mu)\Sigma)+o(1).
 \tag{\(T\E e_Te_T^\top=\Sigma\)}
\end{align*}

\emph{Effective means.}
It remains to compute population excess loss. Both learners have the form
\(x_{n+1}=-\alpha_{n+1}f(a_n)\). For FTL, set
\(a_n=\bar\theta_n\), \(\alpha_{n+1}=1\), and \(k=0\), including
\(a_n=0\) through \(f(0)=0\). For standalone FTRL, use
\eqref{eq:stochastic-effective-mean} and \(k=\sqrt{m_p\gamma/\bar F_2}\).
In this latter case, the strong law and Lemma~\ref{lem:empirical-limits} give
\(x_{n+1}\to-\nabla h(\mu)\) and \(q_{n+1}/\sqrt n\to k\) almost surely.
The deterministic temperature bounds upgrade the latter convergence to
every finite \(L^s\). Thus, for either learner,
\begin{equation}
 \sqrt n(a_n-\mu)=\sqrt n e_n-k\nabla h(\mu)+o_{L^2}(1).
 \label{eq:stochastic-local-shift}
\end{equation}
Consequently,
\begin{equation}
 n\E[(a_n-\mu)(a_n-\mu)^\top]\longrightarrow\Sigma+k^2\nabla h(\mu)\nabla h(\mu)^\top.
 \label{eq:stochastic-local-second-moment}
\end{equation}
The cross term vanishes because \(\E e_n=0\),
\(\sqrt n e_n\) is bounded in \(L^2\), and the perturbation error tends
to zero in \(L^2\).

\emph{Population excess loss.}
For both learners, \(\E\norm{a_n-\mu}_2^3=O(n^{-3/2})\), and the
probability of leaving a fixed small neighbourhood of \(\mu\) is
exponentially small. The deterministic temperature bounds also give
\(\E(1-\alpha_{n+1})=O(e^{-c\sqrt n})\) for some \(c>0\). Linearising at
\(a_n\) as in the proof of Theorem~\ref{thm:stochastic-signal},
\begin{align*}
 n\E[h(\mu)+\ip\mu{x_{n+1}}]
 &=n\E[\alpha_{n+1}D_h(\mu,a_n)]+nh(\mu)\E(1-\alpha_{n+1})\\
 &=\tfrac n2\E[(a_n-\mu)^\top\nabla^2h(\mu)(a_n-\mu)]+o(1)
 \tag{by \eqref{eq:support-taylor}}\\
 &\to\tfrac12\operatorname{tr}\!\left[\nabla^2h(\mu)\bigl(\Sigma+k^2\nabla h(\mu)\nabla h(\mu)^\top\bigr)\right].
 \tag{by \eqref{eq:stochastic-local-second-moment}}
\end{align*}
Harmonic summation and \eqref{eq:stochastic-hindsight-decomposition}
give both claimed limits, since the hindsight correction is bounded.
\end{proof}

\section{Baseline reductions}
\label{app:baseline-reductions}

Propositions~\ref{prop:general-prior}, \ref{prop:metagrad-prior},
and~\ref{prop:euclidean-prior} prove rows 2--4 of
Table~\ref{tab:method-comparison}. Throughout, \(\mathcal K\subset\R^d\)
is a centrally symmetric convex body with centre zero and support function
\(h\), \(h(\theta_t)\le1\), and
\(\Reg_T(u)=\sum_{t\le T}\ip{x_t-u}{\theta_t}\), so that
\(\Reg_T=\max_{u\in \mathcal K}\Reg_T(u)\) and \(S_T=\Delta_T+\Reg_T\) as
in~\eqref{eq:shifted}. Cone multiplicative weights and norm perturbations
give fixed-rate bounds in the learner's shifted loss, which we make
anytime by restarting; MetaGrad needs no restarts.

For \(b>0\) and a family of learners \(\mathsf L_B\), the doubling scheme
runs \(\mathsf L_{B_j}\) in epochs \(j=0,1,\ldots\) with \(B_j=2^jb\),
restarting after the first round at which the current epoch's shifted loss
exceeds \(B_j\).

\begin{lemma}[Doubling on the shifted loss]\label{lem:doubling}
Let \(b,C_0>0\). Suppose that for every \(B\ge b\) a learner
\(\mathsf L_B\), started at any round, satisfies \(\Reg(u)\le C_0\sqrt{bB}\)
for every \(u\in \mathcal K\) on every prefix of its run whose shifted loss is at
most \(B+2\). Then the doubling scheme satisfies
\(\Reg_T=O(b+\sqrt{b\Delta_T})\) for every \(T\), with a constant
depending only on \(C_0\).
\end{lemma}

\begin{proof}
Each round's shifted loss lies in \([0,2]\), so every epoch \(j\) has
shifted loss at most \(B_j+2\), and the assumed bound applies to it with
the same comparator \(u\). Let \(J\) be the last epoch containing an
action. If \(J\ge1\), epoch \(J-1\) has shifted loss greater than
\(B_J/2\); hence \(B_J\le2(b+S_T)\) in all cases. With
\(C_1=\sqrt2\,C_0/(1-2^{-1/2})\),
\begin{align*}
 \Reg_T(u)
 &\le C_0\sum_{j\le J}\sqrt{bB_j}
 \tag{assumed bound in each epoch}\\
 &\le\frac{C_0\sqrt{bB_J}}{1-2^{-1/2}}
 \tag{geometric series}\\
 &\le C_1\sqrt{b(b+S_T)}.
 \tag{\(B_J\le2(b+S_T)\)}
\end{align*}
At the best comparator, \(S_T=\Delta_T+\Reg_T\);
Lemma~\ref{lem:exact-closure} with \(A=0\), square-root coefficient
\(C_1\sqrt b\), and slack \(b+\Delta_T\) gives the claim.
\end{proof}

\subsection{The Euclidean bound from cone multiplicative weights}
\label{app:euclidean-prior}

In the Euclidean case, \(\Delta_T\) and \(\Reg_T\) use the support function
\(h=\norm{\cdot}_2\).

\begin{proposition}[Euclidean small-loss regret]\label{prop:euclidean-prior}
On \(B_2^d\) with \(\norm{\theta_t}_2\le1\), symmetric-cone multiplicative weights with
restarts on its realised shifted loss satisfies
\(\Reg_T=O(1+\sqrt{\Delta_T})\) simultaneously for every horizon,
with a universal dimension-independent constant.
\end{proposition}

\begin{proof}
\citet[Section~6]{canyakmaz2023multiplicative} represent \(B_2^d\) in the
Lorentz cone \(\mathcal L=\{(v,a)\in\R^d\times\R:\norm{v}_2\le a\}\). It
has rank two, identity \(e=(0,1)\), eigenvalues \(a\pm\norm v_2\), trace
\(2a\), product \((v,a)\circ(w,a')=(a'v+aw,\ip{v}{w}+aa')\), and inner
product \(\ipM{(v,a)}{(w,a')}{\mathcal L}=2(\ip{v}{w}+aa')\). Lift
\[
 \bar x_t=(x_t/2,1/2),\qquad \bar u=(u/2,1/2),\qquad
 \bar\theta_t=(\theta_t/2,h(\theta_t)/2).
\]
The lifted actions have trace one, the loss eigenvalues are
\(0,h(\theta_t)\in[0,1]\), and, since \(h=\norm{\cdot}_2\),
\[
 \bar\theta_t^2=h(\theta_t)\bar\theta_t,\qquad
 \ipM{\bar\theta_t}{\bar x_t}{\mathcal L}=\frac{\ell^+(x_t,\theta_t)}{2},\qquad
 \ipM{\bar\theta_t}{\bar x_t-\bar u}{\mathcal L}
 =\frac{\ip{x_t-u}{\theta_t}}2.
\]
From \(\bar x_1=e/2\), \citet[Theorem~5.1]{canyakmaz2023multiplicative}
therefore give, for \(0<\eta\le1\) and on every prefix,
\begin{align*}
 \Reg_T(u)
 &\le2\eta\sum_{t\le T}\ipM{\bar\theta_t^2}{\bar x_t}{\mathcal L}+\frac{2\log2}{\eta}
 \tag{cone regret bound}\\
 &=\eta\sum_{t\le T}h(\theta_t)\ell^+(x_t,\theta_t)+\frac{2\log2}{\eta}
 \tag{\(\bar\theta_t^2=h(\theta_t)\bar\theta_t\)}\\
 &\le\eta S_T+\frac{2\log2}{\eta}.
 \tag{\(h(\theta_t)\le1\)}
\end{align*}
The factor \(\log2\) is the logarithm of the rank, so the bound is free of
dimension. For \(B\ge1\), take \(\eta=1/(2\sqrt B)\). A prefix with
shifted loss at most \(B+2\le3B\) then has regret at most
\((\tfrac32+4\log2)\sqrt B\), and Lemma~\ref{lem:doubling} with \(b=1\)
proves the claim. Within an epoch, with \(Y\) the negative sum of its
previous losses, the normalised cone exponential plays
\(x_t=\tanh(\eta\norm{Y}_2/2)Y/\norm{Y}_2\) for \(Y\ne0\), and zero
otherwise.
\end{proof}

\subsection{A general-body bound from norm perturbations}
\label{app:general-prior}

Norm smoothing produces a potential whose gradient is the mean perturbed
leader and whose conjugate is a bounded regulariser. The construction
combines the norm mechanism of
\citet[Definition~4.1 and Theorem~4.3]{hardt2010geometry} with the
differential-stability argument of \citet[Lemma~3.1]{abernethy2019privacy}
and the smoothing framework of \citet[Sections~3--4]{abernethy2014smoothing}.
The norm-density ratio below is also the key stability estimate in
\citet[Appendix~A.5, Eq.~(34)]{wang2022adaptive}.

\begin{lemma}[Norm smoothing as bounded regularisation]
\label{lem:norm-perturbation-regulariser}
Let \(Z\) have density proportional
to \(e^{-h(z)}\), and choose a measurable support maximiser
\(a(w)\in\argmax_{x\in\mathcal K}\ip wx\). Define
\[
 \Phi_\beta(y)=\E h(y+\beta Z)-\beta d,\qquad \beta>0.
\]
Then \(F=\Phi_1^*\) is a bounded regulariser with cap \(d\),
\((\beta F)^*=\Phi_\beta\), and
\(\nabla\Phi_\beta(y)=\E a(y+\beta Z)\). For all \(y,\theta\in\R^d\),
\begin{equation}
 D_{\Phi_\beta}(y-\theta,y)
 \le\frac{h(\theta)}{2\beta}\ell^+(\nabla\Phi_\beta(y),\theta).
 \label{eq:norm-smoothing-stability}
\end{equation}
\end{lemma}

\begin{proof}
The volume identity \(\operatorname{vol}\{h\le t\}=t^d\operatorname{vol}\{h\le1\}\)
gives \(\E h(Z)=d\), and symmetry gives \(\E Z=0\). Jensen's inequality and
the triangle inequality give \(h(y)-\beta d\le\Phi_\beta(y)\le h(y)\).
Symmetry of \(Z\) and convexity of \(h\) give
\(\E h(y+\beta Z)\ge\E h(\beta Z)=\beta d\), so \(\Phi_\beta\ge0=\Phi_\beta(0)\).
Since \(h^*=\iota_{\mathcal K}\), conjugating these bounds at \(\beta=1\)
gives \(\iota_{\mathcal K}\le F\le\iota_{\mathcal K}+d\) and
\(F(0)=-\min\Phi_1=0\). Norms are differentiable almost everywhere with
subgradients in \(\mathcal K\), and \(Z\) has a density, so dominated
convergence gives \(\nabla\Phi_\beta(y)=\E a(y+\beta Z)\); homogeneity
gives \(\Phi_\beta(y)=\beta\Phi_1(y/\beta)=(\beta F)^*(y)\).

Put \(s=h(\theta)\) and
\(z(t)=s+\ip{\nabla\Phi_\beta(y-t\theta)}\theta\). By the triangle
inequality, the densities centred at \(y-t\theta\) and \(y\) have ratio at
least \(e^{-ts/\beta}\). Integrating against \(s+\ip{a(w)}\theta\ge0\)
gives \(z(t)\ge e^{-ts/\beta}z(0)\). Therefore
\begin{align*}
 D_{\Phi_\beta}(y-\theta,y)
 &=\int_0^1[z(0)-z(t)]\dif t
 \tag{fundamental theorem of calculus}\\
 &\le z(0)\int_0^1(1-e^{-ts/\beta})\dif t
 \tag{\(z(t)\ge e^{-ts/\beta}z(0)\)}\\
 &\le\frac{s}{2\beta}z(0).
 \tag{\(1-e^{-x}\le x\)}
\end{align*}
\end{proof}

\begin{proposition}[Regret from norm perturbations]
\label{prop:general-prior}
With exact expected perturbed-leader and support
oracles, there is a deterministic anytime learner satisfying
\(\Reg_T=O(d+\sqrt{d\Delta_T})\) on every loss sequence, with an absolute
implicit constant.
\end{proposition}
\begin{proof}
In one epoch, let \(v_t=-Y_t\) sum the epoch's losses, and choose a
measurable \(\chi(v)\in\argmin_{x\in\mathcal K}\ip vx\). By symmetry of
\(Z\), the mean perturbed leader is
\(x_t=\E\chi(v_{t-1}+\eta Z)=\nabla\Phi_\eta(Y_{t-1})\in\mathcal K\), with
\(\Phi_\eta\) from Lemma~\ref{lem:norm-perturbation-regulariser}.
For every comparator,
\begin{align*}
 \Reg_T(u)
 &\le\sum_{t\le T}D_{\Phi_\eta}(Y_t,Y_{t-1})+(h-\Phi_\eta)(Y_T)
 \tag{by \eqref{eq:one-clock-telescope}, fixed \(\eta\)}\\
 &\le\frac{S_T}{2\eta}+\eta d.
 \tag{by \eqref{eq:norm-smoothing-stability}, cap \(d\)}
\end{align*}
For \(B\ge d\), take \(\eta=\sqrt{B/d}\). A prefix with shifted loss at
most \(B+2\) then has regret at most
\(\tfrac12(B+2)\sqrt{d/B}+\sqrt{dB}\le3\sqrt{dB}\), and
Lemma~\ref{lem:doubling} with \(b=d\) proves the claim. The learner plays
means rather than sampled perturbations, so it is deterministic and the
pathwise bound also covers adaptive loss paths.
\end{proof}

\subsection{A general-body bound from MetaGrad}
\label{app:metagrad-prior}

\begin{proposition}[Regret from a second-order bound]
\label{prop:metagrad-prior}
MetaGrad Full \citep{vanerven2021metagrad}, run in
coordinates where \(B_2^d\subseteq\mathcal K\subseteq\sqrt d\,B_2^d\),
satisfies \(\Reg_T=O(d\log(T+1)+\sqrt{d\Delta_T\log(T+1)})\)
simultaneously for every horizon, with an absolute implicit constant.
\end{proposition}

\begin{proof}
By John's theorem for symmetric bodies, an invertible linear map \(A\)
gives \(B_2^d\subseteq A\mathcal K\subseteq\sqrt d\,B_2^d\). Replacing
actions by \(Ax\) and losses by \(A^{-\top}\theta\) preserves every inner
product, hence \(h\), \(\Delta_T\), and the regret, so we assume
\(B_2^d\subseteq\mathcal K\subseteq\sqrt d\,B_2^d\). Run MetaGrad Full with
gradients \(\theta_t\), parameter \(\sigma=\sqrt d\), and range bounds
\(b_t=2h(\theta_t)\), valid since \(\abs{\ip{w-x_t}{\theta_t}}\le2h(\theta_t)\)
for \(w,x_t\in\mathcal K\). Their running maximum \(B_T\) is at most two,
every \(u\in\mathcal K\) has \(\norm u_2^2\le d=\sigma^2\), and
\(\norm{\theta_t}_2\le h(\theta_t)\le B_T/2\). The matrix
\(\sum_{t\le T}\theta_t\theta_t^\top\) has rank \(k\le d\), and
\(k\mapsto k\log(1+x/k)\) is nondecreasing. With
\(\lceil z\rceil_+=\max\{\lceil z\rceil,1\}\) and
\(\Gamma_T=1+d\log(1+T/8)+2\log\lceil2\log_2T\rceil_+\),
\citet[Theorem~4]{vanerven2021metagrad} then gives, for every
\(u\in\mathcal K\) and every \(T\),
\[
 \Reg_T(u)\le\frac52\sqrt{V_T^u\,\Gamma_T}+10\Gamma_T+4,\qquad
 V_T^u=\sum_{t\le T}\ip{x_t-u}{\theta_t}^2 .
\]
At a best fixed action \(x_\star\), \(\Delta_T=\sum_{t\le T}\ell^+(x_\star,\theta_t)\)
by \eqref{eq:slack-boundary}, and each
\(\ip{x_t-x_\star}{\theta_t}=\ell^+(x_t,\theta_t)-\ell^+(x_\star,\theta_t)\)
is a difference of two numbers in \([0,2]\). Since
\((a-c)^2\le\max\{a,c\}^2\le2(a+c)\) for \(a,c\in[0,2]\),
\begin{align*}
 V_T^{x_\star}
 &\le2(S_T+\Delta_T)=2(\Reg_T+2\Delta_T),
 \tag{by \eqref{eq:shifted}}\\
 \Reg_T
 &\le10\Gamma_T+4+5\sqrt{\Gamma_T/2}\,\sqrt{2\Delta_T+\Reg_T}.
 \tag{MetaGrad bound at \(x_\star\)}
\end{align*}
Lemma~\ref{lem:exact-closure} with \(A=10\Gamma_T+4\),
\(b=5\sqrt{\Gamma_T/2}\), and slack \(2\Delta_T\) gives
\(\Reg_T\le\tfrac32(\tfrac{45}2\Gamma_T+4)+5\sqrt{\Gamma_T\Delta_T}\).
Since \(\Gamma_T=O(d\log(T+1))\), the claim follows.
\end{proof}

\section{Computing the quadratic actions}
\label{app:oracle-errors}

We compute the conjugate optimiser by solving one monotone scalar
equation. Each evaluation requires \(d\) independent coordinate solves.

Before crossover, \(\beta_2=0\) and the action is the explicit radial one.
After crossover, \(\beta_1=\beta_2=\beta>0\). We parameterise the action's radius by \(\chi=g'(\rho)\).
For \(y\ne0\) and \(\chi>0\), put \(\rho=1-e^{-\chi}\) and
define \(a_i(\chi)\ge0\) by
\[
 |y_i|=\beta(\rho a_i+\chi a_i^{p-1}),\qquad i=1,\ldots,d.
\]
The right side increases strictly from zero to infinity in \(a_i\), so
each coordinate has a unique solution, zero exactly when \(y_i=0\).
For \(y_i\ne0\), both coefficients increase with \(\chi\), making
\(a_i(\chi)\) continuous and strictly decreasing. As \(\chi\downarrow0\),
both coefficients vanish, forcing \(a_i(\chi)\to\infty\); as
\(\chi\to\infty\), the bound
\(a_i(\chi)^{p-1}\le |y_i|/(\beta\chi)\) forces it to zero.
The sum \(\sum_i a_i(\chi)^p\) is therefore continuous and strictly
decreasing from infinity to zero, so it equals one at exactly one root.
This reduction specifies the exact oracle.

At this root, set \(x_i=\rho\operatorname{sign}(y_i)a_i(\chi)\).
Then \(\norm{x}_p=\rho<1\), \(g'(\rho)=\chi\), and
\((\nabla\norm x_p)_i=\operatorname{sign}(y_i)a_i(\chi)^{p-1}\). Hence
\begin{align*}
 \bigl(\nabla F^{\beta,\beta}(x)\bigr)_i
 &=\beta g'(\rho)(\nabla\norm x_p)_i+\beta x_i\\
 &=\operatorname{sign}(y_i)\,
   \beta[\rho a_i(\chi)+\chi a_i(\chi)^{p-1}]\\
 &=y_i.
 \tag{coordinate equation}
\end{align*}
This verifies that \(x\) is the unique conjugate optimiser. At \(y=0\),
the optimiser is zero.

\end{document}